\documentclass{article}

\usepackage{microtype}
\usepackage{graphicx}
\usepackage{subfigure}
\usepackage{booktabs}
\usepackage{placeins}

\usepackage{hyperref}

\usepackage{amsmath,bm}
\usepackage{amsthm}
\usepackage{cite}
\usepackage{enumerate}
\usepackage{colortbl}
\usepackage{psfrag}
\usepackage[export]{adjustbox}
\usepackage{wrapfig}
\usepackage{xspace} 
\usepackage{amssymb}
\usepackage{multirow}
\usepackage{afterpage}
\usepackage{xcolor}
\usepackage{float}
\usepackage{pbox}
\usepackage{longtable}

\usepackage{enumitem}

\newenvironment{Itemize}{
    \begin{itemize}[leftmargin=*]
    \setlength{\itemsep}{0pt}
    \setlength{\topsep}{0pt}
    \setlength{\partopsep}{0pt}
    \setlength{\parskip}{0pt}}
{\end{itemize}}
\usepackage[most]{tcolorbox}
\tcbset{
    frame code={}
    center title,
    left=0pt,
    right=0pt,
    top=0pt,
    bottom=0pt,
    colframe=white,
    width=\dimexpr\textwidth\relax,
    enlarge left by=0mm,
    boxsep=5pt,
    arc=0pt,outer arc=0pt,
}

\usepackage{xcolor}
\colorlet{darkgreen}{green!65!black}
\colorlet{darkblue}{blue!75!black}
\colorlet{darkred}{red!80!black}
\definecolor{lightblue}{HTML}{0071bc}
\definecolor{lightgreen}{HTML}{39b54a}
\definecolor{manyshot}{HTML}{6969ff}
\definecolor{medshot}{HTML}{f7c600}
\definecolor{fewshot}{HTML}{ff6969}
\definecolor{mypurple}{HTML}{412F8A}
\definecolor{myorange}{HTML}{fc8e62}

\definecolor{deemph}{gray}{0.55}

\global\long\def\P{\mathbb{P}}

\newcommand{\indep}{\perp \!\!\! \perp}

\newcommand{\tabref}[1]{Table~\ref{#1}}
\newcommand{\secref}[1]{Sec.~\ref{#1}}
\newcommand{\figref}[1]{Fig.~\ref{#1}}

\newcommand{\eqnref}[1]{Eqn.~(\ref{#1})}

\renewcommand\ttdefault{cmtt}

\newcommand{\waterbirds}{\texttt{Waterbirds}\xspace}
\newcommand{\celeba}{\texttt{CelebA}\xspace}
\newcommand{\civilcomments}{\texttt{CivilComments}\xspace}
\newcommand{\multinli}{\texttt{MultiNLI}\xspace}
\newcommand{\metashift}{\texttt{MetaShift}\xspace}
\newcommand{\imagenetbg}{\texttt{ImageNetBG}\xspace}
\newcommand{\nicopp}{\texttt{NICO++}\xspace}
\newcommand{\mimiccxr}{\texttt{MIMIC-CXR}\xspace}
\newcommand{\mimicnotes}{\texttt{MIMICNotes}\xspace}
\newcommand{\chexpert}{\texttt{CheXpert}\xspace}
\newcommand{\cxrmultisite}{\texttt{CXRMultisite}\xspace}
\newcommand{\living}{\texttt{Living17}\xspace}

\renewcommand{\paragraph}[1]{\vspace{1.25mm}\noindent\textbf{#1}}
\newcommand{\grayrow}{\rowcolor[gray]{.9}}
\definecolor{baselinecolor}{gray}{.95}

\newcommand{\thoughts}[1]{\textcolor{deemph}{#1}}

\usepackage[accepted]{icml2023}

\usepackage[textsize=tiny]{todonotes}

\icmltitlerunning{Change is Hard: A Closer Look at Subpopulation Shift}

\begin{document}

\twocolumn[
\icmltitle{Change is Hard: A Closer Look at Subpopulation Shift}



\icmlsetsymbol{equal}{*}

\begin{icmlauthorlist}
\icmlauthor{Yuzhe Yang}{equal,mit}
\icmlauthor{Haoran Zhang}{equal,mit}
\icmlauthor{Dina Katabi}{mit}
\icmlauthor{Marzyeh Ghassemi}{mit}
\end{icmlauthorlist}

\icmlaffiliation{mit}{MIT CSAIL}
\icmlcorrespondingauthor{Yuzhe Yang}{yuzhe@mit.edu}

\icmlkeywords{Machine Learning, ICML}

\vskip 0.3in
]



\printAffiliationsAndNotice{\icmlEqualContribution} 

\begin{abstract}
Machine learning models often perform poorly on \emph{subgroups} that are underrepresented in the training data.
Yet, little is understood on the variation in mechanisms that cause subpopulation shifts, and how algorithms generalize across such diverse shifts at scale.
In this work, we provide a fine-grained analysis of subpopulation shift.
We first propose a unified framework that dissects and explains common shifts in subgroups.
We then establish a comprehensive benchmark of 20 state-of-the-art algorithms evaluated on 12 real-world datasets in vision, language, and healthcare domains.
With results obtained from training over 10,000 models, we reveal intriguing observations for future progress in this space.
First, existing algorithms only improve subgroup robustness over certain types of shifts but not others.
Moreover, while current algorithms rely on group-annotated validation data for model selection, we find that a simple selection criterion based on worst-class accuracy is surprisingly effective even without any group information.
Finally, unlike existing works that solely aim to improve worst-group accuracy (WGA), we demonstrate the fundamental tradeoff between WGA and other important metrics, highlighting the need to carefully choose testing metrics.
Code and data are available at: {\fontsize{9.5}{11.5}\selectfont \url{https://github.com/YyzHarry/SubpopBench}}.

\end{abstract}

\vspace{-0.7cm}
\section{Introduction}
\vspace{-0.1cm}
\label{sec:intro}
Machine learning models frequently exhibit drops in performance under the presence of distribution shifts~\cite{quinonero2008dataset}. Constructing machine learning models that are robust to these shifts is critical to the safe deployment of such models in the real-world~\cite{amodei2016concrete}. One ubiquitous type of distribution shift is \textit{subpopulation shift}, which is characterized by changes in the proportion of some subpopulations between training and deployment \cite{koh2021wilds}.
In such settings, models may have high overall performance but still perform poorly in rare subgroups \cite{zhang2020hurtful, hashimoto2018fairness}.

A well-studied type of subpopulation shift occurs when data contains \textit{spurious correlations} \cite{geirhos2020shortcut} -- non-causal relationships between the input and the label which may shift in deployment \cite{simon1954spurious}.
For example, image classifiers frequently make use of non-robust features such as image backgrounds \cite{xiao2016learning}, textures \cite{geirhos2018imagenet}, and erroneous markings \cite{degrave2021ai}.
However, there has been little work in defining subpopulation shift in a holistic way, understanding \emph{when} these shifts happen, and \emph{how} state-of-the-art (SOTA) algorithms generalize under diverse and realistic shifts.
Subpopulation shift can encompass a much wider array of underlying mechanisms. 
First, different attributes in data often exhibit skewed distributions, inevitably causing \emph{attribute imbalance} \cite{martinez2021blind}.
Moreover, certain labels can have significantly fewer observations, where such long-tailed label distribution induces severe \emph{class imbalance} \cite{liu2019large}.
Finally, certain attributes may have no training data at all, which motivates the need for \emph{attribute generalization} to unseen subpopulations \cite{santurkar2020breeds}.

In this work, we systematically investigate subpopulation shift in realistic evaluation settings.
We first formalize a generic framework of subpopulation shift, which decomposes \emph{attribute} and \emph{class} to enable fine-grained analyses.
We demonstrate that this modeling covers and explains the aforementioned common subgroup shifts, which are basic units of building more complex shifts that arise in real data.
Using this framework, we can quantify the type and degree of different shift components in each given dataset.

We establish a realistic and comprehensive benchmark of subpopulation shift, consisting of \textbf{20} SOTA algorithms that span different learning strategies and \textbf{12} real-world datasets in vision, language, and healthcare domains. 
While existing analysis on subpopulation shift either focus on a single shift type, or have limited severity, our benchmark provides a much larger set of datasets that cover different types of realistic subgroup shifts. Our experimental framework can be easily extended to include new methods, shifts, and datasets.

Our work also evaluates current methods across different settings including attribute availability in training and/or validation set, model selection strategies, and a wide range of metrics for understanding subpopulation shift in-depth.
With the established framework and over 10K trained models, we reveal intriguing observations for future research.

Concretely, we make the following contributions:
\vspace{-0.3cm}
\begin{Itemize}
    \item We formalize a unified framework for subpopulation shift which defines basic types of shift, explains when and why shifts happen, and quantifies their degrees.
    \vspace{1.5pt}
    \item We set up a comprehensive and realistic benchmark for systematic subpopulation shift evaluation, with 20 SOTA methods and 12 diverse datasets across various domains.
    \vspace{1.5pt}
    \item Based on over 10K trained models, we verify that current algorithms only advance subgroup robustness over certain types of shift identified by our framework, but not others.
    \vspace{1.5pt}
    \item We confirm that while successful algorithms rely on the access to group information for model selection, a simple criterion based on worst-class accuracy is surprisingly effective even without group-annotated validation data.
    \vspace{1.5pt}
    \item We establish the fundamental tradeoff between worst-group accuracy (WGA) and important metrics such as worst-case precision, highlighting the need to rethink evaluation metrics in subpopulation shift beyond WGA.
\end{Itemize}

\vspace{-0.35cm}
\section{Related Work}
\vspace{-0.05cm}
\label{sec:related-work}
\textbf{Subpopulation Shift.}
Machine learning models frequently experience performance degradation under \textit{subpopulation shift}, where the proportion of some subpopulations differ between the training and test \cite{koh2021wilds, cai2021theory}. Depending on the definition of such subpopulations, this could lead to vastly different problem settings. Prior works largely focus on the case of shortcut learning \cite{geirhos2020shortcut}, where subpopulations are defined as the product of attributes and labels. In such settings, models trained to minimize overall loss tend to learn spurious correlations, resulting in poor performance in the minority subpopulation \cite{degrave2021ai, joshi2022all}. There have been a large set of methods developed to address this scenario, both when the attribute is known \cite{yao2022improving, sagawa2019distributionally, izmailov2022feature, nam2022spread, menon2020overparameterisation, gowda2021pulling}, and unknown \cite{liu2021just, creager2021environment, idrissi2022simple, han2022umix}.

However, subpopulations may also be defined using only the label. This setting corresponds to class-imbalanced learning, which has also been well studied with extensive proposed methods \cite{yang2020rethinking, yang2021delving, yang2022multi, cao2019learning, cui2019class, li2021targeted}.

Finally, when subpopulations are defined based on a particular attribute (e.g., demographic group) \cite{pfohl2022comparison, zong2022medfair}, the objective of maximizing performance for the worst-case group then becomes identical to minimax fairness \cite{lahoti2020fairness, martinez2020minimax}.

In this work, we present a unified framework of subpopulation shift across these aforementioned scenarios.

\textbf{Distribution Shift Benchmarks.}
There have been few prior works which benchmark the performance of subpopulation shift methods. \citet{koh2021wilds} proposed the WILDS benchmark for domain generalization and subpopulation shift, though they only evaluated four methods over five datasets. \citet{zhang2022nico++} and \citet{gulrajani2020search} proposed the NICO++ and DomainBed benchmarks respectively for domain generalization, and we adapt elements of their benchmark into our subpopulation shift evaluation. \citet{santurkar2020breeds} proposed the BREEDS benchmark, which consists of multiple datasets constructed from ImageNet \cite{deng2009imagenet} using the WordNet hierarchy \cite{miller1995wordnet}, aiming to evaluate generalization across unseen attributes.
Finally, \citet{wiles2021fine} conducted a similar analysis in the general distribution shift setting on four synthetic and two real-world datasets.

Our work differs from these prior works by evaluating a much larger set of algorithms that span different categories on many more real-world datasets. We further define, dissect and quantify the type and degree of shift components in each dataset, and relate it to the performance of each method. In addition, we analyze important yet overlooked factors such as model selection criteria and metrics to evaluate against, and reveal intriguing properties in subpopulation shift.

\vspace{-0.05cm}
\section{Unified Framework of Subpopulation Shift}
\label{sec:method}
\begin{table*}[ht]
\setlength{\tabcolsep}{5pt}
\caption{Formulation summary of basic types of subpopulation shift under our framework.
}
\vspace{-6pt}
\label{table:basic-type-subpop}
\small
\begin{center}
\resizebox{1\textwidth}{!}{
\begin{tabular}{lccc}
\toprule[1.5pt]
    \textbf{Subpopulation Shift Type}  & \textbf{Attribute Bias} & \textbf{Class Bias} & \textbf{Impact on Classification Model} \\ \midrule
Spurious Correlations (SC)
    & \begin{tabular}[c]{@{}c@{}}$p_{\text{train}}(a|y,\mathbf{x}_{\text{core}}) \gg p_{\text{train}}(a|\mathbf{x}_{\text{core}})$\\ $p_{\text{test}}(a|y,\mathbf{x}_{\text{core}}) = p_{\text{test}}(a|\mathbf{x}_{\text{core}})$\end{tabular}
    & $-$
    & $\frac{\P(a|y,\mathbf{x}_{\text{core}})}{\P(a|\mathbf{x}_{\text{core}})} \gg 1 \ \Rightarrow \ \P(y|\mathbf{x}) \uparrow$
    \\ \midrule
Attribute Imbalance (AI)
    & \begin{tabular}[c]{@{}c@{}}$p_{\text{train}}(a|y,\mathbf{x}_{\text{core}}) \gg p_{\text{train}}(a'|y, \mathbf{x}_{\text{core}})$\\ $p_{\text{test}}(a|y,\mathbf{x}_{\text{core}}) = p_{\text{test}}(a'|y, \mathbf{x}_{\text{core}})$\end{tabular}
    & $-$
    & $\frac{\P(a|y,\mathbf{x}_{\text{core}})}{\P(a|\mathbf{x}_{\text{core}})} \gg \frac{\P(a'|y,\mathbf{x}_{\text{core}})}{\P(a'|\mathbf{x}_{\text{core}})} \ \Rightarrow \ \P(y | \mathbf{x}_{\text{core}}, a) \gg \P(y | \mathbf{x}_{\text{core}}, a')$
    \\ \midrule
Class Imbalance (CI)
    & $-$
    & \begin{tabular}[c]{@{}c@{}}$p_{\text{train}}(\mathbf{Y}=y) \gg p_{\text{train}}(\mathbf{Y}=y')$\\ $p_{\text{test}}(\mathbf{Y}=y) = p_{\text{test}}(\mathbf{Y}=y')$\end{tabular}
    & $\P(y) \gg \P(y') \ \Rightarrow \ \P(y | \mathbf{x}) \gg \P(y' | \mathbf{x})$
    \\ \midrule
Attribute Generalization (AG)
    & \begin{tabular}[c]{@{}c@{}}$p_{\text{train}}(a|y,\mathbf{x}_{\text{core}}) = 0, \forall a\in \mathbb{A}^{\text{unseen}}$\\ $p_{\text{test}}(a|y,\mathbf{x}_{\text{core}}) > 0, \forall a \in \mathbb{A}$\end{tabular}
    & Unconstrained
    & Generalize to $\mathbb{A}^{\text{unseen}}$ \\
\bottomrule[1.5pt]
\end{tabular}
}
\end{center}
\vspace{-0.4cm}
\end{table*}

\textbf{Problem Setup.}
In the general subpopulation shift setting, given input $\mathbf{x} \in \mathcal{X}$ and label $y \in \mathcal{Y}$, the goal is to learn $f: \mathcal{X} \rightarrow \mathcal{Y}$. In addition, there exist attributes $a_1, ..., a_i, ..., a_m$, $a_i \in \mathcal{A}_i$, which may or may not be available when learning $f$.
Then, discrete subpopulations can be defined based on the attribute and label, by some function $h : \mathcal{A} \times \mathcal{Y} \rightarrow \mathcal{G}$.

Let $\ell(y, f(\mathbf{x})) \rightarrow \mathbb{R}$ be a loss function. Consider the source distribution where $(\mathbf{x}, y)$ are drawn as a mixture of group-wise distributions: $P_{src} = \sum_{g \in \mathcal{G}} \alpha_g P_g$, where $\alpha \in \Delta_{|\mathcal{G}|}$. Further, consider some target distribution which is not observed: $P_{tar} = \sum_{g \in \mathcal{G}} \beta_g P_g$, where $\beta \in \Delta_{|\mathcal{G}|}$. The objective of subpopulation shift is to find \cite{sagawa2020groupdro}:
\begin{equation*}
f^* = \arg\min_{f} \sup_{\beta \in \Delta_{|\mathcal{G}|}} \mathbb{E}_{(\mathbf{x}, y) \sim_{P_{tar}}} [\ell(y,  f(\mathbf{x}))].
\end{equation*}
This objective is equivalent to minimizing risk for the worst-case group \cite{sagawa2020groupdro}, i.e.,
\begin{equation*}
f^* = \arg\min_{f} \max_{g \in \mathcal{G}} \mathbb{E}_{(\mathbf{x}, y) \sim_{P_{g}}} [\ell(y,  f(\mathbf{x}))].
\end{equation*}

\vspace{-0.3cm}
\subsection{A Generic Framework for Subpopulation Shift}
\vspace{-0.05cm}
\label{subsec:unified-framework-subpop}

As motivated earlier, both attribute $a$ and label $y$ can have specific skewed distributions, resulting in distinct types of subpopulation shift.
To this end, we propose to decompose the effect of $a$ and $y$ given a multi-group dataset, and characterize general subpopulation shift into several \emph{\textbf{basic shift}} components for fine-grained interpretation.

Specifically, we view each input $\mathbf{x}$ as being fully described or generated from a set of underlying core features $\mathbf{x}_{\text{core}}$ (representing the label) and a list of attributes $\mathbf{a}$ \cite{wang2021self, tang2022invariant}.
Here, $\mathbf{x}_{\text{core}}$ denotes the underlying invariant components that are label-specific and support robust classification, whereas attributes $\mathbf{a}$ may have inconsistent distributions and are not label-specific.
Such modeling helps us disentangle the attributes and examine how they affect the classification results $\P(y | \mathbf{x})$.
Following Bayes' theorem, we can rewrite the classification model as:
\begin{align}
\label{eq:classification-model}
    \P(y | \mathbf{x}) & = \ \frac{\P(\mathbf{x} | y)}{\P(\mathbf{x})} \cdot \P(y) \nonumber\\
    & = \ \frac{\P(\mathbf{x}_{\text{core}}, \mathbf{a} | y)}{\P(\mathbf{x}_{\text{core}}, \mathbf{a})} \cdot \P(y) \nonumber\\
    & = \ \thoughts{\underbrace{\frac{\P(\mathbf{x}_{\text{core}} | y)}{\P(\mathbf{x}_{\text{core}})}}_{\text{PMI}}}\ \cdot \ \underbrace{\frac{\P(\mathbf{a} | y, \mathbf{x}_{\text{core}})}{\P(\mathbf{a} | \mathbf{x}_{\text{core}})}}_{\text{attribute}}\ \cdot \ \underbrace{\P(y)}_{\text{class}},
\end{align}
where the first term in \eqnref{eq:classification-model} represents the pointwise mutual information (PMI) between $\mathbf{x}_{\text{core}}$ and $y$, the second term corresponds to the potential bias arising in the \textbf{attribute} distribution, and the third term explains the potential bias arising in the \textbf{class} (label) distribution.
Given invariant $\mathbf{x}_{\text{core}}$ between training and testing distributions, we can ignore changes in first term (which is a robust indicator), and focus on how the second and third term, i.e., the \emph{attribute} and \emph{class}, influence the outcomes under subpopulation shift.

More formally, assuming the mutual independence and conditional independence across different attributes $a_i$ \cite{wiles2021fine}, we can further decompose the attribute term into a fine-grained version:
\begin{equation}
\label{eq:attribute-term-fine-grained}
    \frac{\P(\mathbf{a} | y, \mathbf{x}_{\text{core}})}{\P(\mathbf{a}  | \mathbf{x}_{\text{core}})} \triangleq \prod_{a_i\in \mathbf{a}} \frac{\P(a_i | y, \mathbf{x}_{\text{core}})}{\P(a_i  | \mathbf{x}_{\text{core}})},
\end{equation}
where each $a_i$ corresponds to an attribute. Note that for benign attributes that are independent of $y$ (i.e., $a_i \indep y, \forall a_i \in \mathbf{a}_{\text{benign}}$), we have $\P(a_i | y, \mathbf{x}_{\text{core}}) = \P(a_i  | \mathbf{x}_{\text{core}})$, indicating that the attribute term in \eqnref{eq:attribute-term-fine-grained} is only driven by \emph{biased} attributes that are label-dependent.

Using the formulation of ``attribute-class'' decomposition, we can intuitively explain \emph{when} do common subpopulation shifts happen, and \emph{how} they affect the classification results.

\renewcommand\ttdefault{cmtt}

\begin{table*}[ht]
\setlength{\tabcolsep}{8pt}
\caption{Overview of the datasets for evaluating subpopulation shift. Detailed statistics and example data are provided in Appendix \ref{appendix-sec:benchmark-details}.}
\vspace{-2pt}
\label{table:dataset-overview}
\small
\begin{center}
\resizebox{1\textwidth}{!}{
\begin{tabular}{lcccccccccccc}
\toprule[1.5pt]
\multicolumn{1}{l}{\multirow{2.5}{*}{\textbf{Dataset}}} & \multirow{2.5}{*}{\textbf{Data type}} & \multirow{2.5}{*}{\textbf{\# Attr.}} & \multirow{2.5}{*}{\textbf{\# Classes}} & \multirow{2.5}{*}{\textbf{\# Train set}} & \multirow{2.5}{*}{\textbf{\# Val. set}} & \multirow{2.5}{*}{\textbf{\# Test set}} & \multirow{2.5}{*}{\textbf{Max group}} & \multirow{2.5}{*}{\textbf{Min group}} & \multicolumn{4}{c}{\textbf{Shift type}} \\ \cmidrule(l){10-13} 
 & & & & & & & & & \textbf{SC} & \textbf{AI} & \textbf{CI} & \textbf{AG} \\ \midrule\midrule
\waterbirds & Image & 2 & 2 & 4795 & 1199 & 5794 & 3498 (73.0\%) & 56 (1.2\%) & \checkmark & \checkmark & \checkmark & \\ \midrule
\celeba & Image & 2 & 2 & 162770 & 19867 & 19962 & 71629 (44.0\%) & 1387 (0.9\%) & \checkmark & & \checkmark & \\ \midrule
\metashift & Image & 2 & 2 & 2276 & 349 & 874 & 789 (34.7\%) & 196 (8.6\%) & \checkmark & & & \\ \midrule
\imagenetbg & Image & N/A & 9 & 183006 & 7200 & 4050 & N/A & N/A & & & & \checkmark \\ \midrule
\nicopp & Image & 6 & 60 & 62657 & 8726 & 17483 & 811 (1.3\%) & 0 (0.0\%) & & \checkmark & \checkmark & \checkmark \\ \midrule
\living & Image & N/A & 17 & 39780 & 4420 & 1700 & N/A & N/A & & & & \checkmark \\ \midrule
\multinli & Text & 2 & 3 & 206175 & 82462 & 123712 & 67376 (32.7\%) & 1521 (0.7\%) & & \checkmark & & \\ \midrule
\civilcomments & Text & 8 & 2 & 148304 & 24278 & 71854 & 31282 (21.1\%) & 1003 (0.7\%) & & \checkmark & \checkmark & \\ \midrule
\mimicnotes & Clinical text & 2 & 2 & 16149 & 3229 & 6460 & 8359 (51.8\%) & 676 (4.2\%) & & & \checkmark & \\ \midrule
\mimiccxr & Chest X-rays & 6 & 2 & 303591 & 17859 & 35717 & 68575 (22.6\%) & 7846 (2.6\%) & & \checkmark & & \\ \midrule
\chexpert & Chest X-rays & 6 & 2 & 167093 & 22280 & 33419 & 51606 (30.9\%) & 506 (0.3\%) & & \checkmark & \checkmark & \\ \midrule
\cxrmultisite & Chest X-rays & 2 & 2 & 338134 & 19891 & 39781 & 299089 (88.5\%) & 574 (0.2\%) & \checkmark & \checkmark & \checkmark & \\
\bottomrule[1.5pt]
\end{tabular}
}
\end{center}
\vspace{-0.3cm}
\end{table*}

\vspace{-0.15cm}
\subsection{Characterizing Basic Types of Subpopulation Shift}
\vspace{-0.05cm}
\label{subsec:basic-subpop-types}

We formally define and characterize four basic types of subpopulation shift using our framework: \emph{spurious correlations}, \emph{attribute imbalance}, \emph{class imbalance}, and \emph{attribute generalization} (see \tabref{table:basic-type-subpop}). In practice, we note that dataset often consists of multiple types of shift instead of one.
The four cases constitute the \emph{basic} shift units, and are important elements to explain complex subgroup shifts in real data.

\textbf{Spurious Correlations (SC).}
Spurious correlations happen when certain $a$ is spuriously correlated with $y$ in training but not in test data.
Under our framework, it implies that $p_{\text{train}}(a|y,\mathbf{x}_{\text{core}}) \gg p_{\text{train}}(a|\mathbf{x}_{\text{core}})$, which is not true of $p_{\text{test}}$.
As a result, it introduces bias to the \emph{attribute term}, which induces higher prediction confidence for certain label once given its spuriously correlated attribute (details in \tabref{table:basic-type-subpop}).

\textbf{Attribute Imbalance (AI).}
Attributes often incur biased distributions in the wild. In our framework, it happens when certain attributes are sampled with a much smaller probability than others in $p_{\text{train}}$, but not in $p_{\text{test}}$. To disentangle the effect of labels, we assume no class bias under this basic shift.
As such, it again affects the \emph{attribute term} in \eqnref{eq:classification-model} where $p_{\text{train}}(a|y,\mathbf{x}_{\text{core}}) \gg p_{\text{train}}(a'|y, \mathbf{x}_{\text{core}})$, causing lower prediction confidence for underrepresented attributes.

\textbf{Class Imbalance (CI).}
Similarly, class labels can exhibit imbalanced distributions, causing lower preference for minority labels. Within our framework, CI can be explained by biasing the \emph{class term} in $p_{\text{train}}$, leading to higher prediction confidence for majority classes.

\textbf{Attribute Generalization (AG).}
Certain attributes can be totally missing in $p_{\text{train}}$, but present in $p_{\text{test}}$, which motivates the need for attribute generalization. In our framework, this translates to $p_{\text{train}}(a|y,\mathbf{x}_{\text{core}}) = 0, a\in \mathbb{A}^{\text{unseen}}$, yet we have $p_{\text{test}}(a|y,\mathbf{x}_{\text{core}}) > 0$. AG requires learning robust $\mathbf{x}_{\text{core}}$ in order to generalize across unseen attributes, which is harder but more ubiquitous in real data \cite{santurkar2020breeds}.

\section{Benchmarking Subpopulation Shift}
\label{sec:experiment}
\textbf{Datasets.}
We explore subpopulation shift using \textbf{12} real-world datasets from a variety of modalities and tasks.
First, for \textbf{vision} datasets, we use \waterbirds \cite{wah2011caltech} and \celeba \cite{liu2015deep}, which are commonly used in the spurious correlation literature \cite{liu2021just}. Similarly, we use the \metashift cats \emph{vs.} dogs dataset~\cite{liang2022metashift}.
We further convert the ImageNet backgrounds challenge (\imagenetbg) \cite{xiao2020noise}, the \nicopp \cite{zhang2022nico++} benchmark, and the \living dataset from the BREEDS benchmark \cite{santurkar2020breeds} for subpopulation shift.
Further, for \textbf{language} understanding datasets, we leverage \civilcomments \cite{borkan2019nuanced} and \multinli \cite{williams2017broad}, which are commonly used text datasets in subpopulation shift.
Finally, we curate 4 datasets in the \textbf{medical} domain. We construct \mimiccxr \cite{johnson2019mimic} and \chexpert \cite{irvin2019chexpert} to predict the presence of any pathology from a chest X-ray. We also construct \mimicnotes for mortality classification from clinical notes \cite{chen2019can}. Finally, we follow a recent work in evaluating subgroup shift and construct the \cxrmultisite dataset \cite{puli2021out}.
\tabref{table:dataset-overview} reports the details of each dataset.
We leave full information and descriptions for each of the datasets in Appendix \ref{appendix-subsec:dataset-details}.

\textbf{Algorithms.}
We evaluate \textbf{20} algorithms that span a broad range of learning strategies and categories, and relate their performance to different shifts defined in our framework. We believe this is the first work to comprehensively evaluate a large set of diverse algorithms in subpopulation shift. Concretely, these algorithms cover the following areas:
(1) \emph{vanilla:} \textbf{ERM} \cite{vapnik1999overview}, 
(2) \emph{subgroup robust methods:} \textbf{GroupDRO} \cite{sagawa2020groupdro}, \textbf{CVaRDRO} \cite{duchi2018learning}, \textbf{LfF} \cite{nam2020learning}, \textbf{JTT} \cite{liu2021just}, \textbf{LISA} \cite{yao2022improving}, \textbf{DFR} \cite{izmailov2022feature},
(3) \emph{data augmentation:} \textbf{Mixup}~\cite{zhang2018mixup}, 
(4) \emph{domain-invariant feature learning:} \textbf{IRM} \cite{arjovsky2019irm}, \textbf{CORAL} \cite{sun2016coral}, \textbf{MMD} \cite{li2018mmd}, 
(5) \emph{imbalanced learning:} \textbf{ReSample} \cite{japkowicz2000class}, \textbf{ReWeight} \cite{japkowicz2000class}, \textbf{Focal} \cite{lin2017focal}, \textbf{CBLoss} \cite{cui2019class}, \textbf{LDAM} \cite{cao2019learning}, \textbf{BSoftmax} \cite{ren2020bsoftmax}, \textbf{CRT} \cite{kang2020decoupling}, \textbf{ReWeightCRT} \cite{kang2020decoupling}.
Our framework can be easily extended to include new algorithms.
We provide detailed descriptions for each algorithm in Appendix~\ref{appendix-subsec:algo-details}.

\textbf{Evaluation Metrics.}
Existing works on subpopulation shift mainly report \emph{worst-group accuracy} (WGA) as the gold-standard. While WGA faithfully assesses worst-group performance, other important metrics (e.g., worst-case precision, calibration error, etc.) are also essential especially when involving subpopulation shift. Therefore, in our benchmark we include a variety of metrics aiming for a thorough evaluation from different aspects.
In particular, besides \textbf{Avg Accuracy} and \textbf{Worst Accuracy}, we further include \textbf{Avg Precision}, \textbf{Worst Precision}, \textbf{Avg F1-score}, \textbf{Worst F1-score}, (Class-)\textbf{Balanced Accuracy}, \textbf{Adjusted Accuracy} (accuracy on a \emph{group}-balanced dataset), and expected calibration error (\textbf{ECE}) \cite{guo2017calibration}.
Detailed summaries of all metrics are in Appendix \ref{appendix-subsec:eval-metrics}.

\begin{figure*}[!t]
\begin{center}
\includegraphics[width=\linewidth]{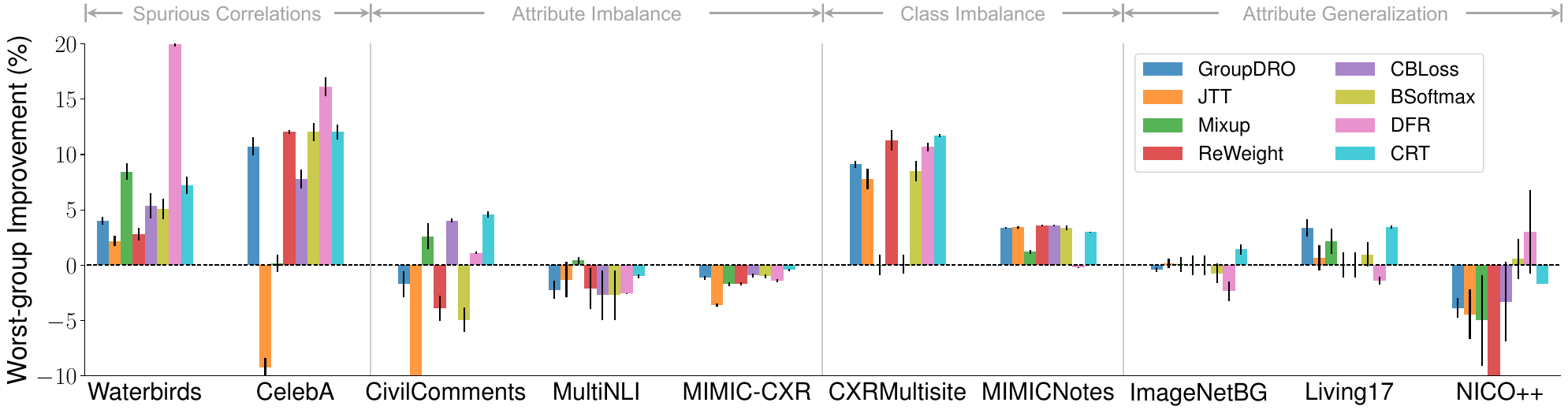}
\end{center}
\vspace{-0.5cm}
\caption{Worst-group improvements over ERM across different datasets when attributes are \emph{unknown} in both training and validation set. SOTA algorithms only enhance subgroup robustness on certain types of shift (i.e., \textbf{SC} and \textbf{CI}). Complete results are in Appendix \ref{appendix-subsec:improvements-diff-settings}.}
\label{fig:improve-over-erm-valattrNo}
\vspace{-0.3cm}
\end{figure*}

\begin{figure}[!t]
\begin{center}
\includegraphics[width=\linewidth]{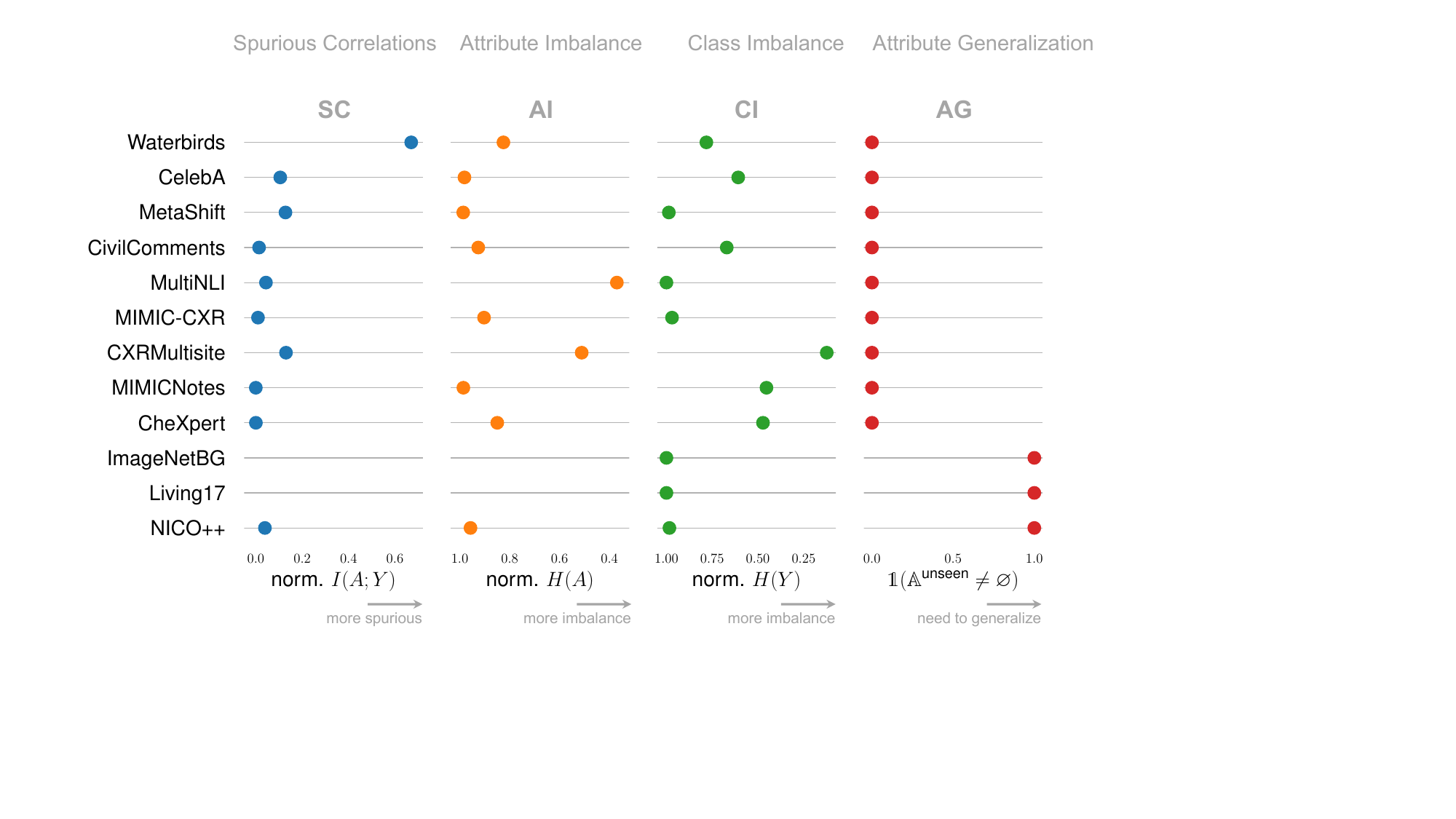}
\end{center}
\vspace{-0.5cm}
\caption{Quantification of the degree of different shifts over all datasets. Additional metrics are provided in Appendix \ref{appendix-subsec:quantify-shift}.}
\label{fig:quantify-shift}
\vspace{-0.5cm}
\end{figure}

\textbf{Attribute Availability.}
Whether attribute is known in both (1) \emph{training set} and (2) \emph{validation set} has long been a vital factor for almost all subgroup algorithms \cite{izmailov2022feature}. Specifically, classic methods (e.g., GroupDRO) assume access to attributes during training to define meaningful groups. Recently, a number of methods (e.g., JTT, LfF, DFR) try to improve worst-group accuracy without knowing the training attributes.
Nevertheless, current approaches still require access to group-annotated validation set for model selection and hyperparameter tuning \cite{idrissi2022simple}.

We systematically investigate this phenomenon by considering three settings in our benchmark: (1) \emph{attributes are known in both training \& validation}, (2) \emph{attributes are unknown in training, but known in validation}, and (3) \emph{attributes are unknown in both training \& validation}. Note that when training attributes are unknown, methods that operate over \emph{subgroups} degenerate to operate over \emph{classes}.
Without further specification, we report results under the third setting, which is the hardest but the most realistic one. We include full results across all settings in Appendix \ref{appendix-sec:complete-results}.

\textbf{Model Selection.}
As mentioned earlier, model selection becomes essential when attributes are completely unknown. Significant drop (over 20\%) in worst-group test accuracy has been observed if using the highest \emph{average} validation accuracy as the model selection criterion without any group annotations \cite{idrissi2022simple}. To this end, we provide a rigorous analysis on different model selection strategies, especially when attributes are fully unknown. Further details are provided in Appendix \ref{appendix-subsec:model-select-attr-availability}.

\textbf{Implementation.}
For a fair evaluation, following \cite{gulrajani2020domainbed}, for each algorithm we conduct a random search of 16 trials over a joint distribution of all hyperparameters (details are provided in Appendix \ref{appendix-sec:exp-setting}). We then use the validation set to select the best hyperparameters for each algorithm, fix them and rerun the experiments under three different random seeds to report the final average results with standard deviation. Such process ensures the comparison is best-versus-best, and the hyperparameters are optimized for all algorithms.

\section{A Fine-Grained Analysis}
\label{sec:analysis}
\begin{table*}[!t]
\setlength{\tabcolsep}{5pt}
\caption{Results on all tested subpopulation benchmarks, when attributes are \emph{unknown} in both training and validation set. Full results for each dataset and other settings are in Appendix~\ref{appendix-sec:complete-results}. Methods that re-train classifier using a two-stage strategy are marked in \colorbox{baselinecolor}{gray}.}
\vspace{-9pt}
\label{table:main-full-results}
\small
\begin{center}
\adjustbox{max width=\textwidth}{%
\begin{tabular}{lcccccccccccc|c}
\toprule[1.5pt]
\textbf{Algorithm}         & \waterbirds        & \celeba            & \civilcomments & \multinli          & \metashift         & \imagenetbg        & \nicopp            & \mimiccxr    & \mimicnotes        & \cxrmultisite      & \chexpert & \living          & \textbf{Avg}               \\
\midrule
ERM                        & 69.1 \scriptsize$\pm4.7$   & 57.6 \scriptsize$\pm0.8$   & 63.2 \scriptsize$\pm1.2$   & 66.4 \scriptsize$\pm2.3$   & 82.1 \scriptsize$\pm0.8$   & 76.8 \scriptsize$\pm0.9$   & 35.0 \scriptsize$\pm4.1$   & 68.6 \scriptsize$\pm0.2$   & 80.4 \scriptsize$\pm0.2$   & 50.1 \scriptsize$\pm0.9$   & 41.7 \scriptsize$\pm3.4$   & 27.7 \scriptsize$\pm1.1$   & 59.9                       \\[1.2pt]
Mixup                      & 77.5 \scriptsize$\pm0.7$   & 57.8 \scriptsize$\pm0.8$   & 65.8 \scriptsize$\pm1.5$   & 66.8 \scriptsize$\pm0.3$   & 79.0 \scriptsize$\pm0.8$   & 76.9 \scriptsize$\pm0.7$   & 30.0 \scriptsize$\pm4.1$   & 66.8 \scriptsize$\pm0.6$   & 81.6 \scriptsize$\pm0.6$   & 50.1 \scriptsize$\pm0.9$   & 37.4 \scriptsize$\pm3.5$   & 29.8 \scriptsize$\pm1.8$   & 60.0                       \\[1.2pt]
GroupDRO                   & 73.1 \scriptsize$\pm0.4$   & 68.3 \scriptsize$\pm0.9$   & 61.5 \scriptsize$\pm1.8$   & 64.1 \scriptsize$\pm0.8$   & 83.1 \scriptsize$\pm0.7$   & 76.4 \scriptsize$\pm0.2$   & 31.1 \scriptsize$\pm0.9$   & 67.4 \scriptsize$\pm0.5$   & 83.7 \scriptsize$\pm0.1$   & 59.2 \scriptsize$\pm0.3$   & 74.7 \scriptsize$\pm0.3$   & 31.1 \scriptsize$\pm1.0$   & 64.5                       \\[1.2pt]
CVaRDRO                    & 75.5 \scriptsize$\pm2.2$   & 60.2 \scriptsize$\pm3.0$   & 62.9 \scriptsize$\pm3.8$   & 48.2 \scriptsize$\pm3.4$   & 83.5 \scriptsize$\pm0.5$   & 74.8 \scriptsize$\pm0.8$   & 27.8 \scriptsize$\pm2.3$   & 68.0 \scriptsize$\pm0.2$   & 65.6 \scriptsize$\pm1.5$   & 50.2 \scriptsize$\pm0.9$   & 50.2 \scriptsize$\pm1.8$   & 27.3 \scriptsize$\pm1.6$   & 57.8                       \\[1.2pt]
JTT                        & 71.2 \scriptsize$\pm0.5$   & 48.3 \scriptsize$\pm1.5$   & 51.0 \scriptsize$\pm4.2$   & 65.1 \scriptsize$\pm1.6$   & 82.6 \scriptsize$\pm0.4$   & 77.0 \scriptsize$\pm0.4$   & 30.6 \scriptsize$\pm2.3$   & 64.9 \scriptsize$\pm0.3$   & 83.8 \scriptsize$\pm0.1$   & 57.9 \scriptsize$\pm2.1$   & 60.4 \scriptsize$\pm4.8$   & 28.3 \scriptsize$\pm1.1$   & 60.1                       \\[1.2pt]
LfF                        & 75.0 \scriptsize$\pm0.7$   & 53.0 \scriptsize$\pm4.3$   & 42.2 \scriptsize$\pm7.2$   & 57.3 \scriptsize$\pm5.7$   & 72.3 \scriptsize$\pm1.3$   & 70.1 \scriptsize$\pm1.4$   & 28.8 \scriptsize$\pm2.0$   & 62.2 \scriptsize$\pm2.4$   & 84.0 \scriptsize$\pm0.1$   & 50.1 \scriptsize$\pm0.9$   & 13.7 \scriptsize$\pm9.8$   & 26.4 \scriptsize$\pm1.3$   & 52.9                       \\[1.2pt]
LISA                       & 77.5 \scriptsize$\pm0.7$   & 57.8 \scriptsize$\pm0.8$   & 65.8 \scriptsize$\pm1.5$   & 66.8 \scriptsize$\pm0.3$   & 79.0 \scriptsize$\pm0.8$   & 76.9 \scriptsize$\pm0.7$   & 30.0 \scriptsize$\pm4.1$   & 66.8 \scriptsize$\pm0.6$   & 81.6 \scriptsize$\pm0.6$   & 50.1 \scriptsize$\pm0.9$   & 37.4 \scriptsize$\pm3.5$   & 29.8 \scriptsize$\pm1.8$   & 60.0                       \\[1.2pt]
ReSample                  & 70.0 \scriptsize$\pm1.0$  & 74.1 \scriptsize$\pm2.2$  & 61.0 \scriptsize$\pm0.6$  & 66.8 \scriptsize$\pm0.5$  & 81.0 \scriptsize$\pm1.7$  & 77.7 \scriptsize$\pm1.1$  & 30.6 \scriptsize$\pm2.3$  & 67.5 \scriptsize$\pm0.3$  & 82.6 \scriptsize$\pm0.6$  & 55.0 \scriptsize$\pm0.2$  & 74.3 \scriptsize$\pm0.4$  & 31.4 \scriptsize$\pm0.6$  & 64.3                      \\[1.2pt]
ReWeight                   & 71.9 \scriptsize$\pm0.6$   & 69.6 \scriptsize$\pm0.2$   & 59.3 \scriptsize$\pm1.1$   & 64.2 \scriptsize$\pm1.9$   & 83.1 \scriptsize$\pm0.7$   & 76.8 \scriptsize$\pm0.9$   & 25.0 \scriptsize$\pm0.0$   & 67.0 \scriptsize$\pm0.4$   & 84.0 \scriptsize$\pm0.1$   & 61.4 \scriptsize$\pm1.3$   & 73.7 \scriptsize$\pm1.0$   & 27.7 \scriptsize$\pm1.1$   & 63.6                       \\[1.2pt]
SqrtReWeight               & 71.0 \scriptsize$\pm1.4$   & 66.9 \scriptsize$\pm2.2$   & 68.6 \scriptsize$\pm1.1$   & 63.8 \scriptsize$\pm2.4$   & 82.6 \scriptsize$\pm0.4$   & 76.8 \scriptsize$\pm0.9$   & 32.8 \scriptsize$\pm3.5$   & 68.0 \scriptsize$\pm0.4$   & 83.1 \scriptsize$\pm0.2$   & 61.2 \scriptsize$\pm0.6$   & 68.5 \scriptsize$\pm1.6$   & 27.7 \scriptsize$\pm1.1$   & 64.2                       \\[1.2pt]
CBLoss                     & 74.4 \scriptsize$\pm1.2$   & 65.4 \scriptsize$\pm1.4$   & 67.3 \scriptsize$\pm0.2$   & 63.6 \scriptsize$\pm2.4$   & 83.1 \scriptsize$\pm0.0$   & 76.8 \scriptsize$\pm0.9$   & 31.7 \scriptsize$\pm3.6$   & 67.6 \scriptsize$\pm0.3$   & 84.0 \scriptsize$\pm0.1$   & 50.2 \scriptsize$\pm0.9$   & 74.0 \scriptsize$\pm0.7$   & 27.7 \scriptsize$\pm1.1$   & 63.8                       \\[1.2pt]
Focal                      & 71.6 \scriptsize$\pm0.8$   & 56.9 \scriptsize$\pm3.4$   & 61.9 \scriptsize$\pm1.1$   & 62.4 \scriptsize$\pm2.0$   & 81.0 \scriptsize$\pm0.4$   & 71.9 \scriptsize$\pm1.2$   & 30.6 \scriptsize$\pm2.3$   & 68.7 \scriptsize$\pm0.4$   & 70.9 \scriptsize$\pm9.8$   & 50.0 \scriptsize$\pm0.9$   & 42.1 \scriptsize$\pm4.0$   & 26.9 \scriptsize$\pm0.6$   & 57.9                       \\[1.2pt]
LDAM                       & 70.9 \scriptsize$\pm1.7$   & 57.0 \scriptsize$\pm4.1$   & 28.4 \scriptsize$\pm7.7$   & 65.5 \scriptsize$\pm0.8$   & 83.6 \scriptsize$\pm0.4$   & 76.7 \scriptsize$\pm0.5$   & 31.7 \scriptsize$\pm3.6$   & 66.6 \scriptsize$\pm0.6$   & 81.0 \scriptsize$\pm0.3$   & 50.1 \scriptsize$\pm0.9$   & 36.0 \scriptsize$\pm0.7$   & 24.3 \scriptsize$\pm0.8$   & 56.0                       \\[1.2pt]
BSoftmax                   & 74.1 \scriptsize$\pm0.9$   & 69.6 \scriptsize$\pm1.2$   & 58.3 \scriptsize$\pm1.1$   & 63.6 \scriptsize$\pm2.4$   & 82.6 \scriptsize$\pm0.4$   & 76.1 \scriptsize$\pm2.0$   & 35.6 \scriptsize$\pm1.8$   & 67.6 \scriptsize$\pm0.6$   & 83.8 \scriptsize$\pm0.3$   & 58.6 \scriptsize$\pm1.8$   & 73.8 \scriptsize$\pm1.0$   & 28.6 \scriptsize$\pm1.4$   & 64.4                       \\[1.2pt]
\grayrow
DFR                        & 89.0 \scriptsize$\pm0.2$   & 73.7 \scriptsize$\pm0.8$   & 64.4 \scriptsize$\pm0.1$   & 63.8 \scriptsize$\pm0.0$   & 81.4 \scriptsize$\pm0.1$   & 74.4 \scriptsize$\pm1.8$   & 38.0 \scriptsize$\pm3.8$   & 67.1 \scriptsize$\pm0.4$   & 80.2 \scriptsize$\pm0.0$   & 60.8 \scriptsize$\pm0.4$   & 75.8 \scriptsize$\pm0.3$   & 26.3 \scriptsize$\pm0.4$   & 66.2                       \\[1.2pt]
\grayrow
CRT                        & 76.3 \scriptsize$\pm0.8$   & 69.6 \scriptsize$\pm0.7$   & 67.8 \scriptsize$\pm0.3$   & 65.4 \scriptsize$\pm0.2$   & 83.1 \scriptsize$\pm0.0$   & 78.2 \scriptsize$\pm0.5$   & 33.3 \scriptsize$\pm0.0$   & 68.1 \scriptsize$\pm0.1$   & 83.4 \scriptsize$\pm0.0$   & 61.8 \scriptsize$\pm0.1$   & 74.6 \scriptsize$\pm0.4$   & 31.1 \scriptsize$\pm0.1$   & 66.1                       \\[1.2pt]
\grayrow
ReWeightCRT                & 76.3 \scriptsize$\pm0.2$   & 70.7 \scriptsize$\pm0.6$   & 64.7 \scriptsize$\pm0.2$   & 65.2 \scriptsize$\pm0.2$   & 85.1 \scriptsize$\pm0.4$   & 77.5 \scriptsize$\pm0.7$   & 33.3 \scriptsize$\pm0.0$   & 67.9 \scriptsize$\pm0.1$   & 83.4 \scriptsize$\pm0.0$   & 53.1 \scriptsize$\pm2.3$   & 75.1 \scriptsize$\pm0.2$   & 33.1 \scriptsize$\pm0.1$   & 65.4                       \\
\bottomrule[1.5pt]
\end{tabular}}
\end{center}
\vspace{-0.1cm}
\end{table*}

\subsection{Quantifying Subpopulation Shift}
\label{subsec:quantify-shifts}

In order to quantify the degree of each shift for each dataset relative to others, we use several simple metrics. For \emph{spurious correlations}, we use the normalized mutual information between $A$ and $Y$, where $\text{norm } I(A; Y) = 1$ means that the two are perfectly correlated: $\text{norm } I(A; Y) = \frac{2 I(A; Y)}{H(Y) + H(A)}$.

For \emph{attribute} and \emph{class imbalance}, we use the normalized entropy, where $\text{norm } H(Y) = 1$ indicates that the distribution is uniform (i.e., no imbalance): $\text{norm } H(Y) = \frac{H(Y)}{\log |\text{supp}(Y)|}$.

For \emph{attribute generalization}, we simply examine whether there exist any subpopulations in the test set which do not appear during training via an indicator function (see \figref{fig:quantify-shift}).
We provide several additional metrics in Appendix \ref{appendix-subsec:quantify-shift}.

We find that different datasets exhibit very different types of shift, and the degrees also greatly vary (\figref{fig:quantify-shift}). To further study how algorithms perform across various types of shift, we categorize each dataset into its most dominant shift type.

\subsection{Performance across Different Types of Shift}
\vspace{-0.03cm}
\label{subsec:main-perf-across-shifts}

As described earlier, we run experiments for all algorithms, datasets, and attribute availability settings. We use \emph{worst-group accuracy} as the model selection criterion, and provide analysis for other metrics in Appendix \ref{appendix-subsec:model-selection-metrics}.
When attributes are unknown in the validation set, this criterion degenerates to \emph{worst-class accuracy}. Interestingly, we discover that this simple method is surprisingly effective (related results in Sec.~\ref{subsec:model-select-attr-availability}).
In total, we trained over 10,000 models.

We study model performance over different shifts. Specifically, we report results when attributes are \emph{unknown} in both training and validation. Results for other settings are in Appendix \ref{appendix-subsec:improvements-diff-settings}. We present main results in \figref{fig:improve-over-erm-valattrNo} and \tabref{table:main-full-results}, where we make intriguing observations as follows.

\begin{figure*}[h]
\begin{center}
\includegraphics[width=\linewidth]{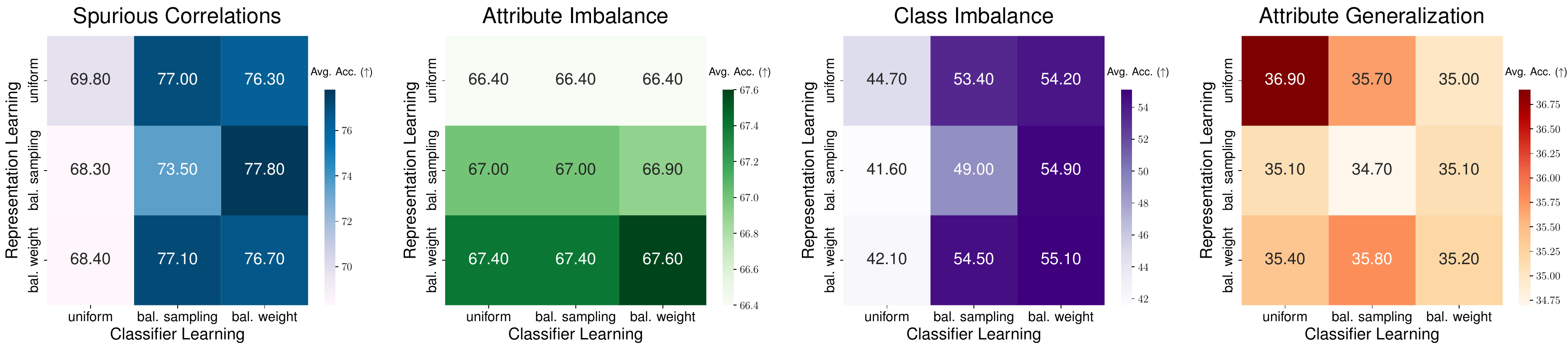}
\end{center}
\vspace{-0.5cm}
\caption{Averaged worst-group accuracy of different manners for representation learning and classifier learning under different shifts. Within each shift type, we average the results across datasets that belong to this shift to report the final accuracy. As observed, balanced classifier learning substantially improves the results for \textbf{SC} and \textbf{CI}, while balanced representation learning gives reasonable gains for \textbf{AI}; Yet, no stratified learning manners lead to performance gains under \textbf{AG} compared to vanilla ERM. Experimental details are in \secref{subsec:rep-vs-vls}.}
\label{fig:rep-cls}
\vspace{-0.4cm}
\end{figure*}

\begin{table}[!t]
\setlength{\tabcolsep}{8pt}
\caption{Relative improvements over ERM when using stratified balanced representation or classifier learning under different shifts.}
\label{table:rep-cls-relative-improves}
\small
\begin{center}
\adjustbox{max width=0.4\textwidth}{
\begin{tabular}{lcccc}
\toprule[1.5pt]
 & \textbf{SC} & \textbf{AI} & \textbf{CI} & \textbf{AG} \\ \midrule
\textsc{Representation} & \textcolor{lightblue}{\texttt{-}\textbf{0.3}} & \textcolor{darkgreen}{\texttt{+}\textbf{1.1}} & \textcolor{lightblue}{\texttt{-}\textbf{0.2}} & \textcolor{lightblue}{\texttt{-}\textbf{0.4}} \\[1.2pt]
\textsc{Classifier} & \textcolor{darkgreen}{\texttt{+}\textbf{8.1}} & \textcolor{deemph}{\texttt{+}\textbf{0.0}} & \textcolor{darkgreen}{\texttt{+}\textbf{11.9}} &\textcolor{lightblue}{\texttt{-}\textbf{0.4}} \\
\bottomrule[1.5pt]
\end{tabular}
}
\end{center}
\vspace{-0.5cm}
\end{table}

\textbf{SOTA algorithms only improve subgroup robustness on certain types of shift, but not others.}
As \figref{fig:improve-over-erm-valattrNo} illustrates, for \emph{spurious correlations} and \emph{class imbalance}, existing algorithms can provide consistent worst-group gains over ERM even in the absence of validation attributes, indicating that progress has been made for tackling these two specific shifts. Interestingly however, when it comes to \emph{attribute imbalance}, little improvement is observed across datasets. In addition, the performance becomes even worse for \emph{attribute generalization}. These findings stress that current advances are only made for specific shifts (i.e., SC and CI), while no progress has been made for the more challenging shifts such as AG.

\textbf{Methods that decouple representation and classifier are more effective.}
When further zoom into the performance across all datasets in \tabref{table:main-full-results}, a set of methods that decouple the training of representation and classifier \cite{izmailov2022feature, kang2020decoupling} achieve remarkable gains over all other algorithms (highlighted in \textcolor{deemph}{gray}).
As prior works also confirmed \cite{izmailov2022feature}, features learned by ERM seem to be good enough under spurious correlations.
These findings inspire us to further understand the role of \emph{representation} and \emph{classifier} in subpopulation shift, especially their behaviors under different subgroup shifts.

\begin{table*}[!t]
\setlength{\tabcolsep}{8pt}
\caption{Test-set worst-group accuracy difference (\%) between each selection strategy on each dataset, relative to the oracle which selects the best worst-group accuracy. Complete results across all datasets and all selection strategies are provided in Appendix \ref{appendix-subsec:model-selection-metrics}.}
\vspace{-10pt}
\label{table:selection_criteria}
\small
\begin{center}
\resizebox{1\textwidth}{!}{
\begin{tabular}{lrrrrrr|r}
\toprule[1.5pt]
\textbf{Selection Strategy}     & \celeba     & \chexpert & \civilcomments & \mimiccxr & \mimicnotes & \metashift     & \textbf{Avg}   \\ \midrule
Max Worst-Class Accuracy     & -5.0 \scriptsize$\pm6.3$   & \textbf{-0.4} \scriptsize$\pm0.8$          & \textbf{-3.2} \scriptsize$\pm5.2$          & \textbf{-0.9} \scriptsize$\pm1.0$       & \textbf{-0.1} \scriptsize$\pm0.5$   & \textbf{-1.5} \scriptsize$\pm3.0$   & \textbf{-1.8}  \\
Max Balanced Accuracy & \textbf{-4.4} \scriptsize$\pm5.4$   & -1.3 \scriptsize$\pm2.5$          & -3.5 \scriptsize$\pm5.8$          & -2.9 \scriptsize$\pm4.9$       & -2.3 \scriptsize$\pm6.2$   & -1.7 \scriptsize$\pm3.0$    & -2.7  \\
Min Class Accuracy Diff   & -6.1 \scriptsize$\pm9.1$  & -1.9 \scriptsize$\pm5.3$          & -4.1 \scriptsize$\pm8.0$          & -1.9 \scriptsize$\pm5.0$       & -0.3 \scriptsize$\pm1.2$   & -2.2 \scriptsize$\pm4.6$  &  -2.7  \\
Max Worst-Class F1   & -13.4 \scriptsize$\pm10.4$ & -5.4 \scriptsize$\pm6.7$          & \textbf{-3.2} \scriptsize$\pm3.8$          & -2.5 \scriptsize$\pm2.2$       & -4.4 \scriptsize$\pm8.7$   & -1.8 \scriptsize$\pm3.3$    & -5.1  \\
Max Overall AUROC        & -12.2 \scriptsize$\pm10.3$ & -10.4 \scriptsize$\pm13.0$        & -8.2 \scriptsize$\pm9.0$          & -6.6 \scriptsize$\pm9.9$       & -10.0 \scriptsize$\pm16.5$ & -3.2 \scriptsize$\pm7.0$  &  -8.4  \\
Max Overall Accuracy      & -18.6 \scriptsize$\pm12.0$ & -30.9 \scriptsize$\pm24.9$        & -13.7 \scriptsize$\pm9.5$         & -5.1 \scriptsize$\pm6.3$       & -19.9 \scriptsize$\pm26.0$ & -1.9 \scriptsize$\pm3.3$    & -15.0 \\
\bottomrule[1.5pt]
\end{tabular}
}
\end{center}
\end{table*}

\begin{figure*}[!t]
\begin{center}
\includegraphics[width=0.9\linewidth]{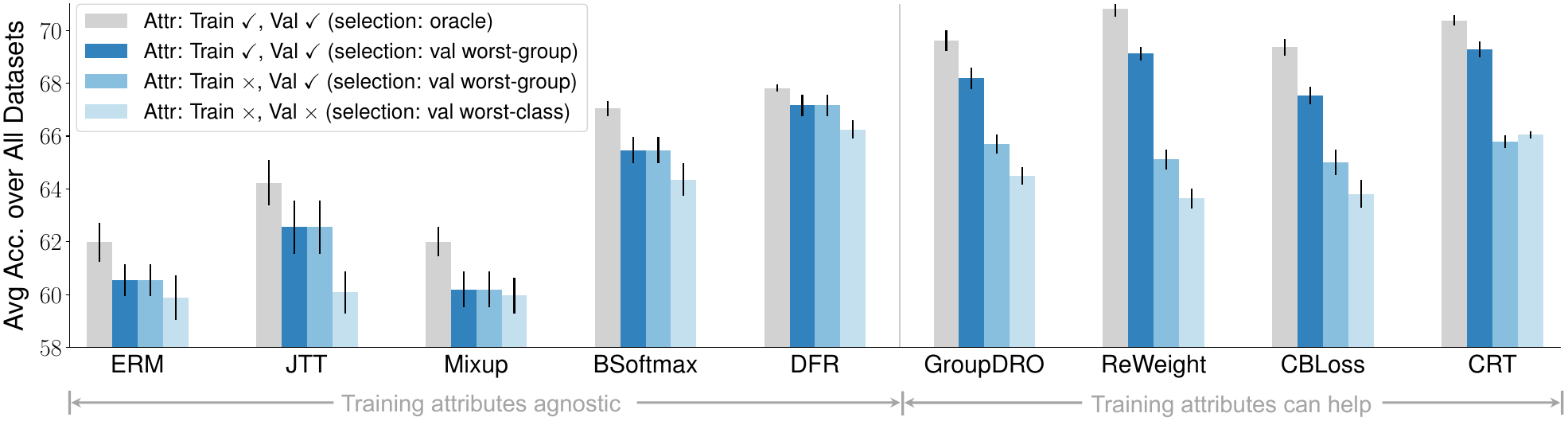}
\end{center}
\vspace{-0.5cm}
\caption{Averaged worst-group accuracy of various algorithms under different model selection and attribute availability settings.}
\label{fig:diff-selection-settings}
\vspace{-0.2cm}
\end{figure*}

\vspace{-0.01cm}
\subsection{The Role of Representation and Classifier}
\label{subsec:rep-vs-vls}

We are motivated to explore the role of representation and classifier in subpopulation shift. 
In particular, we separate the whole network into two parts: the feature extractor and the classifier.
We then employ three training strategies for representation and classifier learning, respectively: (1) \emph{uniform}, which follows the normal ERM training; (2) \emph{balanced sampling}, where balanced samples are drawn from each group (class if attribute not available) during training, and (3) \emph{re-weighting}, where we re-weight all the samples by the inverse of the sample size of their groups (classes). Note that classifier re-balancing resembles CRT \cite{kang2020decoupling} and DFR \cite{izmailov2022feature}.
We train models following the above settings across all datasets, and average the results over datasets according to the type of shift.

\textbf{Representation \& classifier quality play different roles under different shifts.}
As \figref{fig:rep-cls} reveals, for \textbf{SC} and \textbf{CI}, balanced classifier learning (i.e., both re-sampling and re-weighting) can substantially improve the performance when fixing the representation, whereas different representation learning schemes do not lead to notable gains when fixing the classifier learning manner. Interestingly, for \textbf{AI}, balancing the classifier does not lead to better performance, while balanced representation schemes can bring notable gains.

\textbf{ERM features are not sufficient for subpopulation shift.}
Unlike recent works that claim ERM features are sufficient for out-of-distribution generalization \cite{rosenfeld2022domain, izmailov2022feature}, our above intriguing findings suggest that features learned via ERM may only be good enough for \textbf{certain} shifts. Concretely, improving the feature extractor still leads to notable gains especially for \textbf{AI}. The results in turn well explain the performance differences in \figref{fig:improve-over-erm-valattrNo}, that SOTA algorithms with two-stage training do not improve worst-case accuracy under \textbf{AI} or \textbf{AG}.

\textbf{Stratified balanced learning does not outperform ERM under AG.}
Finally, no stratified learning manners lead to performance gains under \textbf{AG}. As \tabref{table:rep-cls-relative-improves} summarizes, both stratified representation and classifier learning manners even exhibit negative gains for datasets that require \textbf{AG}. This reveals the intrinsic limitation of SOTA algorithms \cite{izmailov2022feature} against diverse types of subpopulation shift.

\begin{figure*}[h]
\begin{center}
\subfigure[\textbf{Accuracy on the line:} Adjusted accuracy is \emph{positively} correlated with WGA.]{
    \label{subfig:tradeoff-adjusted}
    \includegraphics[width=0.995\linewidth]{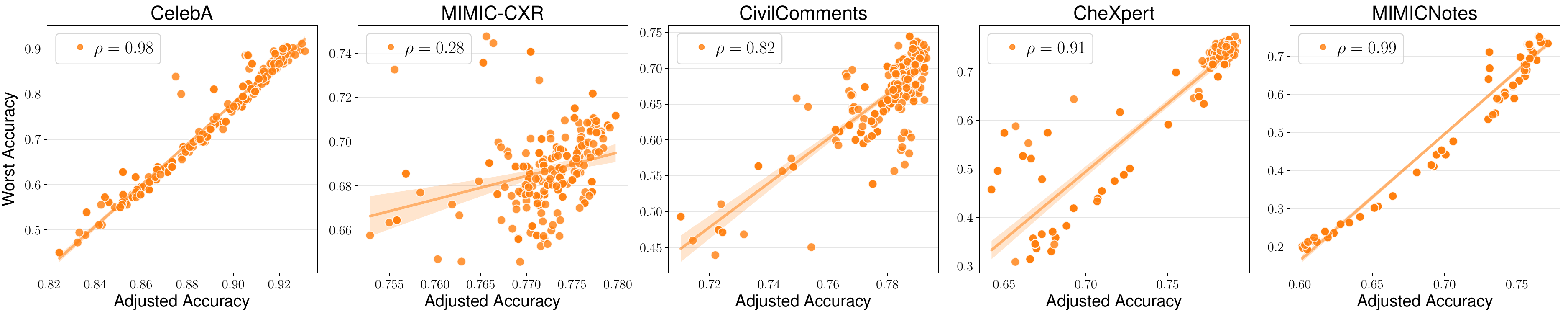}
}
\\
\subfigure[\textbf{Accuracy on the inverse line:} Worst-case precision is \emph{negatively} correlated with WGA.]{
    \label{subfig:tradeoff-worstprecision}
    \includegraphics[width=0.995\linewidth]{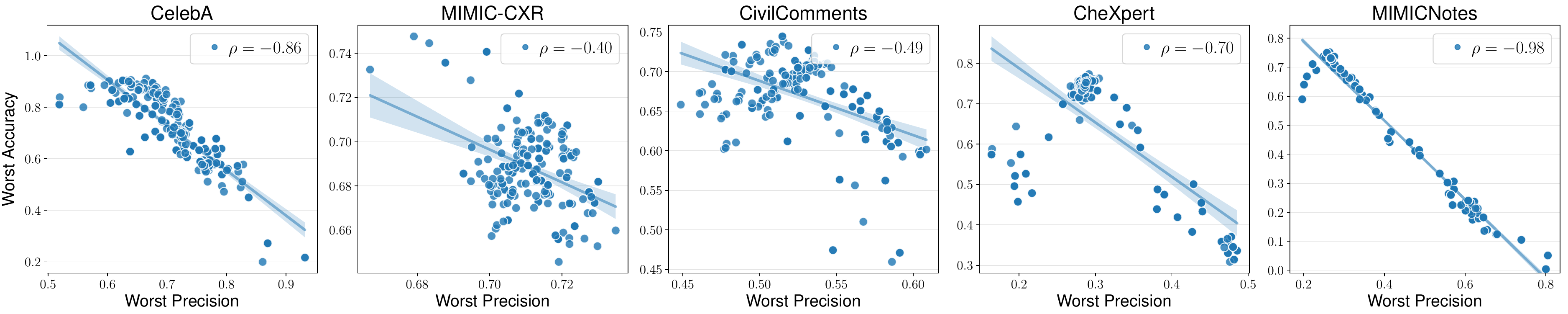}
}
\end{center}
\vspace{-0.5cm}
\caption{Fundamental tradeoff between WGA and other evaluation metrics. Complete results for all metrics are in Appendix \ref{appendix-subsec:tradeoff-metrics}.}
\label{fig:tradeoff-main-paper}
\vspace{-0.1cm}
\end{figure*}

\subsection{On Model Selection and Attribute Availability}
\label{subsec:model-select-attr-availability}

Model selection (e.g., choice of hyperparameters, training checkpoints) and attribute availability affect subpopulation shift evaluation considerably, especially given that almost all SOTA algorithms need access to a group-annotated validation set for model selection \cite{idrissi2022simple}.
We study this problem in-depth, where we follow three settings mentioned earlier (i.e., the availability of both \emph{training} and \emph{validation} attributes), and summarize the results in \figref{fig:diff-selection-settings}.

\textbf{The importance of training attribute availability relies on algorithm properties.}
As \figref{fig:diff-selection-settings} verifies, when training attribute is available, it can greatly boost the performance of algorithms that need group information (e.g., GroupDRO), while it does not bring benefits for attribute-agnostic methods (e.g., ERM, JTT).

\textbf{Validation attribute may not be necessary once you have a good selection metric.}
We further investigate the performance without validation attributes. It is widely known that SOTA subpopulation shift methods rely on group labels for validation. Surprisingly however, we observe a relatively small accuracy drop over all methods when using a simple \emph{worst-class accuracy} (degenerated from \emph{worst-group} as attributes are unknown in validation) as selection metric.
Specifically, comparing the last two bars across all methods in \figref{fig:diff-selection-settings}, the average accuracy drop is less than merely 2\%.
This striking finding contrasts with the literature, where large degradation (over 20\%) is observed when using \emph{average} accuracy as the metric without validation attributes. This suggests that if carefully choosing a metric for model selection, we can achieve minimal worst-group accuracy loss even in the absence of any attribute information.

\textbf{Simple selection criterion using worst-class accuracy is surprisingly effective even without validation attribute.}
We examine different strategies for choosing when to stop during model training when no attribute annotations are available in both training and validation.
We select six representative datasets and six representative selection strategies, respectively (full results across all datasets and all selection strategies are in Appendix \ref{appendix-subsec:model-selection-metrics}).
For each model, we utilize each stopping criterion over the validation set metrics computed throughout training, to determine its corresponding stopping point. We evaluate a variety of selection criteria in this way for a large variety of methods trained on each dataset. We compare each strategy with the oracle selection criteria, summarizing our results in \tabref{table:selection_criteria}.
We observe that simply stopping when the \emph{worst-class accuracy} reaches a maxima achieves the best worst-group accuracy on average. As expected, any selection criterion based on overall performance (e.g., accuracy, AUROC) performs much worse.

\subsection{Metrics Beyond Worst-Group Accuracy}
\label{subsec:metrics-beyond-wga}

Worst-group accuracy (WGA) has long been treated as the gold-standard for assessing the model performance in subpopulation shift. Recent studies also discovered that WGA and model average performance are linearly correlated, a phenomenon called ``\emph{Accuracy on the line}'' \cite{miller2021accuracy, izmailov2022feature}.
However, WGA essentially assesses the worst-case (top-1) recall conditioned on attribute \cite{yang2022multi}, which does not reflect other important metrics such as worst-case precision and calibration error.
Whether models with high WGA will also perform better across these metrics remains unknown.
Therefore, we further examine the relationship between WGA and other evaluation metrics that proposed in our benchmark.

\textbf{Intrinsic tradeoff: Accuracy can be on the inverse line.}
Interestingly, we observe that not all metrics are positively correlated with WGA. In particular, we show scatter plots of WGA \emph{vs.} other metrics for representative datasets.
As \figref{subfig:tradeoff-adjusted} confirms, adjusted accuracy is linearly correlated with WGA, which is well aligned with existing observations \cite{izmailov2022feature}.
Interestingly however, for worst-case precision, the \emph{positive} correlation does not hold anymore; instead, we observe a strong \emph{\textbf{negative}} linear correlation, indicating an intrinsic tradeoff between WGA and worst-case precision. We show in Appendix \ref{appendix-subsec:tradeoff-metrics} that many metrics also possess such ``\emph{accuracy on the \textbf{inverse} line}'' property, further verifying the inherent tradeoff between testing metrics.

\textbf{Fundamental limitations of WGA as the only metric.}
The above observations highlight the complex relationship between WGA and other metrics: Certain metrics display high positive correlation, while many others show the opposite case. This finding uncovers the fundamental limitation of using only WGA to assess model performance in subpopulation shift: A well performed model with high WGA can however have low worst-case precision, which is alarming especially in critical applications such as medical diagnosis (e.g., \chexpert).
Our observations emphasize the need for more realistic evaluation metrics in subpopulation shift.

\subsection{Further Analysis}
\label{subsec:further-analysis}

\textbf{Impact of model architecture (Appendix \ref{appendix-subsec:arch-pretrain}).}
We study the effect of different model architectures on subpopulation shift across various datasets and modalities. In particular, we employ ResNets and vision transformers (ViTs) for the image modality, and five different transformer-based language models for the text modality. We observe that on text datasets, base BERT models are already competitive over other architecture variants (Table \ref{tab-appendix:text_archs}). Yet, the results on image datasets are mixed when comparing the worst-group performance for ResNets and ViTs (Tables \ref{tab-appendix:image_archs_known} and \ref{tab-appendix:image_archs_unknown}).

\textbf{Impact of pretraining methods (Appendix \ref{appendix-subsec:arch-pretrain}).}
We investigate how different pretraining methods affect the model performance under subpopulation shift. We consider both \emph{supervised} and \emph{self-supervised} pretraining using various SOTA methods. Similar to previous findings \cite{izmailov2022feature}, we observe that supervised pretraining outperforms self-supervised counterparts for most of the experiments. The results may also suggest that better self-supervised schemes could be developed for tackling subgroup shifts.

\textbf{Impact of pretraining datasets (Appendix \ref{appendix-subsec:arch-pretrain}).}
Finally, we investigate whether increasing the pretraining dataset size could lead to better subgroup performance. We leverage ImageNet-21K \cite{ridnik2021imagenet} and SWAG \cite{singh2022revisiting} in addition to the default ImageNet-1K.
Interestingly, we find consistent and significant worst-group performance gains when going from ImageNet-1K to ImageNet-21K to SWAG, indicating that larger and more diverse pretraining datasets seem to increase worst-group performance.

\vspace{-0.15cm}
\section{Conclusion}
\label{sec:conclusion}
We systematically study the subpopulation shift problem, formalize a unified framework to define and quantify different types of subpopulation shift, and further set up a comprehensive benchmark for realistic evaluation.
Our benchmark includes 20 SOTA methods and 12 real-world datasets across different domains.
Based on over 10K trained models, we reveal several intriguing properties in subpopulation shift that have implications for future research, including divergent performance on different shifts, model selection criteria, and metrics to evaluate against.
We hope our benchmark and findings will promote realistic and rigorous evaluations and inspire new advances in subpopulation shift.

\section*{Acknowledgements}
This work was supported in part by the MIT-IBM Watson AI Lab, and a grant from Quanta Computing.

\bibliography{subpop}
\bibliographystyle{icml2023}

\newpage
\appendix
\onecolumn
\section{Limitations and Broader Impacts}
\label{appendix-sec:broader-impact-limitations}

\paragraph{Limitations.}
We acknowledge several limitations of our benchmark and analyses.
First, we have used 12 real-world predictive datasets in our benchmark. However, real-world data can have many complexities including potential mislabelling in both attributes and labels. We do not consider this effect, though it would be interesting to examine it in a synthetic setting.
Moreover, prior work has shown that in the case of multiple spurious attributes, reducing reliance on one can increase reliance on another \cite{li2022whac}. We only consider a single attribute in this benchmark, though an evaluation of this effect in the context of model selection criteria would be an interesting direction of future research.
 
\paragraph{Potential Negative Impacts.}
There are several potential negative social impacts of our work. First, we assume throughout the work that we would like to have models that are robust to subpopulation shift. However, in practice, this comes at the cost of overall accuracy on the training distribution. There may be cases where the practitioner would like to maximize overall accuracy regardless of spurious correlations, and thus subpopulation shift methods would worsen overall performance and potentially cause excess harm. Next, we recognize that the large grid of deep models trained for our evaluations likely resulted in a significant carbon footprint \cite{anthony2020carbontracker}. However, we hope that the insights provided in this paper will reduce the number of models and training steps (and therefore carbon emissions) required by future practitioners. Finally, we have constructed several models in this paper that utilize clinical data for clinical predictive tasks. We do not advocate for blind deployment of these models in any way, as there are many issues that need to be verified and resolved before their deployment, such as real-world clinical testing, privacy, fairness, interpretability, and regulatory requirements \cite{maleki2020machine, wiens2019no}.

\section{Details of the Subpopulation Shift Benchmark}
\label{appendix-sec:benchmark-details}

\subsection{Dataset Details}
\label{appendix-subsec:dataset-details}

We explore subpopulation shift using \textbf{12} real-world datasets from a variety of domains including computer vision, natural language processing, and healthcare applications. We provide example inputs for each dataset in \tabref{appendix:table:sample-inputs-images} and \tabref{appendix:table:sample-inputs-texts}. Note that we omit showing examples for \mimiccxr, \mimicnotes, and \cxrmultisite to comply with the \textit{PhysioNet Credentialed Health Data Use Agreement}.
Below, we provide detailed descriptions for each dataset in our benchmark.

\begin{table}[!t]
\setlength{\tabcolsep}{10pt}
\caption{Example inputs for \textbf{image datasets} in our benchmark. We omit showing samples for \mimiccxr and \cxrmultisite to comply with the \emph{PhysioNet Credentialed Health Data Use Agreement}.}
\label{appendix:table:sample-inputs-images}
\begin{center}
\begin{tabular}{lcccccc}
\toprule[1.5pt]
    \textbf{Dataset} & \multicolumn{6}{l}{\textbf{Examples}} \\
    \midrule\midrule
    \waterbirds &
        \raisebox{-.4\height}{\includegraphics[width=35pt, height=35pt]{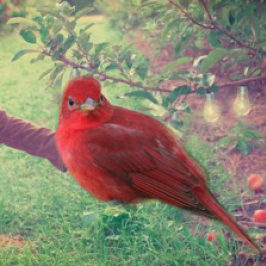}} &
        \raisebox{-.4\height}{\includegraphics[width=35pt, height=35pt]{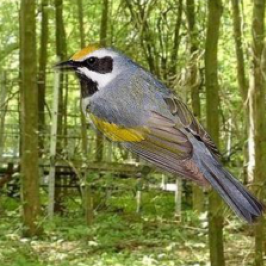}} &
        \raisebox{-.4\height}{\includegraphics[width=35pt, height=35pt]{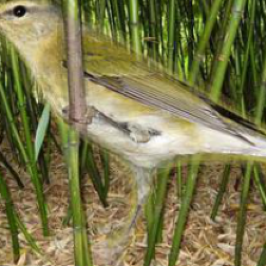}} &
        \raisebox{-.4\height}{\includegraphics[width=35pt, height=35pt]{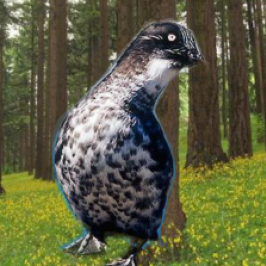}} &
        \raisebox{-.4\height}{\includegraphics[width=35pt, height=35pt]{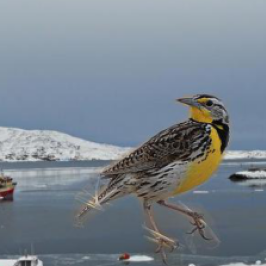}} &
        \raisebox{-.4\height}{\includegraphics[width=35pt, height=35pt]{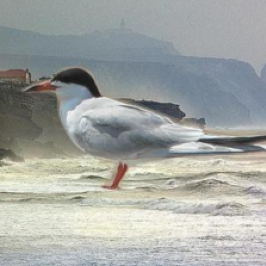}}
        \\
    \midrule
    \celeba &
        \raisebox{-.4\height}{\includegraphics[width=35pt, height=35pt]{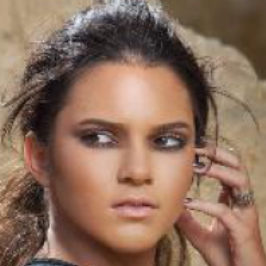}} &
        \raisebox{-.4\height}{\includegraphics[width=35pt, height=35pt]{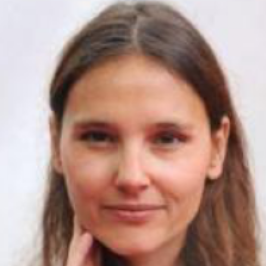}} &
        \raisebox{-.4\height}{\includegraphics[width=35pt, height=35pt]{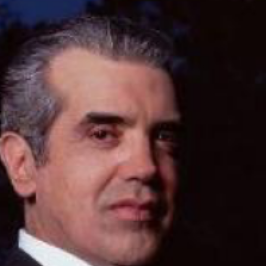}} &
        \raisebox{-.4\height}{\includegraphics[width=35pt, height=35pt]{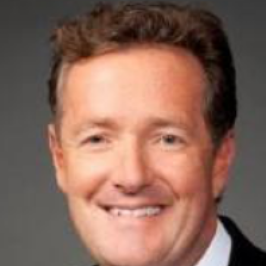}} &
        \raisebox{-.4\height}{\includegraphics[width=35pt, height=35pt]{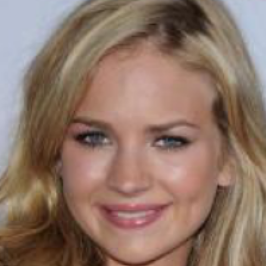}} &
        \raisebox{-.4\height}{\includegraphics[width=35pt, height=35pt]{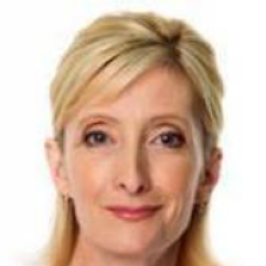}}
        \\
    \midrule
    \metashift &
        \raisebox{-.4\height}{\includegraphics[width=35pt, height=35pt]{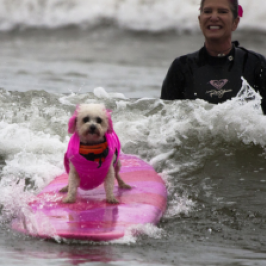}} &
        \raisebox{-.4\height}{\includegraphics[width=35pt, height=35pt]{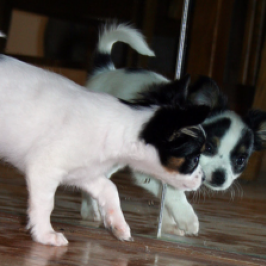}} &
        \raisebox{-.4\height}{\includegraphics[width=35pt, height=35pt]{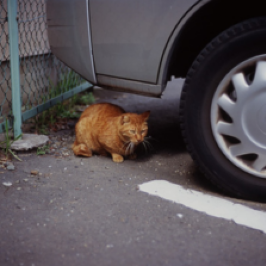}} &
        \raisebox{-.4\height}{\includegraphics[width=35pt, height=35pt]{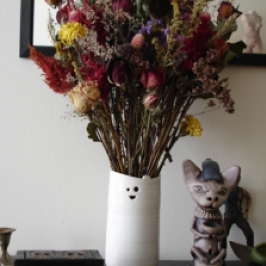}} &
        \raisebox{-.4\height}{\includegraphics[width=35pt, height=35pt]{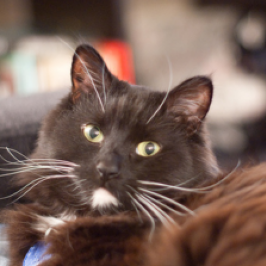}} &
        \raisebox{-.4\height}{\includegraphics[width=35pt, height=35pt]{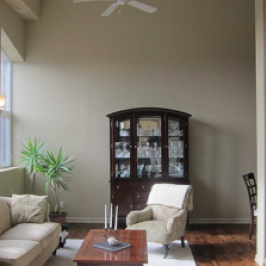}}
        \\
    \midrule
    \chexpert &
        \raisebox{-.4\height}{\includegraphics[width=35pt, height=35pt]{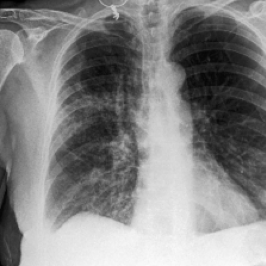}} &
        \raisebox{-.4\height}{\includegraphics[width=35pt, height=35pt]{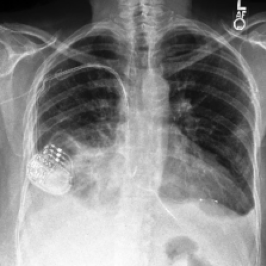}} &
        \raisebox{-.4\height}{\includegraphics[width=35pt, height=35pt]{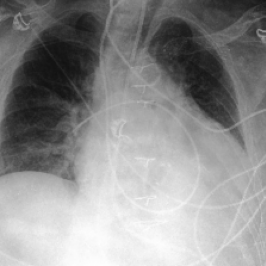}} &
        \raisebox{-.4\height}{\includegraphics[width=35pt, height=35pt]{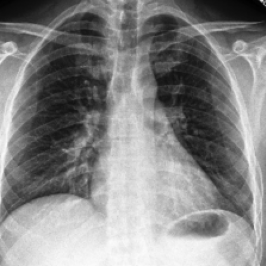}} &
        \raisebox{-.4\height}{\includegraphics[width=35pt, height=35pt]{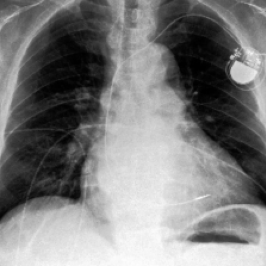}} &
        \raisebox{-.4\height}{\includegraphics[width=35pt, height=35pt]{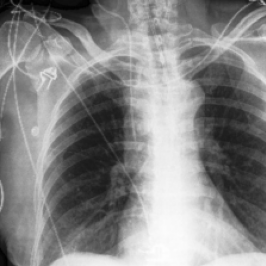}}
        \\
    \midrule
    \nicopp &
        \raisebox{-.4\height}{\includegraphics[width=35pt, height=35pt]{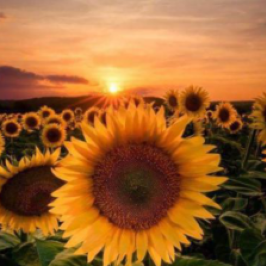}} &
        \raisebox{-.4\height}{\includegraphics[width=35pt, height=35pt]{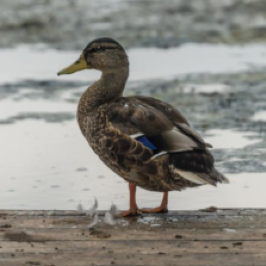}} &
        \raisebox{-.4\height}{\includegraphics[width=35pt, height=35pt]{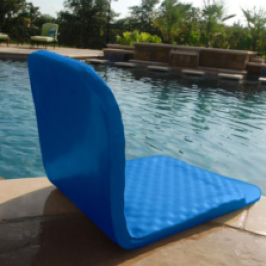}} &
        \raisebox{-.4\height}{\includegraphics[width=35pt, height=35pt]{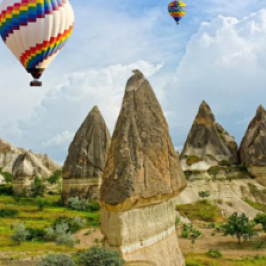}} &
        \raisebox{-.4\height}{\includegraphics[width=35pt, height=35pt]{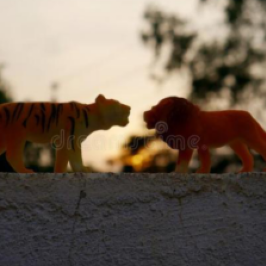}} &
        \raisebox{-.4\height}{\includegraphics[width=35pt, height=35pt]{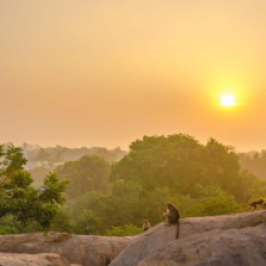}}
        \\
    \midrule
    \imagenetbg &
        \raisebox{-.4\height}{\includegraphics[width=35pt, height=35pt]{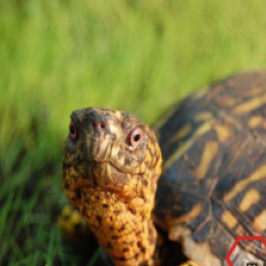}} &
        \raisebox{-.4\height}{\includegraphics[width=35pt, height=35pt]{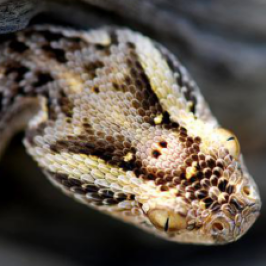}} &
        \raisebox{-.4\height}{\includegraphics[width=35pt, height=35pt]{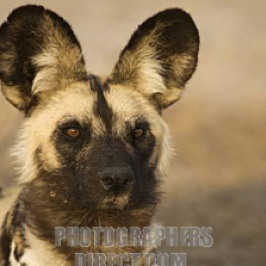}} &
        \raisebox{-.4\height}{\includegraphics[width=35pt, height=35pt]{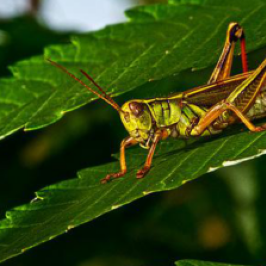}} &
        \raisebox{-.4\height}{\includegraphics[width=35pt, height=35pt]{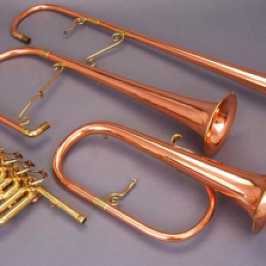}} &
        \raisebox{-.4\height}{\includegraphics[width=35pt, height=35pt]{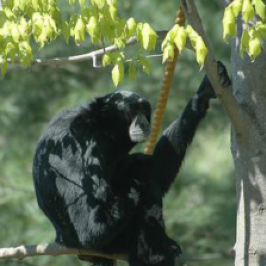}}
        \\
    \midrule
    \living &
        \raisebox{-.4\height}{\includegraphics[width=35pt, height=35pt]{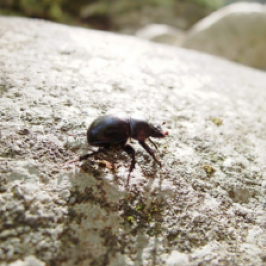}} &
        \raisebox{-.4\height}{\includegraphics[width=35pt, height=35pt]{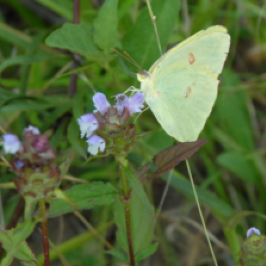}} &
        \raisebox{-.4\height}{\includegraphics[width=35pt, height=35pt]{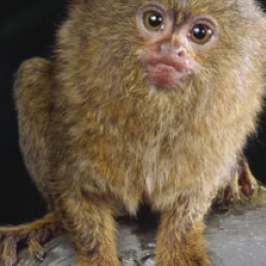}} &
        \raisebox{-.4\height}{\includegraphics[width=35pt, height=35pt]{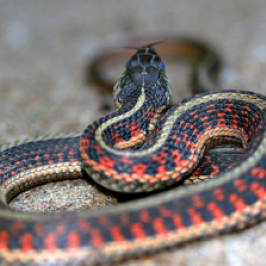}} &
        \raisebox{-.4\height}{\includegraphics[width=35pt, height=35pt]{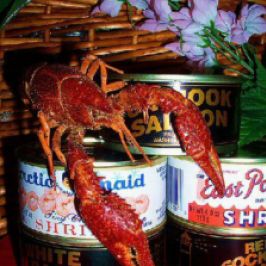}} &
        \raisebox{-.4\height}{\includegraphics[width=35pt, height=35pt]{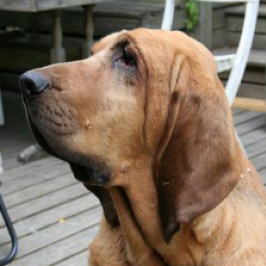}}
        \\
\bottomrule[1.5pt]
\end{tabular}
\end{center}
\vspace{-0.2cm}
\end{table}

\begin{table}[!t]
\setlength{\tabcolsep}{10pt}
\caption{Example inputs for \textbf{text datasets} in our benchmark. We omit showing samples for \mimicnotes to comply with the \emph{PhysioNet Credentialed Health Data Use Agreement}.}
\label{appendix:table:sample-inputs-texts}
\small
\begin{center}
\begin{tabular}{ll}
\toprule[1.5pt]
    \textbf{Dataset} & \textbf{Examples} \\
    \midrule\midrule
    \civilcomments & \begin{tabular}[c]{@{}l@{}}``Munchins looks like a munchins. The man who dont want to show his taxes, will tell you everything...''\\ ``The democratic party removed the filibuster to steamroll its agenda.  Suck it up boys and girls.''\\ ``so you dont use oil? no gasoline? no plastic? man you ignorant losers are pathetic.''\end{tabular} \\
    \midrule
    \multinli & \begin{tabular}[c]{@{}l@{}}``The analysis proves that there is no link between PM and bronchitis.'' \\ ``Postal Service were to reduce delivery frequency.''\\ ``The famous tenements (or lands) began to be built.''\end{tabular} \\
\bottomrule[1.5pt]
\end{tabular}
\end{center}
\vspace{-0.2cm}
\end{table}

\paragraph{\waterbirds \cite{wah2011caltech}.} \waterbirds is a commonly used binary classification image dataset in the spurious correlations setting, constructed by placing images from the Caltech-UCSD Birds-200-2011 (CUB) dataset \cite{wah2011caltech} over backgrounds from the Places dataset \cite{zhou2017places}. The task is to classify whether a bird is a landbird or a waterbird, where the spurious attribute is the background (water or land). We use standard train/val/test splits given by prior work \cite{idrissi2022simple}.

\paragraph{\celeba \cite{liu2015deep}.} \celeba is a binary classification image dataset consisting of over 200,000 celebrity face images. The task, which is also used widely in the spurious correlations literature, is to predict the hair color of the person (blond \emph{vs.} non-blond), where the spurious correlation is the gender. We also use standard dataset splits from prior work \cite{idrissi2022simple}. The dataset is licensed under the \textit{Creative Commons Attribution 4.0 International} license.

\paragraph{\metashift \cite{liang2022metashift}.} \metashift is a general method of creating image datasets from the Visual Genome project \cite{krishna2016connecting}. Here, we make use of the pre-processed Cat \emph{vs.} Dog dataset, where the goal is to distinguish between the two animals. The spurious attribute is the image background, where cats and more likely to be indoors, and dogs are more likely to be outdoors. We use the ``unmixed'' version generated from the authors' codebase.

\paragraph{\civilcomments \cite{borkan2019nuanced}.} \civilcomments is a binary classification text dataset, where the goal is to predict whether a internet comment contains toxic language. The spurious attribute is whether the text contains reference to eight demographic identities (\textit{male, female, LGBTQ, Christian, Muslim, other religions, Black,} and \textit{White}). We use the standard splits provided by the WILDS benchmark \cite{koh2021wilds}.

\paragraph{\multinli \cite{williams2017broad}.} \multinli is a text classification dataset with 3 classes, where the target is the natural language inference relationship between the premise and the hypothesis (neutral, contradiction, or entailment). The spurious attribute is whether negation appears in the text, as negation is highly correlated with the contradiction label. We use standard train/val/test splits given by prior work \cite{idrissi2022simple}.

\paragraph{\mimiccxr \cite{johnson2019mimic}.} \mimiccxr is a chest X-ray dataset originating from the Beth Israel Deaconess Medical Center from Boston, Massachusetts containing over 300,000 images. We use ``No Finding'' as the label, where a positive label means that the patient has no illness. Inspired by prior work \cite{seyyed2021underdiagnosis}, we use the intersection of race (\textit{White, Black, Other}) and gender as attributes. We randomly split the dataset into 85\% train, 5\% validation, and 10\% test splits.

\paragraph{\chexpert \cite{irvin2019chexpert}.} \chexpert is a chest X-ray dataset originating from the Stanford University Medical center containing over 200,000 images. We use the same data processing setup as \mimiccxr.

\paragraph{\cxrmultisite \cite{puli2021out}.} \cxrmultisite is a dataset proposed by \citet{puli2021out} which combines MIMIC-CXR \cite{johnson2019mimic} and CheXpert \cite{irvin2019chexpert} to create a semi-synthetic spurious correlation. The task is to predict pneumonia, and the dataset is constructed such that 90\% of the patients with pnuemonia are from MIMIC-CXR, and 90\% of the healthy patients are from CheXpert. Thus, the site where the image was taken is the spurious correlation. We create this correlation by subsampling. We randomly split the dataset into 85\% train, 5\% validation, and 10\% test splits.

\paragraph{\mimicnotes \cite{johnson2016mimic}.} \mimicnotes is a dataset used in a prior work \cite{chen2019can} showing differences in error rate between demographic groups in predicting mortality from clinical notes in MIMIC-III \cite{johnson2016mimic}. Following their work, we reproduce their dataset which consists of featurizing the first 48 hours of clinical text from a patient's hospital stay using the top 5,000 TF-IDF features. We use gender as the attribute.

\paragraph{\nicopp \cite{zhang2022nico++}.} \nicopp is a large-scale benchmark for domain generalization. Here, we use data from Track 1 (the common context generalization) of their challenge. We only use their training dataset, which consists of 60 classes and 6 common attributes (\textit{autumn, dim, grass, outdoor, rock, water}). To transform this dataset into the attribute generalization setting, we select all (attribute, label) pairs with less than 75 samples, and remove them from our training split, so they are only used for validation and testing. For each (attribute, label) pair, we use 25 samples for validation and 50 samples for testing, and use the remaining data as training samples.

\paragraph{\imagenetbg \cite{xiao2020noise}.} \imagenetbg is a benchmark created with the goal of evaluating the reliance of ImageNet classifiers on the background. The authors first created a subset of ImageNet with 9 classes (ImageNet-9), and annotated bounding boxes so that backgrounds can be removed. In our setup, we train models on the original IN-9L (with backgrounds), and evaluate our model on \texttt{MIXED-RAND}. Note that attribute (i.e., the label of the background) is not available for this dataset. This can be thought of as an attribute generalization setting, as we do not observe test backgrounds during training.

\paragraph{\living \cite{santurkar2020breeds}.} \living is a dataset created as part of the BREEDS benchmark for subpopulation shift. Their setup is slightly different from a traditional subpopulation shift setting, where subpopulations are defined using a WordNet hierarchy, and the goal is to generalize to unseen subclasses in the same hierarchy level. As such, it is difficult to define the notion of an ``attribute'' in this setting. In particular, the \living dataset consists of images of living objects across 17 classes. We train our models on the source subclasses and evaluate them on the target subclasses.

\begin{figure*}[!t]
\begin{center}
\subfigure[Spurious correlations.]{
    \label{subfig-appendix:label-distribution-sc}
    \includegraphics[width=0.235\linewidth]{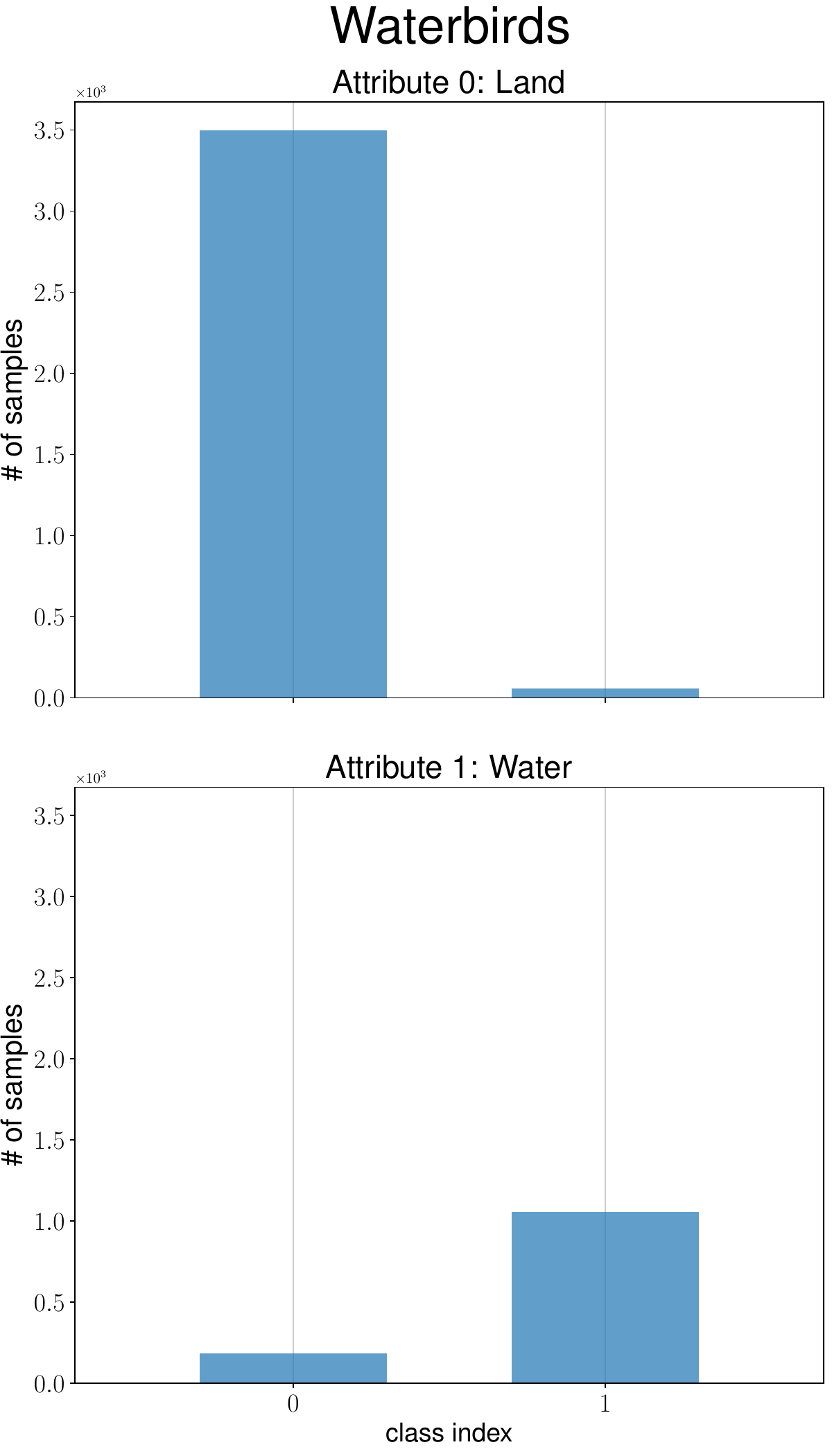}
}
\subfigure[Attribute imbalance.]{
    \label{subfig-appendix:label-distribution-ai}
    \includegraphics[width=0.235\linewidth]{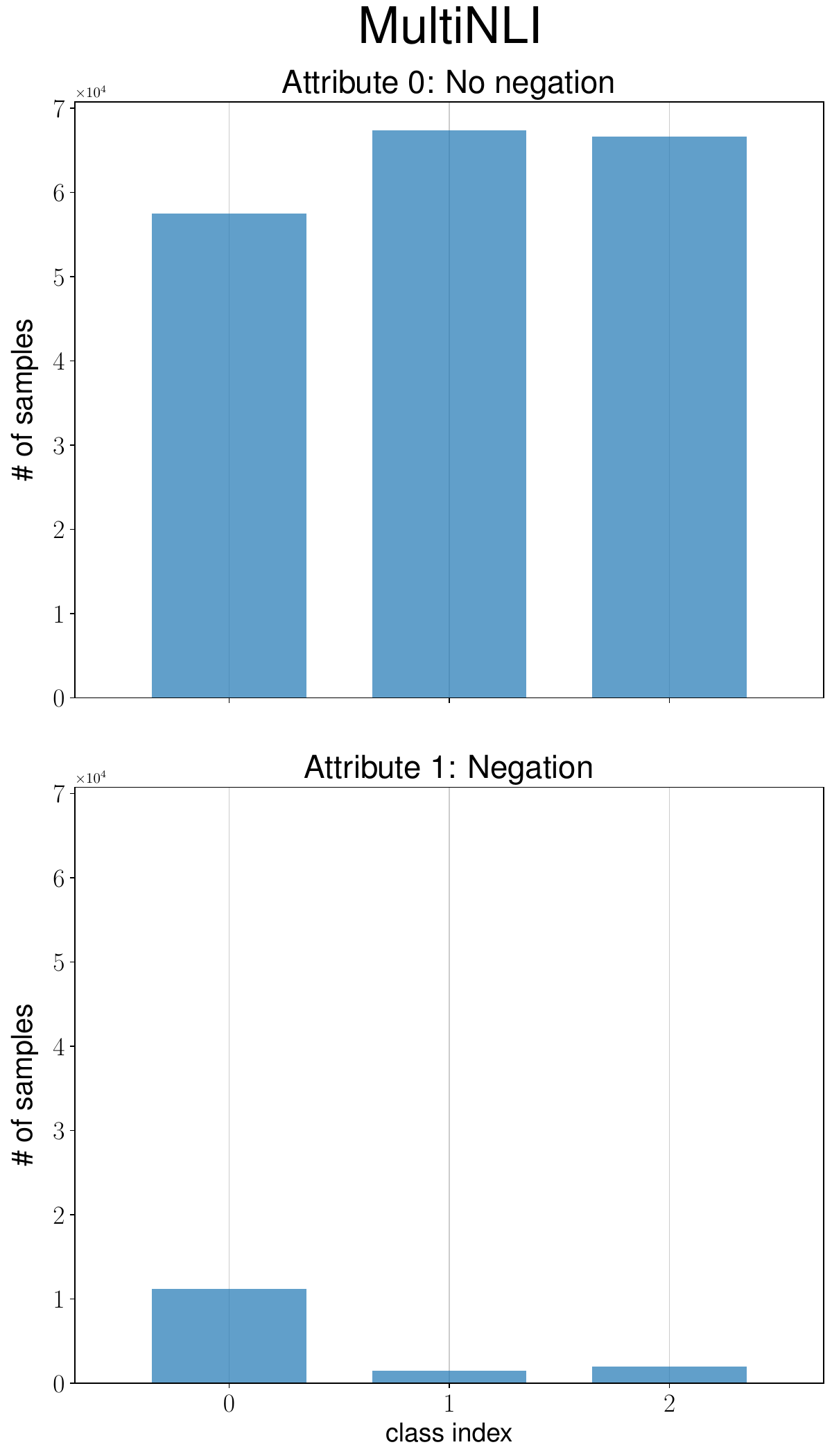}
}
\subfigure[Class imbalance.]{
    \label{subfig-appendix:label-distribution-ci}
    \includegraphics[width=0.235\linewidth]{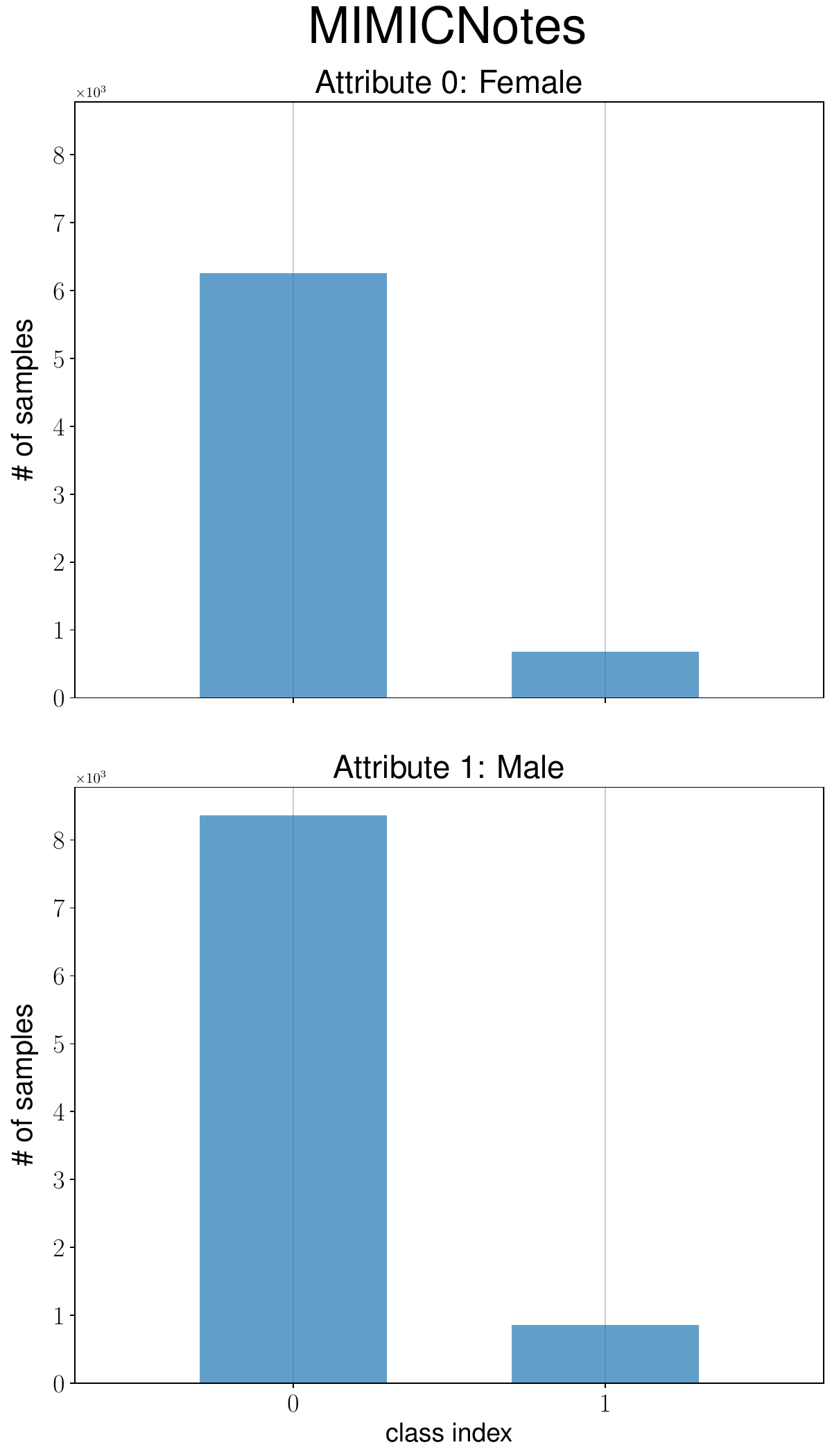}
}
\subfigure[Attribute generalization.]{
    \label{subfig-appendix:label-distribution-ag}
    \includegraphics[width=0.235\linewidth]{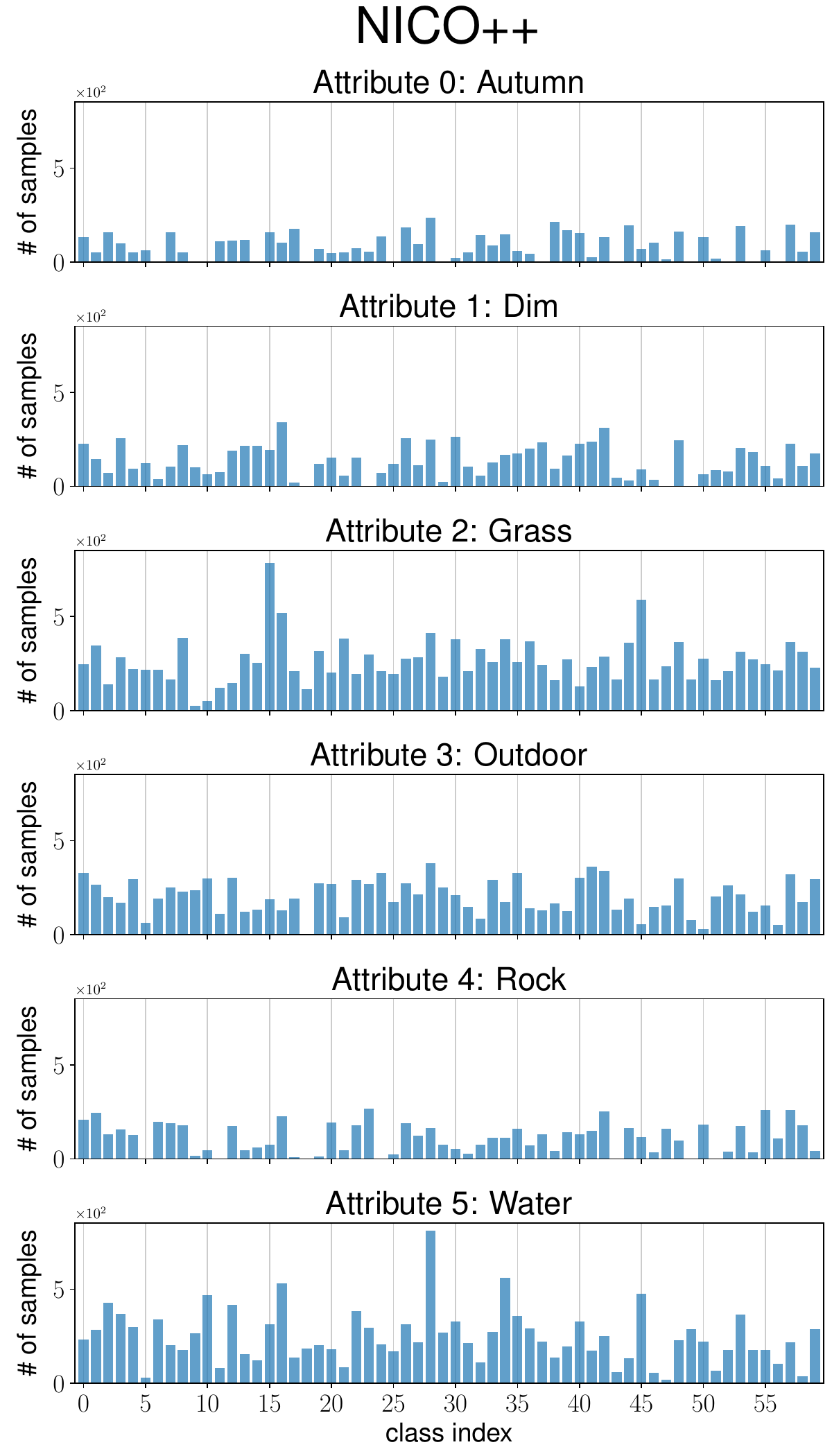}
}
\end{center}
\vspace{-0.5cm}
\caption{Typical label distributions for different types of subpopulation shift.}
\label{fig-appendix:label-distribution-diff-types}
\vspace{-0.2cm}
\end{figure*}

\paragraph{Label distribution for different types of subpopulation shift.}
Finally, we provide typical label distributions for different subpopulation shift types in \figref{fig-appendix:label-distribution-diff-types}. As highlighted, different shifts exhibit distinct types of label distributions, resulting in different properties in learning. For \nicopp (\figref{subfig-appendix:label-distribution-ag}), certain attributes have no training samples in certain classes.

\subsection{Algorithm Details}
\label{appendix-subsec:algo-details}

Our benchmark contains a large number of algorithms that span different learning strategies. We group them according to their categories, and provide detailed descriptions for each algorithm below.
\begin{itemize}
    \item \emph{Vanilla:} The empirical risk minimization (\textbf{ERM})~\cite{vapnik1999overview} minimizes the sum of errors across all samples.
    \item \emph{Subgroup robust methods:} Group distributionally robust optimization (\textbf{GroupDRO}) \cite{sagawa2020groupdro} performs ERM while increasing the importance of groups with larger errors. \textbf{CVaRDRO} \cite{duchi2018learning} proposes a variant of GroupDRO that dynamically weights data samples that have the highest losses. \textbf{LfF} \cite{nam2020learning} trains two models simultaneously, where the first model is biased and the second one is debiased by re-weighting the gradient of the loss. Just train twice (\textbf{JTT}) \cite{liu2021just} first trains an ERM model to identify minority groups in the training set and then trains a second ERM model with the identified samples being re-weighted. \textbf{LISA} \cite{yao2022improving} learns invariant predictors through data interpolation within and across attributes. Deep feature re-weighting (\textbf{DFR}) \cite{izmailov2022feature} first trains an ERM model, then retrains the last layer of the model using a balanced validation set with group annotations.
    \item \emph{Data augmentation:} \textbf{Mixup}~\cite{zhang2018mixup} performs ERM on linear interpolations of randomly sampled training examples and their labels.
    \item \emph{Domain-invariant representation learning:} Invariant risk minimization (\textbf{IRM}) \cite{arjovsky2019irm} learns a feature representation such that the optimal linear classifier on top of that representation matches across domains. Deep correlation alignment (\textbf{CORAL}) \cite{sun2016coral} matches the mean and covariance of feature distributions. Maximum mean discrepancy (\textbf{MMD}) \cite{li2018mmd} matches the MMD \cite{gretton2012kernel} of feature distributions. Note that all methods in this category require group annotations during training.
    \item \emph{Imbalanced learning:} \textbf{ReSample} \cite{japkowicz2000class} and \textbf{ReWeight} \cite{japkowicz2000class} simply re-sample or re-weight the inputs according to the number of samples per class. Focal loss (\textbf{Focal}) \cite{lin2017focal} reduces the relative loss for well-classified samples and focuses on difficult samples. Class-balanced loss (\textbf{CBLoss}) \cite{cui2019class} proposes re-weighting by the inverse effective number of samples. The LDAM loss (\textbf{LDAM}) \cite{cao2019learning} employs a modified marginal loss that favors minority samples more. Balanced-Softmax (\textbf{BSoftmax}) \cite{ren2020bsoftmax} extends Softmax to an unbiased estimation that considers the number of samples in each class. Classifier re-training (\textbf{CRT}) \cite{kang2020decoupling} decomposes the representation and classifier learning into two stages, where it fine-tunes the classifier using class-balanced sampling with representation fixed in the second stage. \textbf{ReWeightCRT} \cite{kang2020decoupling} is a re-weighting variant of CRT.
\end{itemize}

\subsection{Evaluation Metrics}
\label{appendix-subsec:eval-metrics}

We describe in detail all the evaluation metrics we used in our experiments.

\textbf{Average \& Worst Accuracy.} The average accuracy is defined as the accuracy over all samples. For worst-group accuracy (WGA), we compute the accuracy over all subgroups in the test set and report the worst one. When viewing each class as a subgroup, WGA degenerates to the worst-class accuracy.

\textbf{Average \& Worst Precision.} Precision is defined as $\text{TP} / (\text{TP} + \text{FP})$, where $\text{TP}$ is the number of true positives and $\text{FP}$ the number of false positives. Average precision simply takes the average precision score over all classes, whereas the worst precision reports the lowest precision value across classes.

\textbf{Average \& Worst F1-score.} The F1-score is defined as the harmonic mean of precision and recall. Average F1-score simply takes the average F1-score over all classes, whereas the worst F1-score reports the lowest value across all classes.

\textbf{Adjusted Accuracy.} Adjusted accuracy is defined as the average accuracy on a group-balanced dataset, which accounts for the data imbalance over subgroups.

\textbf{Balanced Accuracy.} Balanced accuracy is defined as the average of recall obtained on each class, taking the imbalance over classes into account.

\textbf{AUROC.} Following the common evaluation practice for the medical datasets used in our benchmark \cite{johnson2019mimic, irvin2019chexpert}, we also include the area under the receiver operating characteristic curve (AUROC) for evaluation.

\textbf{ECE \cite{guo2017calibration}.}
The expected calibration error (ECE) is defined as the difference in expected accuracy and expected confidence, which measures how close the output pseudo-probabilities match with the actual probabilities of a correct prediction (lower the better).

\subsection{Model Selection Protocol}
\label{appendix-subsec:model-select-attr-availability}

There has been an increasing interest in model selection within the literature on out-of-distribution generalization \cite{gulrajani2020domainbed}.
In subpopulation shift, model selection becomes essential especially when attributes are completely unknown in both training and validation set. Significant drop (over 20\%) in worst-group test accuracy has been reported if using the highest \emph{average} validation accuracy as the model selection criterion without any group annotations \cite{izmailov2022feature}.

Our benchmark provides different model selection strategies based on various evaluation metrics as described in Appendix \ref{appendix-subsec:eval-metrics}. Throughout the paper, we mainly use \emph{worst-group accuracy} as the metric for model selection (which degenerates to \emph{worst-class accuracy} when attributes are unknown in the validation set). Nevertheless, one can specify any aforementioned metric during model selection stage for experimenting with different selection strategies.

\section{Experimental Settings}
\label{appendix-sec:exp-setting}

\subsection{Implementation Details}
\label{appendix-subsec:implementation-details}

Following \cite{gulrajani2020domainbed, izmailov2022feature}, we use pretrained ResNet-50 model \cite{he2016deep} as the backbone network for image datasets (except for \living, which we train from scratch), and use pretrained BERT model \cite{idrissi2022simple} for all text datasets. We employ a three-layer MLP for \mimicnotes dataset given its simplicity.
For all image datasets, we follow standard pre-processing steps \cite{idrissi2022simple}: resize and center crop the image to $224\times 224$ pixels, and perform normalization using the ImageNet channel statistics.
Following the literature \cite{izmailov2022feature, idrissi2022simple}, we use the AdamW optimizer \cite{kingma2015adam} for all text datasets, and use SGD with momentum for all image datasets.
We train all models for 5,000 steps on \waterbirds and \metashift, 10,000 steps on \mimicnotes and \imagenetbg, 20,000 steps on \chexpert and \cxrmultisite, and 30,000 steps on all other datasets to ensure convergence.

\subsection{Hyperparameters Search Protocol}
\label{appendix-subsec:hyper-params}

For a fair evaluation across different algorithms, following the training protocol in \cite{gulrajani2020domainbed}, for each algorithm we conduct a random search of 16 trials over a joint distribution of its all hyperparameters. We then use the validation set to select the best hyperparameters for each algorithm, fix them and rerun the experiments under 3 different random seeds to report the final average results (and standard deviation). Such process ensures the comparison is best-versus-best, and the hyperparameters are optimized for all algorithms.

We detail the hyperparameter choices for each algorithm in Table \ref{appendix:table:hyperparameters}.

\begin{table}[!t]
\setlength{\tabcolsep}{10pt}
\caption{Hyperparameters search space for all experiments.}
\vspace{5pt}
\label{appendix:table:hyperparameters}
\small
\begin{center}
\begin{tabular}{llll}
\toprule[1.5pt]
\textbf{Condition} & \textbf{Parameter} & \textbf{Default value} & \textbf{Random distribution} \\
\midrule\midrule
\multicolumn{4}{l}{\emph{\textbf{General:}}} \\
\midrule
\multirow{2}{*}{ResNet}     & learning rate & 0.001 & $10^{\text{Uniform}(-4, -2)}$ \\
                            & batch size & 108 & $2^{\text{Uniform}(6, 7)}$ \\
\midrule
\multirow{3}{*}{BERT}       & learning rate & 0.00001 & $10^{\text{Uniform}(-5.5, -4)}$ \\
                            & batch size & 32 & $2^{\text{Uniform}(3, 5.5)}$ \\
                            & dropout & 0.5 & $\text{RandomChoice}([0, 0.1, 0.5])$ \\
\midrule
\multirow{2}{*}{MLP}       & learning rate & 0.001 & $10^{\text{Uniform}(-4, -2)}$ \\
                            & batch size & 256 & $2^{\text{Uniform}(7, 10)}$ \\
\midrule\midrule
\multicolumn{4}{l}{\emph{\textbf{Algorithm-specific:}}} \\
\midrule
\multirow{2}{*}{IRM}        & lambda & 100 & $10^{\text{Uniform}(-1, 5)}$ \\
                            & iterations of penalty annealing & 500 & $10^{\text{Uniform}(0, 4)}$ \\
\midrule
GroupDRO                    & eta & 0.01 & $10^{\text{Uniform}(-3, -1)}$ \\
\midrule
Mixup                       & alpha & 0.2 & $10^{\text{Uniform}(0, 4)}$ \\
\midrule
CVaRDRO                     & alpha & 0.1 & $10^{\text{Uniform}(-2, 0)}$ \\
\midrule
\multirow{2}{*}{JTT}        & first stage step fraction & 0.5 & $\text{Uniform}(0.2, 0.8)$ \\
                            & lambda & 10 & $10^{\text{Uniform}(0, 2.5)}$ \\
\midrule
\multirow{2}{*}{LISA}       & alpha & 2 & $10^{\text{Uniform}(-1, 1)}$ \\
                            & p\_select & 0.5 & $\text{Uniform}(0, 1)$ \\
\midrule
LfF                         & q & 0.7 & $\text{Uniform}(0.05, 0.95)$ \\
\midrule
DFR                         & regularization & 0.1 & $10^{\text{Uniform}(-2, 0.5)}$ \\
\midrule
CORAL, MMD                  & gamma & 1 & $10^{\text{Uniform}(-1, 1)}$ \\
\midrule
Focal                       & gamma & 1 & $0.5 * 10^{\text{Uniform}(0, 1)}$ \\
\midrule
CBLoss                      & beta & 0.9999 & $1 - 10^{\text{Uniform}(-5, -2)}$ \\
\midrule
\multirow{2}{*}{LDAM}       & max\_m & 0.5 & $10^{\text{Uniform}(-1, -0.1)}$ \\
                            & scale & 30 & $\text{RandomChoice}([10, 30])$ \\
\bottomrule[1.5pt]
\end{tabular}
\end{center}
\end{table}

\section{Additional Analysis and Studies}
\label{appendix-sec:additional-analysis}

\subsection{Quantifying the Degree of Different Shifts}
\label{appendix-subsec:quantify-shift}

In order to quantify the degree of each shift for each dataset relative to others, we use several simple metrics (see \tabref{appendix-table:quantify-shift-sc}, \tabref{appendix-table:quantify-shift-ai}, and \tabref{appendix-table:quantify-shift-ci}). For spurious correlations, we use:
\begin{itemize}
    \item The Mutual Information (\textbf{MI}) between $A$ and $Y$, $I(A; Y)$.
    \item The Normalized Mutual Information (\textbf{NMI}) between $A$ and $Y$, where \text{norm } $I(A; Y) = 1$ indicates that the two are perfectly correlated:
        \begin{equation*}
            \text{norm } I(A; Y) = \frac{2 I(A; Y)}{H(Y) + H(A)}.
        \end{equation*} 
    \item \textbf{Cramer's V}, which is an association measure based on the Chi-squared test statistic. It has a range of $[0, 1]$, where 1 indicates perfect correlation.
    \item \textbf{Tschuprow's T}, which is closely related to Cramer's V. It also has a range of $[0, 1]$.
\end{itemize}

Note that we only examine the correlation between $A$ and $Y$, but not the degree of effectiveness to which $A$ can be inferred from $X$. This is an important component, as the model can not take advantage of the spurious correlation if it could not be learnt easily. However, we would expect that most attributes (e.g., words in text, image backgrounds) should be easily inferred from the inputs for the datasets we examine.

For attribute and class imbalance, we use the following metrics (shown for the class imbalance case):
\begin{itemize}
    \item \textbf{Entropy}: $H(Y)$.
    \item \textbf{Normalized Entropy}, where $\text{norm } H(Y) = 1$ means that the distribution is uniform (i.e., no imbalance):
        \begin{equation*}
            \text{norm } H(Y) = \frac{H(Y)}{\log |\text{supp}(Y)|}.
        \end{equation*}
    \item Difference between the probability of the most frequent class and the probability of the least frequent class ($p_{\text{max}} - p_{\text{min}}$).
\end{itemize}

For attribute generalization, we simply examine whether there exist any subpopulations in the test set which do not appear during training.

\begin{table}[H]
\setlength{\tabcolsep}{10pt}
\caption{Metrics for quantifying the degree of \emph{spurious correlations}.}
\vspace{3pt}
\label{appendix-table:quantify-shift-sc}
\small
\begin{center}
\begin{tabular}{lcccc}
\toprule[1.5pt]
\textbf{Dataset} & \textbf{MI}$^\uparrow$ & \textbf{NMI}$^\uparrow$ &  \textbf{Cramer}$^\uparrow$ & \textbf{Tschuprow}$^\uparrow$ \\ \midrule
\waterbirds & 0.37 & 0.67 &  0.87 & 0.87 \\[1.2pt]
\celeba & 0.06 & 0.11 &  0.31 & 0.31 \\[1.2pt]
\metashift & 0.09 & 0.13 &  0.41 & 0.41 \\[1.2pt]
\civilcomments & 0.02 & 0.02 & 0.19 & 0.11 \\[1.2pt]
\multinli & 0.03 & 0.04 & 0.25 & 0.21 \\[1.2pt]
\mimiccxr & 0.01 & 0.01 &  0.15 & 0.10 \\[1.2pt]
\mimicnotes & $<1e^{-4}$ & $<1e^{-4}$   & 0.01 & 0.01 \\[1.2pt]
\cxrmultisite & 0.03 & 0.13  & 0.32 & 0.32 \\[1.2pt]
\chexpert & $<1e^{-3}$ & $<1e^{-3}$  & 0.03 & 0.02 \\[1.2pt]
\nicopp & 0.11 & 0.04 & 0.20 & 0.11 \\[1.2pt]
\imagenetbg & $-$ & $-$ & $-$ &  $-$ \\[1.2pt]
\living & $-$ & $-$ & $-$ & $-$ \\
\bottomrule[1.5pt]
\end{tabular}
\end{center}
\vspace{-0.6cm}
\end{table}

\begin{table}[H]
\setlength{\tabcolsep}{8pt}
\caption{Metrics for quantifying the degree of \emph{attribute imbalance}.}
\vspace{3pt}
\label{appendix-table:quantify-shift-ai}
\small
\begin{center}
\begin{tabular}{lccc}
\toprule[1.5pt]
\textbf{Dataset} & \textbf{Entropy}$^\downarrow$ & \textbf{N. Entropy}$^\downarrow$ & {$p_{\text{max}} - p_{\text{min}}$}$^\uparrow$  \\ \midrule
\waterbirds & 0.82 & 0.82 & 0.48  \\[1.2pt]
\celeba & 0.98 & 0.98 & 0.16 \\[1.2pt]
\metashift & 0.99 & 0.99 & 0.14  \\[1.2pt]
\civilcomments & 2.78 & 0.93 & 0.20  \\[1.2pt]
\multinli & 0.37 & 0.37 & 0.86  \\[1.2pt]
\mimiccxr & 2.33 & 0.90 & 0.27  \\[1.2pt]
\mimicnotes & 0.99 & 0.99 & 0.14  \\[1.2pt]
\cxrmultisite & 0.51 & 0.51 & 0.77  \\[1.2pt]
\chexpert & 2.20 & 0.85 & 0.32  \\[1.2pt]
\nicopp & 2.47 & 0.96 & 0.17  \\[1.2pt]
\imagenetbg & $-$ & $-$ & $-$  \\[1.2pt]
\living & $-$ & $-$ & $-$ \\
\bottomrule[1.5pt]
\end{tabular}
\end{center}
\vspace{-0.6cm}
\end{table}

\begin{table}[H]
\setlength{\tabcolsep}{8pt}
\caption{Metrics for quantifying the degree of \emph{class imbalance}.}
\vspace{3pt}
\label{appendix-table:quantify-shift-ci}
\small
\begin{center}
\begin{tabular}{lccc}
\toprule[1.5pt]
\textbf{Dataset} & \textbf{Entropy}$^\downarrow$ & \textbf{N. Entropy}$^\downarrow$ & {$p_{\text{max}} - p_{\text{min}}$}$^\uparrow$  \\ \midrule
\waterbirds & 0.78 & 0.78 & 0.54  \\[1.2pt]
\celeba & 0.61 & 0.61 & 0.70  \\[1.2pt]
\metashift & 0.99 & 0.99 & 0.13 \\[1.2pt]
\civilcomments & 0.67 & 0.67 & 0.65 \\[1.2pt]
\multinli & 1.58 & 0.99 & 0.001 \\[1.2pt]
\mimiccxr & 0.97 & 0.97 & 0.20 \\[1.2pt]
\mimicnotes & 0.45 & 0.45 & 0.81  \\[1.2pt]
\cxrmultisite & 0.12 & 0.12 & 0.97  \\[1.2pt]
\chexpert & 0.47 & 0.47 & 0.80  \\[1.2pt]
\nicopp & 5.81 & 0.98 & 0.03  \\[1.2pt]
\imagenetbg & 3.17 & 1 & 0  \\[1.2pt]
\living & 4.09 & 1 & 0  \\
\bottomrule[1.5pt]
\end{tabular}
\end{center}
\vspace{-0.3cm}
\end{table}

\subsection{Improvements across Different Shifts \& Settings}
\label{appendix-subsec:improvements-diff-settings}

\begin{figure*}[!t]
\begin{center}
\subfigure[Train \& validation attributes both known (\emph{oracle selection}).]{
    \label{subfig-appendix:improvements-oracle}
    \includegraphics[width=0.95\linewidth]{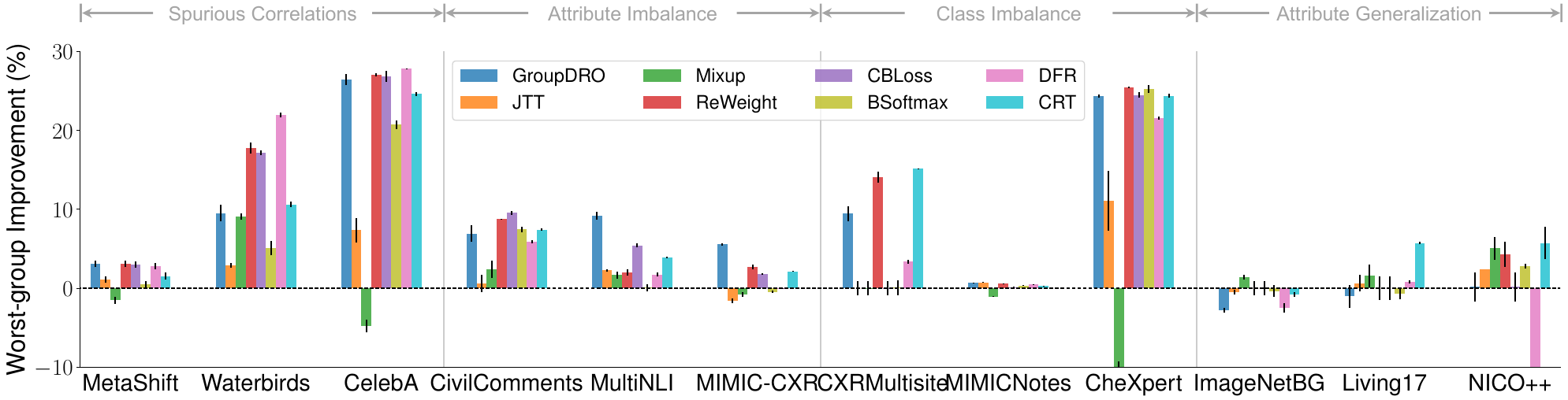}
}
\\
\subfigure[Train attributes unknown, but validation attributes known (\emph{worst-group accuracy selection}).]{
    \label{subfig-appendix:improvements-valAttrY}
    \includegraphics[width=0.95\linewidth]{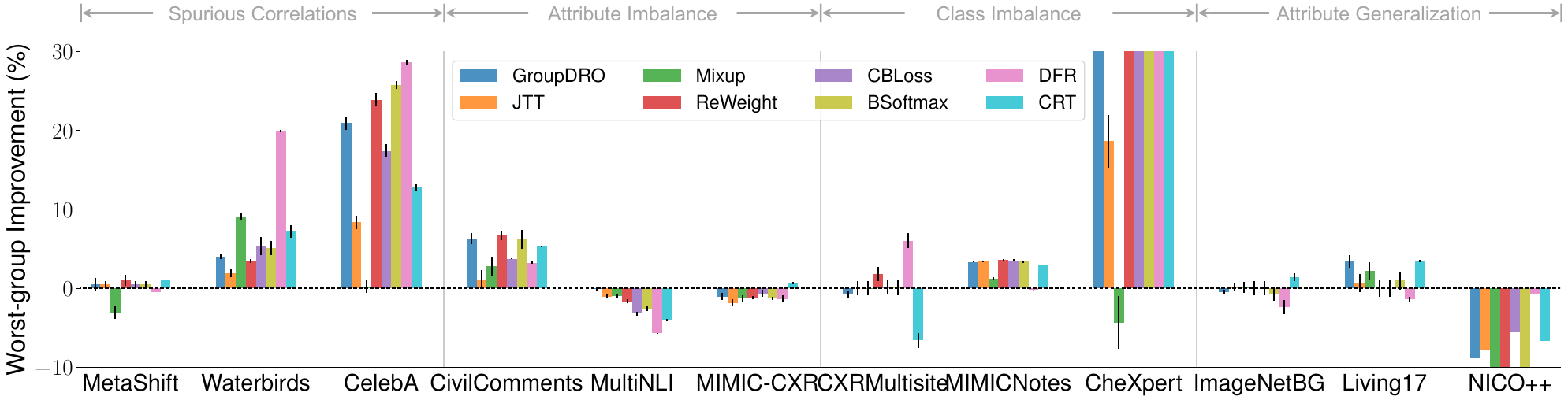}
}
\\
\subfigure[Train \& validation attributes both unknown (\emph{worst-class accuracy selection}).]{
    \label{subfig-appendix:improvements-valAttrN}
    \includegraphics[width=0.95\linewidth]{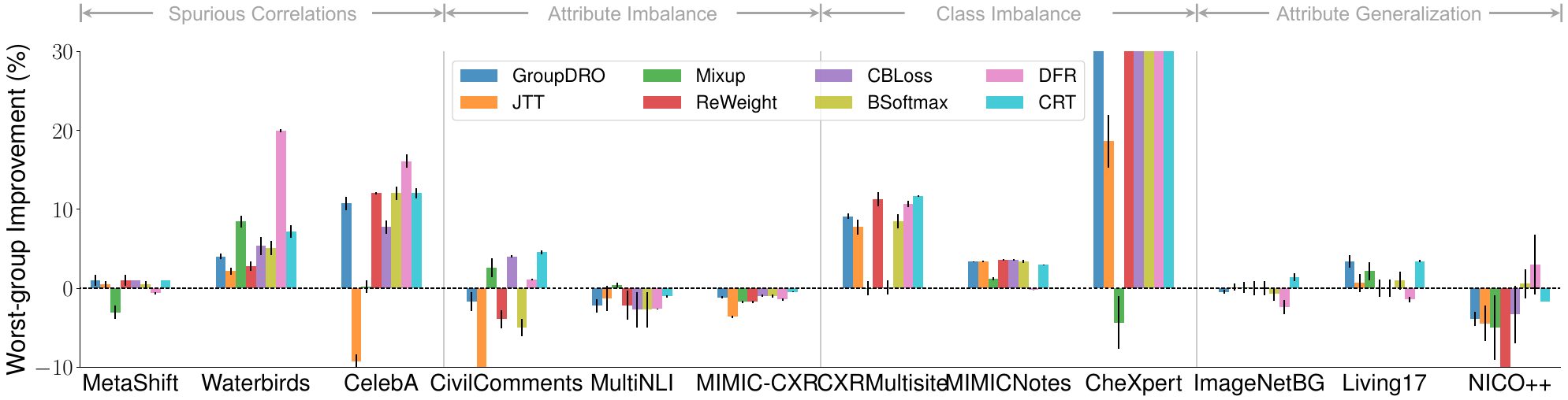}
}
\end{center}
\vspace{-0.5cm}
\caption{Complete results on worst-group performance improvements over ERM under different settings.}
\label{fig-appendix:improvements}
\vspace{-0.1cm}
\end{figure*}

We show in \figref{fig-appendix:improvements} the complete results on worst-group performance improvements over ERM under different settings. As can be observed from all figures, algorithmic advances have been made for \emph{spurious correlations} and \emph{class imbalance}, where consistent improvements can be obtained across different training \& validation settings.
Yet, small overall improvements are observed for \emph{attribute imbalance}, while almost no performance gains can be obtained for \emph{attribute generalization}, indicating the limitation of SOTA algorithms on tackling these types of subpopulation shift.

\begin{table}[!t]
\caption{Test-set worst-group accuracy difference (\%) between each selection strategy on each dataset, relative to the oracle which selects the best test-set worst-group accuracy. Note that we have only defined AUPRC and Brier score for the binary classification case.}
\vspace{3pt}
\label{appendix-table:model-selection-full-results}
\resizebox{1\textwidth}{!}{
\begin{tabular}{lrrrrrrrrrrrr|r}
\toprule[1.5pt]
\textbf{Selection Strategy} & \cxrmultisite              & \celeba          & \chexpert & \civilcomments & \imagenetbg     & \living       & \mimiccxr  &  \mimicnotes        & \metashift       & \multinli        & \nicopp           &  \waterbirds       & \textbf{Avg}  \\ \midrule
Max Worst-Class Accuracy      & -6.9 \scriptsize$\pm10.7$          & -5.0 \scriptsize$\pm6.3$   &\textbf{-0.4} \scriptsize$\pm0.8$   & \textbf{-3.2} \scriptsize$\pm5.2$   & \textbf{-0.7} \scriptsize$\pm1.3$  & \textbf{-1.6} \scriptsize$\pm2.3$  & \textbf{-0.9} \scriptsize$\pm1.0$ & \textbf{-0.1} \scriptsize$\pm0.5$   & \textbf{-1.5} \scriptsize$\pm3.0$ & \textbf{-1.9} \scriptsize$\pm2.9$ & \textbf{-5.3} \scriptsize$\pm5.6$   & \textbf{-0.8} \scriptsize$\pm1.4$  & \textbf{-2.4} \\[1.2pt]
Max Balanced Accuracy        & -6.9 \scriptsize$\pm10.7$          & -4.4 \scriptsize$\pm5.4$   & -1.3 \scriptsize$\pm2.5$   & -3.5 \scriptsize$\pm5.8$   & -0.9 \scriptsize$\pm1.6$  & -4.5 \scriptsize$\pm5.4$  & -2.9 \scriptsize$\pm4.9$ & -2.3 \scriptsize$\pm6.2$   & -1.7 \scriptsize$\pm3.0$ & -3.7 \scriptsize$\pm3.9$ & -7.0 \scriptsize$\pm5.8$   & -1.3 \scriptsize$\pm1.9$  & -3.4          \\[1.2pt]
Min Class Accuracy Diff      & \textbf{-6.2 \scriptsize$\pm10.3$} & -6.1 \scriptsize$\pm9.1$   & -1.9 \scriptsize$\pm5.3$   & -4.1 \scriptsize$\pm8.0$   & -2.8 \scriptsize$\pm13.0$ & -5.1 \scriptsize$\pm10.0$ & -1.9 \scriptsize$\pm5.0$ & -0.3 \scriptsize$\pm1.2$   & -2.2 \scriptsize$\pm4.6$ & -5.7 \scriptsize$\pm8.6$ & -27.2 \scriptsize$\pm15.4$ & -2.4 \scriptsize$\pm4.8$  & -5.5          \\[1.2pt]
Max Worst-Class F1           & -7.7 \scriptsize$\pm11.3$          & -13.4 \scriptsize$\pm10.4$ & -5.4 \scriptsize$\pm6.7$   & -3.2 \scriptsize$\pm3.8$   & -0.8 \scriptsize$\pm1.2$  & -3.5 \scriptsize$\pm4.4$  & -2.5 \scriptsize$\pm2.2$ & -4.4 \scriptsize$\pm8.7$   & -1.8 \scriptsize$\pm3.3$ & -2.3 \scriptsize$\pm3.0$ & -6.7 \scriptsize$\pm6.3$   & -2.6 \scriptsize$\pm3.5$  & -4.5          \\[1.2pt]
Max Macro Avg F1             & -8.2 \scriptsize$\pm11.6$  & -14.3 \scriptsize$\pm10.6$ & -7.7 \scriptsize$\pm9.8$   & -5.1 \scriptsize$\pm4.7$   & -0.9 \scriptsize$\pm1.5$   & -4.4 \scriptsize$\pm5.3$   & -2.8 \scriptsize$\pm4.5$   & -8.2 \scriptsize$\pm13.2$  & -1.8 \scriptsize$\pm2.9$   & -3.3 \scriptsize$\pm3.4$   & -7.0 \scriptsize$\pm5.8$   & -3.1 \scriptsize$\pm4.0$   & -5.6         \\[1.2pt]
Min Per-Class Recall Stdev.    & -6.2 \scriptsize$\pm10.3$  & -6.1 \scriptsize$\pm9.1$   & -1.9 \scriptsize$\pm5.3$   & -4.1 \scriptsize$\pm8.0$   & -2.3 \scriptsize$\pm11.5$  & -5.5 \scriptsize$\pm9.1$   & -1.9 \scriptsize$\pm5.0$   & -0.3 \scriptsize$\pm1.2$   & -2.2 \scriptsize$\pm4.6$   & -5.6 \scriptsize$\pm8.7$   & -29.7 \scriptsize$\pm14.3$ & -2.4 \scriptsize$\pm4.8$   & -5.7         \\[1.2pt]
Max Weighted Avg Precision   & -8.3 \scriptsize$\pm11.5$  & -13.5 \scriptsize$\pm10.1$ & -6.3 \scriptsize$\pm11.1$  & -5.7 \scriptsize$\pm8.6$   & -0.8 \scriptsize$\pm1.3$   & -7.5 \scriptsize$\pm7.8$   & -4.3 \scriptsize$\pm6.4$   & -12.6 \scriptsize$\pm21.5$ & -3.3 \scriptsize$\pm8.0$   & -3.4 \scriptsize$\pm4.7$   & -6.8 \scriptsize$\pm5.5$   & -4.9 \scriptsize$\pm10.1$  & -6.5         \\[1.2pt]
Max Overall AUROC            & -10.0 \scriptsize$\pm12.5$         & -12.2 \scriptsize$\pm10.3$ & -10.4 \scriptsize$\pm13.0$ & -8.2 \scriptsize$\pm9.0$   & -1.1 \scriptsize$\pm2.1$  & -5.5 \scriptsize$\pm6.7$  & -6.6 \scriptsize$\pm9.9$ & -10.0 \scriptsize$\pm16.5$ & -3.2 \scriptsize$\pm7.0$ & -4.4 \scriptsize$\pm5.8$ & -6.9 \scriptsize$\pm6.3$   & -2.6 \scriptsize$\pm6.1$  & -6.7          \\[1.2pt]
Max Overall AUPRC            & -10.0 \scriptsize$\pm12.5$ & -13.0 \scriptsize$\pm10.3$ & -11.6 \scriptsize$\pm11.9$ & -8.1 \scriptsize$\pm8.9$   & -    & -    & -7.3 \scriptsize$\pm10.2$  & -9.6 \scriptsize$\pm16.3$  & -2.7 \scriptsize$\pm6.2$   & -    & -    & -4.0 \scriptsize$\pm9.5$   & -8.3         \\[1.2pt]
Min Overall BCE              & -8.2 \scriptsize$\pm11.5$  & -18.1 \scriptsize$\pm13.2$ & -18.7 \scriptsize$\pm16.4$ & -13.1 \scriptsize$\pm12.3$ & -0.9 \scriptsize$\pm1.6$   & -7.2 \scriptsize$\pm7.3$   & -7.2 \scriptsize$\pm12.0$  & -14.3 \scriptsize$\pm20.7$ & -3.7 \scriptsize$\pm7.7$   & -6.2 \scriptsize$\pm7.8$   & -7.6 \scriptsize$\pm6.1$   & -12.5 \scriptsize$\pm18.4$ & -9.8         \\[1.2pt]
Max Per-class Precision      & -8.2 \scriptsize$\pm11.7$  & \textbf{-3.0} \scriptsize$\pm8.9$   & -6.8 \scriptsize$\pm12.5$  & -14.8 \scriptsize$\pm24.3$ & -7.6 \scriptsize$\pm18.4$  & -19.3 \scriptsize$\pm15.9$ & -9.4 \scriptsize$\pm12.7$  & -12.6 \scriptsize$\pm22.4$ & -9.9 \scriptsize$\pm17.4$  & -6.6 \scriptsize$\pm10.1$  & -14.8 \scriptsize$\pm11.8$ & -5.3 \scriptsize$\pm12.4$  & -9.8         \\[1.2pt]
Max Overall Accuracy         & -8.2 \scriptsize$\pm11.4$          & -18.6 \scriptsize$\pm12.0$ & -30.9 \scriptsize$\pm24.9$ & -13.7 \scriptsize$\pm9.5$  & -0.9 \scriptsize$\pm1.6$  & -4.5 \scriptsize$\pm5.4$  & -5.1 \scriptsize$\pm6.3$ & -19.9 \scriptsize$\pm26.0$ & -1.9 \scriptsize$\pm3.3$ & -3.7 \scriptsize$\pm3.9$ & -7.1 \scriptsize$\pm5.8$   & -7.2 \scriptsize$\pm11.7$ & -10.2 \\[1.2pt]
Min Overall Brier Score           & -8.2 \scriptsize$\pm11.5$ & -18.8 \scriptsize$\pm13.1$ & -19.6 \scriptsize$\pm16.6$ & -13.5 \scriptsize$\pm12.3$ & -   & -    & -7.1 \scriptsize$\pm12.0$  & -15.1 \scriptsize$\pm21.6$ & -2.7 \scriptsize$\pm5.3$   & -    & -    & -6.9 \scriptsize$\pm11.0$  & -11.5        \\[1.2pt]
Min Overall ECE              & -8.2 \scriptsize$\pm11.5$ & -20.5 \scriptsize$\pm15.7$ & -20.3 \scriptsize$\pm17.4$ & -14.4 \scriptsize$\pm13.5$ & -16.9 \scriptsize$\pm33.6$ & -28.8 \scriptsize$\pm19.6$ & -12.3 \scriptsize$\pm18.2$ & -16.2 \scriptsize$\pm22.7$ & -20.9 \scriptsize$\pm28.8$ & -24.6 \scriptsize$\pm19.0$ & -20.0 \scriptsize$\pm14.3$ & -11.0 \scriptsize$\pm17.9$ & -17.9    \\  
\bottomrule[1.5pt]
\end{tabular}
}
\vspace{-0.1cm}
\end{table}

\subsection{Model Selection without Validation Attributes}
\label{appendix-subsec:model-selection-metrics}

In the main paper, we examine the feasibility of different metrics for model selection without group-annotated validation data. We further confirm this in \tabref{appendix-table:model-selection-full-results} by showing the results for more selection strategies with all metrics across all datasets in our benchmark.
Specifically, when using \emph{worst-class accuracy} as the model selection criterion, on average we achieve only \textbf{2.4\%} degrade of worst-group accuracy compared to oracle selection method. The selection criterion also performs the best over all other selection metrics on 10 out of 12 datasets, indicating its effectiveness for reliable model selection without \emph{any} attribute information.

\begin{figure*}[!t]
\begin{center}
\subfigure[Adjusted accuracy.]{
    \label{subfig-appendix:tradeoff-adjusted}
    \includegraphics[width=0.95\linewidth]{figures/tradeoff_main_Adjusted.pdf}
}
\\[-0.5pt]
\subfigure[Balanced accuracy.]{
    \label{subfig-appendix:tradeoff-balanced}
    \includegraphics[width=0.95\linewidth]{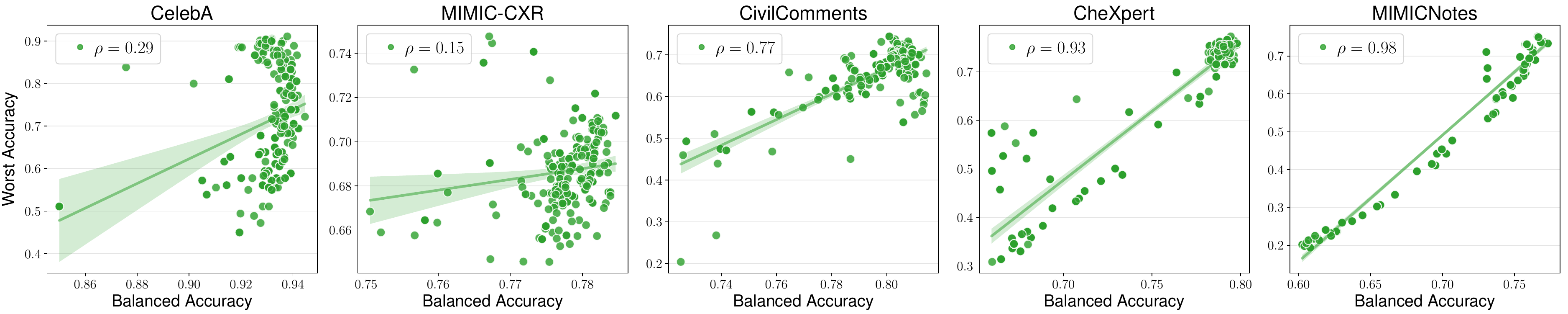}
}
\end{center}
\vspace{-0.5cm}
\caption{\textbf{Accuracy on the line}. We show metrics that are \emph{positively} correlated with worst-group accuracy.}
\label{fig-appendix:tradeoff-acc-on-the-line}
\end{figure*}

\begin{figure*}[!t]
\begin{center}
\subfigure[Worst precision.]{
    \label{subfig-appendix:tradeoff-worstprecision}
    \includegraphics[width=0.95\linewidth]{figures/tradeoff_main_WorstPrec.pdf}
}
\\[-0.5pt]
\subfigure[ECE.]{
    \label{subfig-appendix:tradeoff-ece}
    \includegraphics[width=0.95\linewidth]{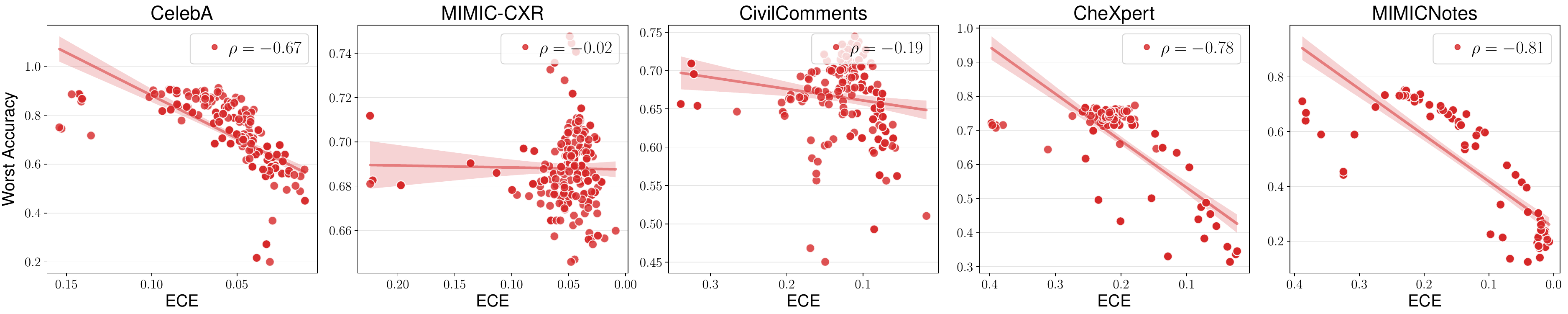}
}
\end{center}
\vspace{-0.5cm}
\caption{\textbf{Accuracy on the inverse line}. We show metrics that are \emph{negatively} correlated with worst-group accuracy.}
\label{fig-appendix:tradeoff-acc-on-the-inverse-line}
\end{figure*}

\begin{figure*}[!t]
\begin{center}
\subfigure[Worst F1-score.]{
    \label{subfig-appendix:tradeoff-worstf1}
    \includegraphics[width=0.95\linewidth]{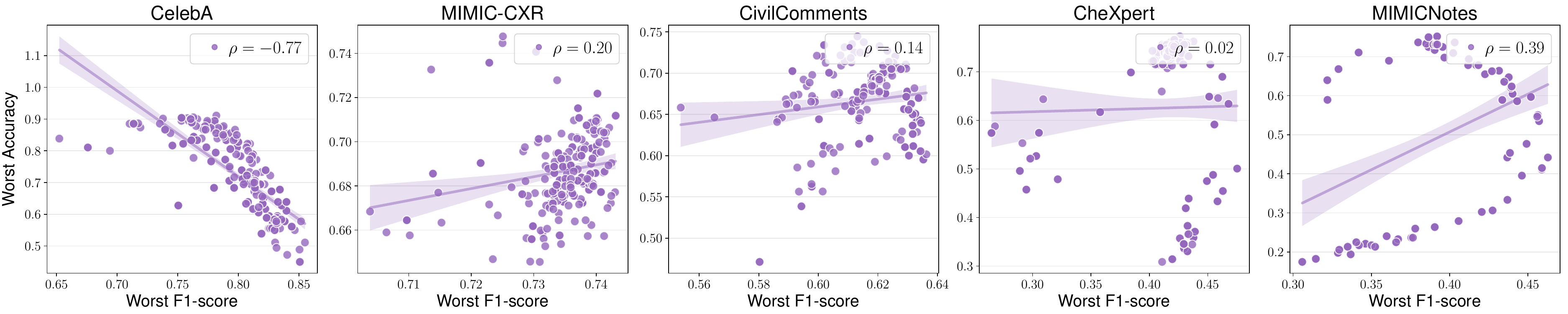}
}
\\[-0.5pt]
\subfigure[Average accuracy.]{
    \label{subfig-appendix:tradeoff-avg}
    \includegraphics[width=0.95\linewidth]{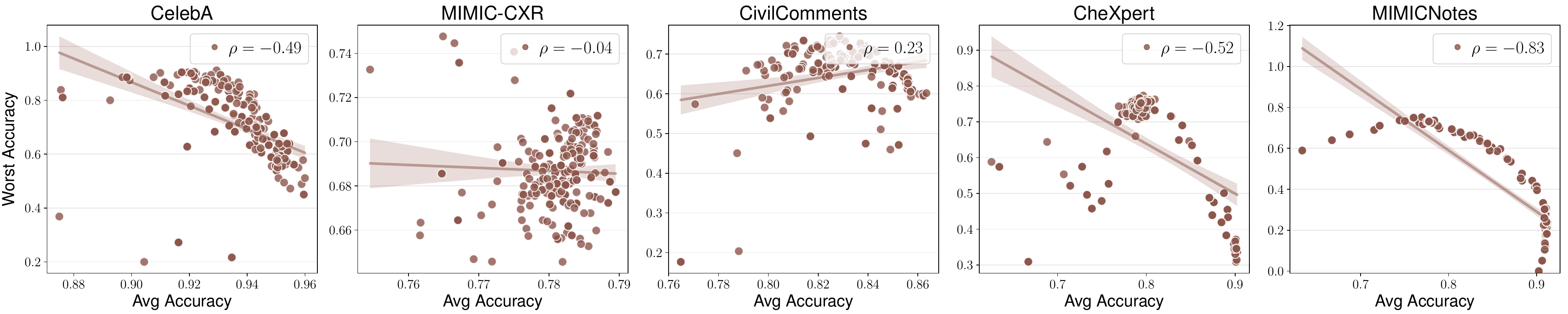}
}
\\[-0.5pt]
\subfigure[Average precision.]{
    \label{subfig-appendix:tradeoff-avgprecision}
    \includegraphics[width=0.95\linewidth]{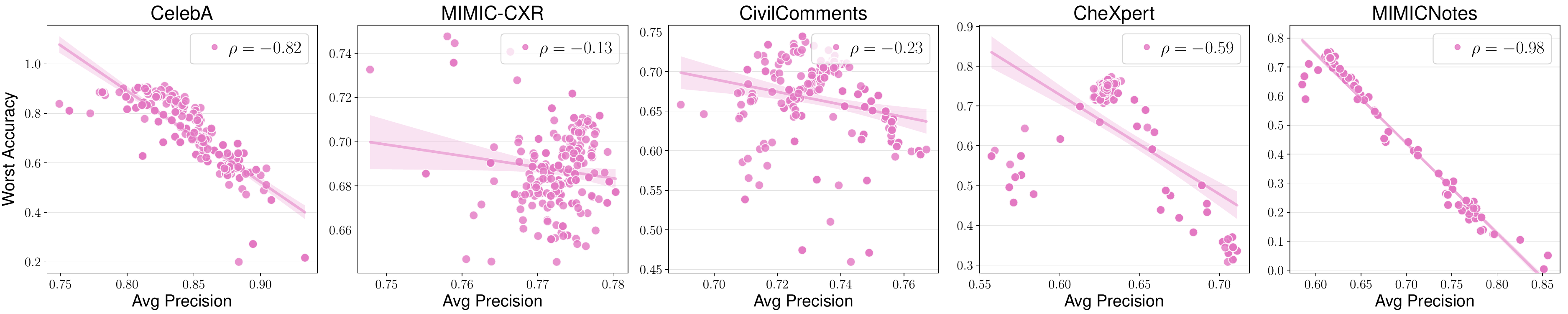}
}
\\[-0.5pt]
\subfigure[Average F1-score.]{
    \label{subfig-appendix:tradeoff-avgf1}
    \includegraphics[width=0.95\linewidth]{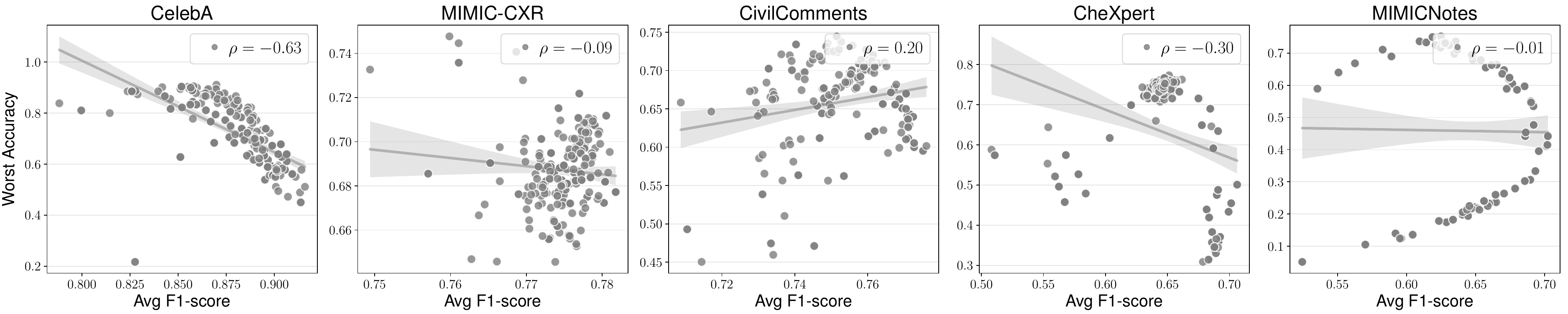}
}
\end{center}
\vspace{-0.5cm}
\caption{\textbf{Accuracy not on the line}. We show metrics that do not demonstrate consistent correlations across datasets with worst-group accuracy.}
\label{fig-appendix:tradeoff-acc-not-on-the-line}
\vspace{-0.3cm}
\end{figure*}

\subsection{Rethinking Evaluation Metrics in Subpopulation Shift}
\label{appendix-subsec:tradeoff-metrics}

We provide complete results on the correlation between worst-group accuracy (WGA) and other metrics we consider in our benchmark.

\textbf{Accuracy on the line.}
In the main paper we show that certain metrics exhibit high linear correlation with WGA. We further show in \figref{fig-appendix:tradeoff-acc-on-the-line} with a full list of metrics that exhibit consistent positive correlation across diverse datasets. Specifically, both adjusted accuracy and balanced accuracy display the ``\emph{accuracy on the line}'' property, which has also been confirmed in prior work \cite{izmailov2022feature}.

\textbf{Accuracy on the inverse line.}
More interestingly, we further establish the intrinsic tradeoff between WGA and certain metrics. \figref{fig-appendix:tradeoff-acc-on-the-inverse-line} shows that both worst-case precision and ECE exhibit clear \emph{negative} correlation with WGA, demonstrating the fundamental tradeoff between WGA and several important metrics in subpopulation shift. These intriguing observations highlight the need for considering more realistic evaluation metrics in subpopulation shift beyond just using WGA.

\vspace{0.2cm}
\textbf{Accuracy not on the line.}
Finally, we display also other metrics that do not show either positive or negative correlation with WGA (\figref{fig-appendix:tradeoff-acc-not-on-the-line}). As observed, the correlation between these metrics and WGA shows inconsistent behavior across datasets. Interestingly, this phenomenon also indicates the potential bad performance on these metrics when merely optimizing for better WGA. We leave the exploration of other metrics and the rationale behind these behaviors for future work.

\subsection{Impact of Architecture, Pretraining Method, and Pretraining Dataset}
\label{appendix-subsec:arch-pretrain}

In this section, we examine the impact of model architecture and the source of the initial model weights on the worst group accuracy. Similar to the experiments above, we consider the following settings:
\begin{itemize}
    \item \textbf{Known Attributes.} Attributes are known in both training and validation, and validation set worst-group accuracy is used as the model selection criteria.
    \item \textbf{Unknown Attributes.} Attributes are unknown during training and validation. Following our findings in Sec.~\ref{subsec:model-select-attr-availability}, we use worst-class accuracy as the model selection criteria.
\end{itemize}

We experiment with ERM, JTT, and DFR as representative methods; \civilcomments as the representative text dataset, and \waterbirds, \chexpert, and \nicopp as representative image datasets.

For the text modality, we consider the following architectures and initial weights:
\begin{itemize}
    \item \textbf{BERT}\textsubscript{BASE} \cite{devlin2018bert}: A contextual language model based on the transformer architecture pretrained on BookCorpus and English Wikipedia data using the masked language model and next sentence prediction tasks.
    \item \textbf{SciBERT} \cite{beltagy2019scibert}: Same architecture as BERT\textsubscript{BASE}, but pretrained on scientific papers from Semantic Scholar, and has higher reported performance on scientific NLP tasks.
    \item \textbf{DistilBERT} \cite{sanh2019distilbert}: A knowledge distilled \cite{hinton2015distilling} version of BERT\textsubscript{BASE} with 40\% fewer parameters, pretrained using the same datasets as BERT\textsubscript{BASE}.
    \item \textbf{GPT-2} \cite{radford2019language}: An autoregressive language model based on the transformer decoder, pretrained using text from webpages upvoted on Reddit.
    \item \textbf{RoBERTa}\textsubscript{BASE} \cite{liu2019roberta}: Same architecture as BERT\textsubscript{BASE}, but pretrained with a more efficient procedure and using a collection of corpora much larger than BERT\textsubscript{BASE}.    
 \end{itemize}

For the image modality, we consider \textbf{ResNet-50}~\cite{he2016deep} and vision transformers (\textbf{ViT-B}) \cite{steiner2021train}. We consider model weights initialized with the following pretraining methods that span supervised and self-supervised manners:
\begin{itemize}
    \item \textbf{Supervised} pretraining \cite{kornblith2019better}.
    \item \textbf{SimCLR} \cite{chen2020simple}: Self-supervised contrastive pretraining using image augmentations.
    \item \textbf{Barlow} Twins \cite{zbontar2021barlow}: Self-supervised pretraining via redundancy reduction.
    \item \textbf{DINO} \cite{caron2021emerging}: Self-distillation with no labels.
    \item \textbf{CLIP} \cite{radford2021learning}: Using associated text as supervision. We select only the vision encoder.
\end{itemize}

We consider model weights initialized using the above pretraining methods on the following pretraining datasets:
\begin{itemize}
    \item \textbf{ImageNet-1K} \cite{deng2009imagenet}: 1.2 million images belonging to 1,000 classes, introduced as part of the ILSVRC2012 visual recognition challenge \cite{russakovsky2015imagenet}. 
    \item \textbf{ImageNet-21K} \cite{ridnik2021imagenet}: A superset of ImageNet-1K, consisting of 14 million images belonging to 21,841 classes.
    \item \textbf{SWAG} \cite{singh2022revisiting}: 3.6 billion images collected from public Instagram posts, weakly supervised using their associated hashtags.
    \item \textbf{LAION-2B} \cite{schuhmann2022laion}: 2.32 billion English image-text pairs constructed from Common Crawl.
    \item \textbf{OpenAI-CLIP} \cite{radford2021learning}: 400 million image-text pairs collected by OpenAI in training their CLIP model.
\end{itemize}

As model weights for many combinations of the above architectures, pretraining methods, and pretraining datasets are not available, we only experiment with the subset of combinations of weights that exist in public repositories.

Based on our experimental results on \civilcomments (Table \ref{tab-appendix:text_archs}), we find that BERT\textsubscript{BASE} is competitive in performance, even outperforming its successor RoBERTa\textsubscript{BASE} on many tasks. In addition, DistilBERT and GPT-2 exhibits much worse performance especially on ERM models.

Based on our experimental results on image datasets (Tables \ref{tab-appendix:image_archs_known} and \ref{tab-appendix:image_archs_unknown}), we find the following:
\begin{itemize}
    \item \textbf{Optimal architecture is dataset dependent.} Contrary to prior work \cite{paul2022vision}, we find mixed results when comparing the worst-group performance for ResNet and ViT-B. Specifically, ResNets seem to work better on \chexpert and \waterbirds, while vision transformers work better on \nicopp. 
    \item \textbf{Supervised pretraining outperforms others.} Similar to prior work \cite{izmailov2022feature}, we find that supervised pretraining outperforms self-supervised learning for the most part, though some self-supervised pretraining methods are still competitive. The results also warrant better self-supervised schemes for subgroup shifts \cite{yang2023simper}.
    \item \textbf{Larger pretraining datasets yield better results.} The biggest impact on worst-group accuracy by far appears to be the dataset on which the initial model weights are derived. This is especially true for \nicopp and \waterbirds, where going from ImageNet-1K to ImageNet-21K to SWAG almost always leads to a significant increase in worst-group accuracy, indicating that larger and more diverse pretraining datasets seem to increase performance. The effectiveness of SWAG-pretrained ViTs on \waterbirds has also been discussed in prior work \cite{mehta2022you}.
\end{itemize}
\vspace{-0.3cm}

\begin{table}[!htbp]
\setlength{\tabcolsep}{10pt}
\caption{Test-set worst-group accuracy on \civilcomments for different text architectures and pretraining methods.}
\vspace{4pt}
\label{tab-appendix:text_archs}
\small
\begin{center}
\begin{tabular}{lcccccc}
\toprule[1.5pt]
\multicolumn{1}{l}{\multirow{2.5}{*}{\textbf{Arch}}} & \multicolumn{3}{c}{\textbf{Unknown Attributes}} & \multicolumn{3}{c}{\textbf{Known Attributes}} \\
\cmidrule(lr){2-4} \cmidrule(lr){5-7}
 & ERM & JTT & DFR & ERM & JTT & DFR \\ \midrule
BERT       & \textbf{65.6}  & \textbf{69.6}  & 62.4          & \textbf{66.2} & 65.0          & \textbf{69.7} \\[1.2pt]
SciBERT    & 61.1           & 58.3           & \textbf{62.5} & 61.1          & 58.3          & 68.0 \\[1.2pt]
DistilBERT & 51.8           & 55.1           & 61.8          & 59.6          & 66.2          & 67.6  \\[1.2pt]
GPT-2      & 14.7           & 49.0           & 51.7          & 14.7          & 49.0          & 51.9 \\[1.2pt]
RoBERTa    & 61.0           & 58.0           & 61.6          & 63.1          & \textbf{66.7} & 68.2 \\
\bottomrule[1.5pt]
\end{tabular}
\end{center}
\vspace{-0.4cm}
\end{table}

\begin{table}[!htbp]
\setlength{\tabcolsep}{8pt}
\caption{Test-set worst-group accuracy for three image datasets with \textit{known attributes}, varying the model architecture and source of model initial weights. Best results of each column are in \textbf{bold} and the second best are \underline{underlined}.}
\vspace{-9pt}
\label{tab-appendix:image_archs_known}
\small
\begin{center}
\adjustbox{max width=\textwidth}{
\begin{tabular}{lllcccccccccr}
\toprule[1.5pt]
\multicolumn{1}{l}{\multirow{2.5}{*}{\textbf{Arch}}} & \multicolumn{1}{l}{\multirow{2.5}{*}{\textbf{Pretrain Method}}} & \multicolumn{1}{l}{\multirow{2.5}{*}{\textbf{Pretrain Dataset}}} & \multicolumn{3}{c}{\chexpert} & \multicolumn{3}{c}{\nicopp} & \multicolumn{3}{c}{\waterbirds} & \multicolumn{1}{l}{\multirow{2.5}{*}{\textbf{Avg}}} \\
\cmidrule(lr){4-6}  \cmidrule(lr){7-9}  \cmidrule(lr){10-12}
 & & & ERM & JTT & DFR & ERM & JTT & DFR & ERM & JTT & DFR &  \\ \midrule
\multirow{5}{*}{ResNet} & Barlow                   & ImageNet-1K               & 46.2          & 66.0          & \underline{74.7} & 40.0          & 40.0          & 20.0          & 67.3          & 72.4          & 88.3          & 57.2                                 \\
                        & DINO                     & ImageNet-1K               & 43.0          & 71.5          & 72.8          & 39.5          & 40.0          & 4.0           & 72.9          & 72.5          & 89.1          & 56.1                                 \\
                        & SimCLR                   & ImageNet-1K               & 47.9          & \textbf{72.3} & \textbf{74.8} & 30.0          & 30.0          & 16.0          & 70.1          & 68.1          & 81.2          & 54.5                                 \\
                        & Supervised               & ImageNet-1K               & \textbf{59.2} & 61.7          & 72.2          & 25.0          & 30.0          & 20.0          & \underline{76.5} & 74.3          & \textbf{90.2} & 56.6                                 \\
                        & Supervised               & ImageNet-21K              & \underline{51.4} & 68.0          & 70.0          & 40.0          & 46.0          & \textbf{40.0} & 74.5          & \underline{75.9} & \textbf{90.2} & \underline{61.8}                        \\ \midrule
\multirow{6}{*}{ViT-B}  & CLIP                     & Laion-2B                  & 49.2          & 58.5          & 69.1          & 33.3          & 40.0          & 33.3          & 39.6          & 46.9          & 75.5          & 49.5                                 \\
                        & CLIP                     & OpenAI-CLIP               & 42.2          & 55.8          & 68.8          & 33.3          & 40.0          & 30.0          & 40.4          & 40.4          & 78.2          & 47.7                                 \\
                        & DINO                     & ImageNet-1K               & 43.4          & \underline{71.8} & 72.4          & 30.0          & 40.0          & 32.0          & 63.9          & 64.6          & \textbf{90.2} & 56.5                                 \\
                        & Supervised               & ImageNet-1K               & 40.4          & 64.5          & 70.1          & 20.0          & 33.3          & 0.0           & 51.2          & 52.6          & 80.4          & 45.8                                 \\
                        & Supervised               & ImageNet-21K              & 47.5          & 69.1          & 69.1          & \underline{48.0} & \textbf{50.0} & 18.0          & 69.9          & 73.8          & 87.2          & 59.2                                 \\
                        & Supervised               & SWAG                      & 48.7          & 67.3          & 72.5          & \textbf{50.0} & \textbf{50.0} & \underline{34.0} & \textbf{82.7} & \textbf{81.2} & 87.5          & \textbf{63.8}                        \\
\bottomrule[1.5pt]
\end{tabular}
}
\end{center}
\vspace{-0.2cm}
\end{table}

\begin{table}[!htbp]
\setlength{\tabcolsep}{8pt}
\centering
\caption{Test-set worst-group accuracy for three image datasets with \textit{unknown attributes}, varying the model architecture and source of model initial weights. Best results of each column are in \textbf{bold} and the second best are \underline{underlined}.}
\vspace{-9pt}
\label{tab-appendix:image_archs_unknown}
\small
\begin{center}
\adjustbox{max width=\textwidth}{
\begin{tabular}{lllcccccccccr}
\toprule[1.5pt]
\multicolumn{1}{l}{\multirow{2.5}{*}{\textbf{Arch}}} & \multicolumn{1}{l}{\multirow{2.5}{*}{\textbf{Pretrain Method}}} & \multicolumn{1}{l}{\multirow{2.5}{*}{\textbf{Pretrain Dataset}}} & \multicolumn{3}{c}{\chexpert} & \multicolumn{3}{c}{\nicopp} & \multicolumn{3}{c}{\waterbirds} & \multicolumn{1}{l}{\multirow{2.5}{*}{\textbf{Avg}}} \\
\cmidrule(lr){4-6}  \cmidrule(lr){7-9}  \cmidrule(lr){10-12}
 & & & ERM & JTT & DFR & ERM & JTT & DFR & ERM & JTT & DFR &  \\ \midrule
\multirow{5}{*}{ResNet} & Barlow                   & ImageNet-1K               & 46.2          & 66.0          & 73.7          & 33.3          & 40.0          & \underline{40.0} & 67.3          & 72.4          & \underline{89.8} & 58.7                                 \\
                        & DINO                     & ImageNet-1K               & 43.0          & \underline{71.5} & 73.3          & 39.5          & 40.0          & 12.0          & 72.9          & 72.5          & 87.9          & 57.0                                 \\
                        & SimCLR                   & ImageNet-1K               & 47.9          & \textbf{72.3} & \underline{74.6} & 30.0          & 30.0          & 26.0          & 70.1          & 69.0          & 79.2          & 55.5                                 \\
                        & Supervised               & ImageNet-1K               & \textbf{59.2} & 61.7          & \textbf{75.4} & 40.0          & 30.0          & 33.3          & 67.0          & 74.3          & 89.6          & 58.9                                 \\
                        & Supervised               & ImageNet-21K              & 45.3          & 69.3          & 69.9          & 40.0          & 40.0          & \underline{40.0} & \underline{74.5} & \underline{75.9} & 88.3          & 60.4                                 \\ \midrule
\multirow{6}{*}{ViT-B}  & CLIP                     & Laion-2B                  & 49.2          & 58.5          & 69.7          & 30.0          & 30.0          & \underline{40.0} & 45.2          & 46.9          & 78.4          & 49.8                                 \\
                        & CLIP                     & OpenAI-CLIP               & 42.2          & 57.4          & 70.4          & 33.3          & 40.0          & \underline{40.0} & 26.5          & 44.4          & 77.4          & 48.0                                 \\
                        & DINO                     & ImageNet-1K               & 43.4          & 69.4          & 72.3          & 40.0          & 41.2          & 37.5          & 63.9          & 64.6          & \textbf{90.0} & 58.0                                 \\
                        & Supervised               & ImageNet-1K               & 40.4          & 69.5          & 71.5          & 33.3          & 33.3          & 16.7          & 49.4          & 52.6          & 81.2          & 49.8                                 \\
                        & Supervised               & ImageNet-21K              & 47.5          & 69.7          & 71.3          & \textbf{50.0} & \textbf{50.0} & 38.0          & 69.9          & 73.8          & 88.9          & \underline{62.1}                        \\
                        & Supervised               & SWAG                      & \underline{52.5} & 63.8          & 71.3          & \textbf{50.0} & \textbf{50.0} & \textbf{50.0} & \textbf{82.7} & \textbf{81.2} & 88.6          & \textbf{65.6}                        \\
\bottomrule[1.5pt]
\end{tabular}
}
\end{center}
\vspace{-0.4cm}
\end{table}

\newpage

\section{Complete Results}
\label{appendix-sec:complete-results}

We provide complete evaluation results in this section. As confirmed earlier, model selection and attribute availability play critical roles in subpopulation shift evaluation. To provide a thorough analysis, we investigate the following three settings:
\begin{Itemize}
    \item \textbf{Attributes are known in both training \& validation (Appendix~\ref{appendix-subsec:complete-results-oracle-selection}).} When attributes are known in both training and validation set, which corresponds to the most ideal scenario, we use ``\emph{test set worst-group accuracy}'' as an oracle selection method to identify the best possible performance for each algorithm.
    \vspace{8pt}
    \item \textbf{Attributes are unknown in training, but known in validation (Appendix~\ref{appendix-subsec:complete-results-val-known-selection}).} When attributes are still known in validation, we use ``\emph{validation set worst-group accuracy}'' to select models. We ignore algorithms that require attribute information in the training set (i.e., IRM, MMD, CORAL) when reporting results under this setting.
    \vspace{8pt}
    \item \textbf{Attributes are unknown in both training \& validation (Appendix~\ref{appendix-subsec:complete-results-val-unknown-selection}).} When attributes are completely unknown, we still use ``\emph{validation set worst-group accuracy}'' for model selection, which however degenerates to ``\emph{worst-class accuracy}''. We again ignore algorithms that require attribute information in the training set.
\end{Itemize}

\subsection{Attributes Known in Both Training \& Validation}
\label{appendix-subsec:complete-results-oracle-selection}

\subsubsection{Waterbirds}

\vspace{-10pt}
\begin{center}
\adjustbox{max width=\textwidth}{%
\begin{tabular}{lcccccccccc}
\toprule[1.5pt]
\textbf{Algorithm}         & \textbf{Avg Acc.}               & \textbf{Worst Acc.}             & \textbf{Avg Prec.}           & \textbf{Worst Prec.}         & \textbf{Avg F1}             & \textbf{Worst F1}           & \textbf{Adjusted Acc.}          & \textbf{Balanced Acc.}          & \textbf{AUROC}             & \textbf{ECE}               \\
\midrule
ERM                        & 84.1 \scriptsize$\pm1.7$   & 69.1 \scriptsize$\pm4.7$   & 77.4 \scriptsize$\pm2.0$   & 60.7 \scriptsize$\pm3.2$   & 79.4 \scriptsize$\pm2.1$   & 69.5 \scriptsize$\pm2.9$   & 83.1 \scriptsize$\pm2.0$   & 83.1 \scriptsize$\pm2.0$   & 91.0 \scriptsize$\pm1.4$   & 12.9 \scriptsize$\pm1.7$   \\
Mixup                      & 89.5 \scriptsize$\pm0.4$   & 78.2 \scriptsize$\pm0.4$   & 83.9 \scriptsize$\pm0.6$   & 71.6 \scriptsize$\pm1.3$   & 85.9 \scriptsize$\pm0.4$   & 78.8 \scriptsize$\pm0.6$   & 88.9 \scriptsize$\pm0.3$   & 88.9 \scriptsize$\pm0.3$   & 94.7 \scriptsize$\pm0.2$   & 7.0 \scriptsize$\pm0.6$    \\
GroupDRO                   & 88.8 \scriptsize$\pm1.8$   & 78.6 \scriptsize$\pm1.0$   & 83.6 \scriptsize$\pm2.4$   & 70.9 \scriptsize$\pm5.0$   & 85.3 \scriptsize$\pm2.0$   & 78.1 \scriptsize$\pm2.6$   & 88.5 \scriptsize$\pm0.8$   & 88.5 \scriptsize$\pm0.8$   & 95.5 \scriptsize$\pm0.5$   & 9.1 \scriptsize$\pm2.2$    \\
IRM                        & 88.4 \scriptsize$\pm0.1$   & 74.5 \scriptsize$\pm1.5$   & 82.5 \scriptsize$\pm0.2$   & 69.5 \scriptsize$\pm0.6$   & 84.3 \scriptsize$\pm0.1$   & 76.4 \scriptsize$\pm0.1$   & 87.1 \scriptsize$\pm0.3$   & 87.1 \scriptsize$\pm0.3$   & 94.0 \scriptsize$\pm0.3$   & 9.5 \scriptsize$\pm0.2$    \\
CVaRDRO                    & 89.8 \scriptsize$\pm0.4$   & 75.5 \scriptsize$\pm2.2$   & 84.5 \scriptsize$\pm0.7$   & 73.2 \scriptsize$\pm1.7$   & 86.1 \scriptsize$\pm0.3$   & 79.0 \scriptsize$\pm0.4$   & 88.5 \scriptsize$\pm0.3$   & 88.5 \scriptsize$\pm0.3$   & 95.4 \scriptsize$\pm0.2$   & 8.2 \scriptsize$\pm0.2$    \\
\end{tabular}}
\end{center}

\subsubsection{CelebA}

\vspace{-10pt}
\begin{center}
\adjustbox{max width=\textwidth}{%
\begin{tabular}{lcccccccccc}
\toprule[1.5pt]
\textbf{Algorithm}         & \textbf{Avg Acc.}               & \textbf{Worst Acc.}             & \textbf{Avg Prec.}           & \textbf{Worst Prec.}         & \textbf{Avg F1}             & \textbf{Worst F1}           & \textbf{Adjusted Acc.}          & \textbf{Balanced Acc.}          & \textbf{AUROC}             & \textbf{ECE}               \\
\midrule
ERM                        & 95.1 \scriptsize$\pm0.2$   & 62.6 \scriptsize$\pm1.5$   & 87.5 \scriptsize$\pm0.5$   & 76.4 \scriptsize$\pm1.2$   & 90.1 \scriptsize$\pm0.3$   & 83.1 \scriptsize$\pm0.5$   & 86.9 \scriptsize$\pm0.2$   & 93.4 \scriptsize$\pm0.2$   & 98.4 \scriptsize$\pm0.1$   & 3.3 \scriptsize$\pm0.5$    \\
Mixup                      & 95.4 \scriptsize$\pm0.1$   & 57.8 \scriptsize$\pm0.8$   & 88.4 \scriptsize$\pm0.3$   & 78.5 \scriptsize$\pm0.7$   & 90.6 \scriptsize$\pm0.2$   & 83.8 \scriptsize$\pm0.3$   & 85.8 \scriptsize$\pm0.2$   & 93.1 \scriptsize$\pm0.1$   & 98.4 \scriptsize$\pm0.1$   & 2.5 \scriptsize$\pm0.2$    \\
GroupDRO                   & 91.4 \scriptsize$\pm0.6$   & 89.0 \scriptsize$\pm0.7$   & 80.4 \scriptsize$\pm0.8$   & 61.5 \scriptsize$\pm1.7$   & 84.9 \scriptsize$\pm0.8$   & 74.9 \scriptsize$\pm1.2$   & 92.6 \scriptsize$\pm0.1$   & 93.3 \scriptsize$\pm0.2$   & 98.1 \scriptsize$\pm0.0$   & 8.0 \scriptsize$\pm0.9$    \\
IRM                        & 94.7 \scriptsize$\pm0.8$   & 63.0 \scriptsize$\pm2.5$   & 87.0 \scriptsize$\pm1.9$   & 75.3 \scriptsize$\pm3.9$   & 89.6 \scriptsize$\pm1.1$   & 82.2 \scriptsize$\pm1.8$   & 86.9 \scriptsize$\pm0.5$   & 93.3 \scriptsize$\pm0.3$   & 98.5 \scriptsize$\pm0.0$   & 3.4 \scriptsize$\pm1.3$    \\
CVaRDRO                    & 95.2 \scriptsize$\pm0.1$   & 64.1 \scriptsize$\pm2.8$   & 88.4 \scriptsize$\pm0.6$   & 78.6 \scriptsize$\pm1.4$   & 90.1 \scriptsize$\pm0.1$   & 83.0 \scriptsize$\pm0.2$   & 86.7 \scriptsize$\pm0.9$   & 92.2 \scriptsize$\pm0.7$   & 98.2 \scriptsize$\pm0.1$   & 2.6 \scriptsize$\pm0.3$    \\
\end{tabular}}
\end{center}

\subsubsection{CivilComments}

\vspace{-10pt}
\begin{center}
\adjustbox{max width=\textwidth}{%
\begin{tabular}{lcccccccccc}
\toprule[1.5pt]
\textbf{Algorithm}         & \textbf{Avg Acc.}               & \textbf{Worst Acc.}             & \textbf{Avg Prec.}           & \textbf{Worst Prec.}         & \textbf{Avg F1}             & \textbf{Worst F1}           & \textbf{Adjusted Acc.}          & \textbf{Balanced Acc.}          & \textbf{AUROC}             & \textbf{ECE}               \\
\midrule
ERM                        & 85.4 \scriptsize$\pm0.2$   & 63.7 \scriptsize$\pm1.1$   & 75.4 \scriptsize$\pm0.2$   & 57.8 \scriptsize$\pm0.6$   & 77.0 \scriptsize$\pm0.0$   & 63.1 \scriptsize$\pm0.1$   & 77.7 \scriptsize$\pm0.2$   & 79.2 \scriptsize$\pm0.3$   & 90.0 \scriptsize$\pm0.0$   & 8.1 \scriptsize$\pm0.2$    \\
Mixup                      & 84.9 \scriptsize$\pm0.3$   & 66.1 \scriptsize$\pm1.3$   & 74.8 \scriptsize$\pm0.4$   & 56.4 \scriptsize$\pm0.9$   & 76.6 \scriptsize$\pm0.2$   & 62.7 \scriptsize$\pm0.2$   & 77.9 \scriptsize$\pm0.3$   & 79.3 \scriptsize$\pm0.3$   & 89.7 \scriptsize$\pm0.0$   & 8.4 \scriptsize$\pm1.0$    \\
GroupDRO                   & 81.8 \scriptsize$\pm0.6$   & 70.6 \scriptsize$\pm1.2$   & 72.0 \scriptsize$\pm0.5$   & 49.6 \scriptsize$\pm1.1$   & 74.2 \scriptsize$\pm0.5$   & 60.3 \scriptsize$\pm0.6$   & 78.5 \scriptsize$\pm0.2$   & 79.9 \scriptsize$\pm0.2$   & 88.8 \scriptsize$\pm0.2$   & 12.2 \scriptsize$\pm0.9$   \\
IRM                        & 85.5 \scriptsize$\pm0.0$   & 63.2 \scriptsize$\pm0.8$   & 75.5 \scriptsize$\pm0.1$   & 57.8 \scriptsize$\pm0.2$   & 77.1 \scriptsize$\pm0.0$   & 63.3 \scriptsize$\pm0.1$   & 77.8 \scriptsize$\pm0.1$   & 79.4 \scriptsize$\pm0.1$   & 89.9 \scriptsize$\pm0.1$   & 7.4 \scriptsize$\pm0.6$    \\
CVaRDRO                    & 83.5 \scriptsize$\pm0.3$   & 68.7 \scriptsize$\pm1.3$   & 73.5 \scriptsize$\pm0.3$   & 52.8 \scriptsize$\pm0.6$   & 75.9 \scriptsize$\pm0.2$   & 62.4 \scriptsize$\pm0.2$   & 78.6 \scriptsize$\pm0.2$   & 80.7 \scriptsize$\pm0.1$   & 89.8 \scriptsize$\pm0.1$   & 32.9 \scriptsize$\pm0.4$   \\
\end{tabular}}
\end{center}

\subsubsection{MultiNLI}

\vspace{-10pt}
\begin{center}
\adjustbox{max width=\textwidth}{%
\begin{tabular}{lcccccccccc}
\toprule[1.5pt]
\textbf{Algorithm}         & \textbf{Avg Acc.}               & \textbf{Worst Acc.}             & \textbf{Avg Prec.}           & \textbf{Worst Prec.}         & \textbf{Avg F1}             & \textbf{Worst F1}           & \textbf{Adjusted Acc.}          & \textbf{Balanced Acc.}          & \textbf{AUROC}             & \textbf{ECE}               \\
\midrule
ERM                        & 80.9 \scriptsize$\pm0.1$   & 66.8 \scriptsize$\pm0.5$   & 81.1 \scriptsize$\pm0.1$   & 76.0 \scriptsize$\pm0.2$   & 80.9 \scriptsize$\pm0.1$   & 77.8 \scriptsize$\pm0.1$   & 79.7 \scriptsize$\pm0.0$   & 80.9 \scriptsize$\pm0.1$   & 93.6 \scriptsize$\pm0.1$   & 8.1 \scriptsize$\pm0.3$    \\
Mixup                      & 81.4 \scriptsize$\pm0.3$   & 68.5 \scriptsize$\pm0.6$   & 81.6 \scriptsize$\pm0.3$   & 76.0 \scriptsize$\pm0.5$   & 81.4 \scriptsize$\pm0.3$   & 78.0 \scriptsize$\pm0.2$   & 80.1 \scriptsize$\pm0.3$   & 81.4 \scriptsize$\pm0.3$   & 93.6 \scriptsize$\pm0.1$   & 9.4 \scriptsize$\pm0.9$    \\
GroupDRO                   & 81.1 \scriptsize$\pm0.3$   & 76.0 \scriptsize$\pm0.7$   & 81.4 \scriptsize$\pm0.3$   & 74.7 \scriptsize$\pm0.5$   & 81.1 \scriptsize$\pm0.3$   & 77.8 \scriptsize$\pm0.1$   & 80.8 \scriptsize$\pm0.3$   & 81.1 \scriptsize$\pm0.3$   & 93.7 \scriptsize$\pm0.1$   & 9.8 \scriptsize$\pm1.0$    \\
IRM                        & 77.8 \scriptsize$\pm0.6$   & 63.6 \scriptsize$\pm1.3$   & 78.3 \scriptsize$\pm0.5$   & 71.0 \scriptsize$\pm1.1$   & 77.9 \scriptsize$\pm0.6$   & 74.8 \scriptsize$\pm0.4$   & 76.6 \scriptsize$\pm0.5$   & 77.8 \scriptsize$\pm0.6$   & 91.5 \scriptsize$\pm0.3$   & 11.2 \scriptsize$\pm1.8$   \\
CVaRDRO                    & 75.1 \scriptsize$\pm0.1$   & 63.0 \scriptsize$\pm1.5$   & 76.2 \scriptsize$\pm0.2$   & 65.6 \scriptsize$\pm0.2$   & 75.2 \scriptsize$\pm0.1$   & 72.1 \scriptsize$\pm0.2$   & 74.2 \scriptsize$\pm0.4$   & 75.1 \scriptsize$\pm0.1$   & 86.3 \scriptsize$\pm0.2$   & 41.4 \scriptsize$\pm0.1$   \\
\end{tabular}}
\end{center}

\subsubsection{MetaShift}

\vspace{-10pt}
\begin{center}
\adjustbox{max width=\textwidth}{%
\begin{tabular}{lcccccccccc}
\toprule[1.5pt]
\textbf{Algorithm}         & \textbf{Avg Acc.}               & \textbf{Worst Acc.}             & \textbf{Avg Prec.}           & \textbf{Worst Prec.}         & \textbf{Avg F1}             & \textbf{Worst F1}           & \textbf{Adjusted Acc.}          & \textbf{Balanced Acc.}          & \textbf{AUROC}             & \textbf{ECE}               \\
\midrule
ERM                        & 91.3 \scriptsize$\pm0.3$   & 82.6 \scriptsize$\pm0.4$   & 91.2 \scriptsize$\pm0.3$   & 90.6 \scriptsize$\pm0.4$   & 91.2 \scriptsize$\pm0.3$   & 90.7 \scriptsize$\pm0.3$   & 89.3 \scriptsize$\pm0.3$   & 91.2 \scriptsize$\pm0.3$   & 97.3 \scriptsize$\pm0.2$   & 6.3 \scriptsize$\pm0.9$    \\
Mixup                      & 91.6 \scriptsize$\pm0.3$   & 81.0 \scriptsize$\pm0.8$   & 91.7 \scriptsize$\pm0.3$   & 90.6 \scriptsize$\pm0.2$   & 91.6 \scriptsize$\pm0.3$   & 91.0 \scriptsize$\pm0.3$   & 89.4 \scriptsize$\pm0.3$   & 91.6 \scriptsize$\pm0.3$   & 97.3 \scriptsize$\pm0.1$   & 2.3 \scriptsize$\pm0.1$    \\
GroupDRO                   & 91.0 \scriptsize$\pm0.1$   & 85.6 \scriptsize$\pm0.4$   & 90.9 \scriptsize$\pm0.1$   & 90.0 \scriptsize$\pm0.6$   & 90.9 \scriptsize$\pm0.1$   & 90.4 \scriptsize$\pm0.0$   & 89.8 \scriptsize$\pm0.1$   & 91.0 \scriptsize$\pm0.1$   & 97.5 \scriptsize$\pm0.0$   & 3.2 \scriptsize$\pm0.5$    \\
IRM                        & 91.8 \scriptsize$\pm0.4$   & 83.0 \scriptsize$\pm0.1$   & 91.8 \scriptsize$\pm0.4$   & 90.5 \scriptsize$\pm1.0$   & 91.7 \scriptsize$\pm0.4$   & 91.3 \scriptsize$\pm0.4$   & 89.7 \scriptsize$\pm0.5$   & 91.7 \scriptsize$\pm0.4$   & 97.6 \scriptsize$\pm0.2$   & 5.3 \scriptsize$\pm0.2$    \\
CVaRDRO                    & 92.1 \scriptsize$\pm0.2$   & 84.6 \scriptsize$\pm0.0$   & 92.1 \scriptsize$\pm0.2$   & 90.8 \scriptsize$\pm0.6$   & 92.1 \scriptsize$\pm0.2$   & 91.6 \scriptsize$\pm0.2$   & 90.4 \scriptsize$\pm0.2$   & 92.1 \scriptsize$\pm0.2$   & 97.7 \scriptsize$\pm0.0$   & 4.9 \scriptsize$\pm0.3$    \\
\end{tabular}}
\end{center}

\subsubsection{ImageNetBG}

\vspace{-10pt}
\begin{center}
\adjustbox{max width=\textwidth}{%
\begin{tabular}{lcccccccccc}
\toprule[1.5pt]
\textbf{Algorithm}         & \textbf{Avg Acc.}               & \textbf{Worst Acc.}             & \textbf{Avg Prec.}           & \textbf{Worst Prec.}         & \textbf{Avg F1}             & \textbf{Worst F1}           & \textbf{Adjusted Acc.}          & \textbf{Balanced Acc.}          & \textbf{AUROC}             & \textbf{ECE}               \\
\midrule
ERM                        & 88.4 \scriptsize$\pm0.1$   & 81.0 \scriptsize$\pm0.9$   & 88.5 \scriptsize$\pm0.1$   & 80.4 \scriptsize$\pm0.1$   & 88.4 \scriptsize$\pm0.1$   & 81.3 \scriptsize$\pm0.1$   & 88.4 \scriptsize$\pm0.1$   & 88.4 \scriptsize$\pm0.1$   & 99.0 \scriptsize$\pm0.0$   & 5.8 \scriptsize$\pm0.2$    \\
Mixup                      & 88.5 \scriptsize$\pm0.3$   & 82.4 \scriptsize$\pm0.3$   & 88.7 \scriptsize$\pm0.2$   & 80.4 \scriptsize$\pm1.3$   & 88.5 \scriptsize$\pm0.3$   & 81.9 \scriptsize$\pm0.8$   & 88.5 \scriptsize$\pm0.3$   & 88.5 \scriptsize$\pm0.3$   & 98.8 \scriptsize$\pm0.1$   & 3.3 \scriptsize$\pm1.0$    \\
GroupDRO                   & 87.3 \scriptsize$\pm0.2$   & 78.2 \scriptsize$\pm0.3$   & 87.5 \scriptsize$\pm0.2$   & 77.9 \scriptsize$\pm1.1$   & 87.3 \scriptsize$\pm0.2$   & 80.0 \scriptsize$\pm0.9$   & 87.3 \scriptsize$\pm0.2$   & 87.3 \scriptsize$\pm0.2$   & 98.9 \scriptsize$\pm0.0$   & 4.3 \scriptsize$\pm0.5$    \\
IRM                        & 88.7 \scriptsize$\pm0.1$   & 81.3 \scriptsize$\pm0.3$   & 88.8 \scriptsize$\pm0.1$   & 81.6 \scriptsize$\pm0.2$   & 88.7 \scriptsize$\pm0.1$   & 81.7 \scriptsize$\pm0.1$   & 88.7 \scriptsize$\pm0.1$   & 88.7 \scriptsize$\pm0.1$   & 99.1 \scriptsize$\pm0.0$   & 5.2 \scriptsize$\pm0.1$    \\
CVaRDRO                    & 88.2 \scriptsize$\pm0.1$   & 80.7 \scriptsize$\pm1.1$   & 88.4 \scriptsize$\pm0.1$   & 78.6 \scriptsize$\pm1.9$   & 88.3 \scriptsize$\pm0.1$   & 80.7 \scriptsize$\pm0.5$   & 88.2 \scriptsize$\pm0.1$   & 88.2 \scriptsize$\pm0.1$   & 99.0 \scriptsize$\pm0.0$   & 4.9 \scriptsize$\pm0.4$    \\
\end{tabular}}
\end{center}

\subsubsection{NICO++}

\vspace{-10pt}
\begin{center}
\adjustbox{max width=\textwidth}{%
\begin{tabular}{lcccccccccc}
\toprule[1.5pt]
\textbf{Algorithm}         & \textbf{Avg Acc.}               & \textbf{Worst Acc.}             & \textbf{Avg Prec.}           & \textbf{Worst Prec.}         & \textbf{Avg F1}             & \textbf{Worst F1}           & \textbf{Adjusted Acc.}          & \textbf{Balanced Acc.}          & \textbf{AUROC}             & \textbf{ECE}               \\
\midrule
ERM                        & 84.5 \scriptsize$\pm0.5$   & 37.6 \scriptsize$\pm2.0$   & 85.5 \scriptsize$\pm0.3$   & 54.5 \scriptsize$\pm2.8$   & 84.6 \scriptsize$\pm0.4$   & 65.8 \scriptsize$\pm1.1$   & 84.0 \scriptsize$\pm0.5$   & 84.3 \scriptsize$\pm0.5$   & 99.3 \scriptsize$\pm0.0$   & 10.4 \scriptsize$\pm0.1$   \\
Mixup                      & 84.0 \scriptsize$\pm0.6$   & 42.7 \scriptsize$\pm1.4$   & 85.2 \scriptsize$\pm0.5$   & 53.0 \scriptsize$\pm1.6$   & 84.2 \scriptsize$\pm0.6$   & 63.4 \scriptsize$\pm1.1$   & 83.7 \scriptsize$\pm0.6$   & 83.9 \scriptsize$\pm0.6$   & 99.3 \scriptsize$\pm0.0$   & 2.5 \scriptsize$\pm1.0$    \\
GroupDRO                   & 83.2 \scriptsize$\pm0.4$   & 37.8 \scriptsize$\pm1.8$   & 84.5 \scriptsize$\pm0.4$   & 55.5 \scriptsize$\pm1.0$   & 83.3 \scriptsize$\pm0.4$   & 63.6 \scriptsize$\pm0.6$   & 82.7 \scriptsize$\pm0.4$   & 83.0 \scriptsize$\pm0.4$   & 99.3 \scriptsize$\pm0.0$   & 8.7 \scriptsize$\pm0.6$    \\
IRM                        & 84.4 \scriptsize$\pm0.7$   & 40.0 \scriptsize$\pm0.0$   & 85.1 \scriptsize$\pm0.5$   & 63.0 \scriptsize$\pm2.0$   & 84.4 \scriptsize$\pm0.6$   & 65.9 \scriptsize$\pm1.2$   & 83.9 \scriptsize$\pm0.7$   & 84.3 \scriptsize$\pm0.7$   & 99.4 \scriptsize$\pm0.0$   & 7.0 \scriptsize$\pm1.4$    \\
CVaRDRO                    & 83.6 \scriptsize$\pm0.6$   & 36.7 \scriptsize$\pm2.7$   & 85.0 \scriptsize$\pm0.4$   & 55.7 \scriptsize$\pm2.3$   & 83.8 \scriptsize$\pm0.6$   & 64.3 \scriptsize$\pm1.5$   & 83.2 \scriptsize$\pm0.6$   & 83.5 \scriptsize$\pm0.6$   & 99.4 \scriptsize$\pm0.0$   & 7.9 \scriptsize$\pm1.1$    \\
\end{tabular}}
\end{center}

\subsubsection{MIMIC-CXR}

\vspace{-10pt}
\begin{center}
\adjustbox{max width=\textwidth}{%
\begin{tabular}{lcccccccccc}
\toprule[1.5pt]
\textbf{Algorithm}         & \textbf{Avg Acc.}               & \textbf{Worst Acc.}             & \textbf{Avg Prec.}           & \textbf{Worst Prec.}         & \textbf{Avg F1}             & \textbf{Worst F1}           & \textbf{Adjusted Acc.}          & \textbf{Balanced Acc.}          & \textbf{AUROC}             & \textbf{ECE}               \\
\midrule
ERM                        & 78.2 \scriptsize$\pm0.1$   & 68.9 \scriptsize$\pm0.3$   & 77.3 \scriptsize$\pm0.1$   & 71.1 \scriptsize$\pm0.1$   & 77.5 \scriptsize$\pm0.1$   & 73.6 \scriptsize$\pm0.1$   & 77.2 \scriptsize$\pm0.0$   & 77.8 \scriptsize$\pm0.1$   & 85.2 \scriptsize$\pm0.1$   & 3.4 \scriptsize$\pm0.2$    \\
Mixup                      & 78.3 \scriptsize$\pm0.0$   & 68.1 \scriptsize$\pm0.9$   & 77.4 \scriptsize$\pm0.0$   & 71.6 \scriptsize$\pm0.2$   & 77.5 \scriptsize$\pm0.0$   & 73.4 \scriptsize$\pm0.1$   & 77.2 \scriptsize$\pm0.1$   & 77.8 \scriptsize$\pm0.1$   & 85.1 \scriptsize$\pm0.1$   & 3.6 \scriptsize$\pm0.2$    \\
GroupDRO                   & 76.9 \scriptsize$\pm0.3$   & 74.4 \scriptsize$\pm0.2$   & 76.1 \scriptsize$\pm0.2$   & 68.7 \scriptsize$\pm0.5$   & 76.3 \scriptsize$\pm0.2$   & 72.7 \scriptsize$\pm0.1$   & 76.7 \scriptsize$\pm0.1$   & 76.9 \scriptsize$\pm0.2$   & 83.7 \scriptsize$\pm0.1$   & 4.7 \scriptsize$\pm0.1$    \\
IRM                        & 78.2 \scriptsize$\pm0.0$   & 67.7 \scriptsize$\pm0.2$   & 77.3 \scriptsize$\pm0.0$   & 71.4 \scriptsize$\pm0.1$   & 77.5 \scriptsize$\pm0.0$   & 73.5 \scriptsize$\pm0.1$   & 77.2 \scriptsize$\pm0.1$   & 77.8 \scriptsize$\pm0.1$   & 85.2 \scriptsize$\pm0.1$   & 3.4 \scriptsize$\pm0.2$    \\
CVaRDRO                    & 78.3 \scriptsize$\pm0.1$   & 68.6 \scriptsize$\pm0.4$   & 77.4 \scriptsize$\pm0.1$   & 71.1 \scriptsize$\pm0.3$   & 77.7 \scriptsize$\pm0.1$   & 73.9 \scriptsize$\pm0.0$   & 77.4 \scriptsize$\pm0.0$   & 78.1 \scriptsize$\pm0.0$   & 85.1 \scriptsize$\pm0.0$   & 7.8 \scriptsize$\pm0.3$    \\
\end{tabular}}
\end{center}

\subsubsection{MIMICNotes}

\vspace{-10pt}
\begin{center}
\adjustbox{max width=\textwidth}{%
\begin{tabular}{lcccccccccc}
\toprule[1.5pt]
\textbf{Algorithm}         & \textbf{Avg Acc.}               & \textbf{Worst Acc.}             & \textbf{Avg Prec.}           & \textbf{Worst Prec.}         & \textbf{Avg F1}             & \textbf{Worst F1}           & \textbf{Adjusted Acc.}          & \textbf{Balanced Acc.}          & \textbf{AUROC}             & \textbf{ECE}               \\
\midrule
ERM                        & 91.1 \scriptsize$\pm0.1$   & 18.7 \scriptsize$\pm2.7$   & 77.6 \scriptsize$\pm0.9$   & 63.1 \scriptsize$\pm2.0$   & 63.2 \scriptsize$\pm1.6$   & 31.2 \scriptsize$\pm3.1$   & 59.7 \scriptsize$\pm1.3$   & 59.9 \scriptsize$\pm1.3$   & 85.3 \scriptsize$\pm0.1$   & 2.1 \scriptsize$\pm0.8$    \\
Mixup                      & 91.1 \scriptsize$\pm0.0$   & 22.7 \scriptsize$\pm3.2$   & 76.8 \scriptsize$\pm0.7$   & 61.2 \scriptsize$\pm1.6$   & 65.1 \scriptsize$\pm1.6$   & 35.0 \scriptsize$\pm3.2$   & 61.5 \scriptsize$\pm1.6$   & 61.7 \scriptsize$\pm1.7$   & 85.4 \scriptsize$\pm0.0$   & 2.0 \scriptsize$\pm0.8$    \\
GroupDRO                   & 76.1 \scriptsize$\pm0.7$   & 72.6 \scriptsize$\pm0.5$   & 61.3 \scriptsize$\pm0.1$   & 25.7 \scriptsize$\pm0.4$   & 61.8 \scriptsize$\pm0.4$   & 38.6 \scriptsize$\pm0.3$   & 76.2 \scriptsize$\pm0.3$   & 76.5 \scriptsize$\pm0.2$   & 85.0 \scriptsize$\pm0.1$   & 22.2 \scriptsize$\pm0.6$   \\
IRM                        & 91.0 \scriptsize$\pm0.0$   & 22.5 \scriptsize$\pm2.5$   & 76.3 \scriptsize$\pm0.5$   & 60.1 \scriptsize$\pm1.2$   & 65.2 \scriptsize$\pm1.2$   & 35.3 \scriptsize$\pm2.4$   & 61.5 \scriptsize$\pm1.2$   & 61.7 \scriptsize$\pm1.2$   & 85.3 \scriptsize$\pm0.0$   & 1.9 \scriptsize$\pm0.2$    \\
CVaRDRO                    & 90.9 \scriptsize$\pm0.1$   & 23.0 \scriptsize$\pm4.6$   & 76.5 \scriptsize$\pm1.2$   & 60.6 \scriptsize$\pm2.9$   & 64.6 \scriptsize$\pm2.4$   & 34.0 \scriptsize$\pm4.8$   & 61.4 \scriptsize$\pm2.3$   & 61.6 \scriptsize$\pm2.4$   & 85.1 \scriptsize$\pm0.1$   & 4.2 \scriptsize$\pm1.6$    \\
\end{tabular}}
\end{center}

\subsubsection{CXRMultisite}

\vspace{-10pt}
\begin{center}
\adjustbox{max width=\textwidth}{%
\begin{tabular}{lcccccccccc}
\toprule[1.5pt]
\textbf{Algorithm}         & \textbf{Avg Acc.}               & \textbf{Worst Acc.}             & \textbf{Avg Prec.}           & \textbf{Worst Prec.}         & \textbf{Avg F1}             & \textbf{Worst F1}           & \textbf{Adjusted Acc.}          & \textbf{Balanced Acc.}          & \textbf{AUROC}             & \textbf{ECE}               \\
\midrule
ERM                        & 98.3 \scriptsize$\pm0.0$   & 0.0 \scriptsize$\pm0.0$    & 49.2 \scriptsize$\pm0.0$   & 0.0 \scriptsize$\pm0.0$    & 49.6 \scriptsize$\pm0.0$   & 0.0 \scriptsize$\pm0.0$    & 50.0 \scriptsize$\pm0.0$   & 50.0 \scriptsize$\pm0.0$   & 93.1 \scriptsize$\pm0.1$   & 0.3 \scriptsize$\pm0.1$    \\
Mixup                      & 98.3 \scriptsize$\pm0.0$   & 0.0 \scriptsize$\pm0.0$    & 49.2 \scriptsize$\pm0.0$   & 0.0 \scriptsize$\pm0.0$    & 49.6 \scriptsize$\pm0.0$   & 0.0 \scriptsize$\pm0.0$    & 50.0 \scriptsize$\pm0.0$   & 50.0 \scriptsize$\pm0.0$   & 92.9 \scriptsize$\pm0.1$   & 0.3 \scriptsize$\pm0.0$    \\
GroupDRO                   & 84.0 \scriptsize$\pm10.9$  & 19.3 \scriptsize$\pm13.7$  & 55.7 \scriptsize$\pm2.0$   & 12.7 \scriptsize$\pm4.1$   & 51.2 \scriptsize$\pm5.0$   & 12.4 \scriptsize$\pm2.9$   & 55.9 \scriptsize$\pm2.5$   & 59.7 \scriptsize$\pm2.5$   & 79.4 \scriptsize$\pm2.9$   & 29.3 \scriptsize$\pm6.3$   \\
IRM                        & 77.5 \scriptsize$\pm17.0$  & 8.8 \scriptsize$\pm7.2$    & 49.6 \scriptsize$\pm0.3$   & 0.7 \scriptsize$\pm0.5$    & 42.4 \scriptsize$\pm5.9$   & 1.3 \scriptsize$\pm1.0$    & 51.1 \scriptsize$\pm0.9$   & 51.8 \scriptsize$\pm1.5$   & 64.2 \scriptsize$\pm7.3$   & 47.3 \scriptsize$\pm1.1$   \\
CVaRDRO                    & 98.3 \scriptsize$\pm0.0$   & 0.0 \scriptsize$\pm0.0$    & 61.2 \scriptsize$\pm4.9$   & 24.0 \scriptsize$\pm9.8$   & 50.7 \scriptsize$\pm0.7$   & 2.2 \scriptsize$\pm1.5$    & 50.2 \scriptsize$\pm0.2$   & 50.6 \scriptsize$\pm0.4$   & 93.0 \scriptsize$\pm0.0$   & 0.9 \scriptsize$\pm0.3$    \\
\end{tabular}}
\end{center}

\subsubsection{CheXpert}

\vspace{-10pt}
\begin{center}
\adjustbox{max width=\textwidth}{%
\begin{tabular}{lcccccccccc}
\toprule[1.5pt]
\textbf{Algorithm}         & \textbf{Avg Acc.}               & \textbf{Worst Acc.}             & \textbf{Avg Prec.}           & \textbf{Worst Prec.}         & \textbf{Avg F1}             & \textbf{Worst F1}           & \textbf{Adjusted Acc.}          & \textbf{Balanced Acc.}          & \textbf{AUROC}             & \textbf{ECE}               \\
\midrule
ERM                        & 86.9 \scriptsize$\pm0.5$   & 50.2 \scriptsize$\pm3.8$   & 66.3 \scriptsize$\pm0.3$   & 37.6 \scriptsize$\pm0.8$   & 68.6 \scriptsize$\pm0.2$   & 44.6 \scriptsize$\pm0.5$   & 72.5 \scriptsize$\pm1.1$   & 72.8 \scriptsize$\pm1.1$   & 84.8 \scriptsize$\pm0.3$   & 8.6 \scriptsize$\pm0.4$    \\
Mixup                      & 81.9 \scriptsize$\pm6.2$   & 37.4 \scriptsize$\pm3.5$   & 63.5 \scriptsize$\pm5.0$   & 33.9 \scriptsize$\pm9.3$   & 62.5 \scriptsize$\pm5.9$   & 35.7 \scriptsize$\pm7.6$   & 63.8 \scriptsize$\pm4.7$   & 64.1 \scriptsize$\pm4.6$   & 76.1 \scriptsize$\pm8.5$   & 16.1 \scriptsize$\pm9.0$   \\
GroupDRO                   & 78.9 \scriptsize$\pm0.3$   & 74.5 \scriptsize$\pm0.2$   & 62.8 \scriptsize$\pm0.1$   & 28.3 \scriptsize$\pm0.3$   & 64.4 \scriptsize$\pm0.2$   & 41.7 \scriptsize$\pm0.3$   & 78.4 \scriptsize$\pm0.1$   & 79.0 \scriptsize$\pm0.1$   & 86.0 \scriptsize$\pm0.1$   & 21.1 \scriptsize$\pm1.0$   \\
IRM                        & 89.8 \scriptsize$\pm0.3$   & 34.4 \scriptsize$\pm1.7$   & 70.1 \scriptsize$\pm0.7$   & 46.5 \scriptsize$\pm1.6$   & 68.6 \scriptsize$\pm0.1$   & 42.7 \scriptsize$\pm0.3$   & 67.5 \scriptsize$\pm0.6$   & 67.5 \scriptsize$\pm0.6$   & 85.8 \scriptsize$\pm0.3$   & 4.4 \scriptsize$\pm1.2$    \\
CVaRDRO                    & 66.2 \scriptsize$\pm2.7$   & 57.9 \scriptsize$\pm0.4$   & 56.4 \scriptsize$\pm0.5$   & 17.7 \scriptsize$\pm1.0$   & 52.9 \scriptsize$\pm1.6$   & 27.9 \scriptsize$\pm1.1$   & 66.1 \scriptsize$\pm0.6$   & 67.0 \scriptsize$\pm0.6$   & 73.0 \scriptsize$\pm0.6$   & 40.4 \scriptsize$\pm0.0$   \\
\end{tabular}}
\end{center}

\subsubsection{Living17}

\vspace{-10pt}
\begin{center}
\adjustbox{max width=\textwidth}{%
\begin{tabular}{lcccccccccc}
\toprule[1.5pt]
\textbf{Algorithm}         & \textbf{Avg Acc.}               & \textbf{Worst Acc.}             & \textbf{Avg Prec.}           & \textbf{Worst Prec.}         & \textbf{Avg F1}             & \textbf{Worst F1}           & \textbf{Adjusted Acc.}          & \textbf{Balanced Acc.}          & \textbf{AUROC}             & \textbf{ECE}               \\
\midrule
ERM                        & 28.2 \scriptsize$\pm1.5$   & 8.7 \scriptsize$\pm1.0$    & 29.5 \scriptsize$\pm1.1$   & 9.0 \scriptsize$\pm1.2$    & 27.8 \scriptsize$\pm1.3$   & 10.0 \scriptsize$\pm1.5$   & 28.2 \scriptsize$\pm1.5$   & 28.2 \scriptsize$\pm1.5$   & 77.9 \scriptsize$\pm1.2$   & 53.5 \scriptsize$\pm2.5$   \\
Mixup                      & 29.8 \scriptsize$\pm1.8$   & 9.3 \scriptsize$\pm1.4$    & 32.5 \scriptsize$\pm2.3$   & 9.5 \scriptsize$\pm1.3$    & 29.8 \scriptsize$\pm1.8$   & 10.1 \scriptsize$\pm1.5$   & 29.8 \scriptsize$\pm1.8$   & 29.8 \scriptsize$\pm1.8$   & 78.3 \scriptsize$\pm1.2$   & 34.6 \scriptsize$\pm1.7$   \\
GroupDRO                   & 27.2 \scriptsize$\pm1.5$   & 9.7 \scriptsize$\pm0.7$    & 29.8 \scriptsize$\pm0.7$   & 7.8 \scriptsize$\pm0.9$    & 27.3 \scriptsize$\pm1.1$   & 9.1 \scriptsize$\pm0.8$    & 27.2 \scriptsize$\pm1.5$   & 27.2 \scriptsize$\pm1.5$   & 77.8 \scriptsize$\pm0.6$   & 55.8 \scriptsize$\pm0.5$   \\
IRM                        & 28.2 \scriptsize$\pm1.5$   & 8.7 \scriptsize$\pm1.0$    & 29.5 \scriptsize$\pm1.1$   & 9.0 \scriptsize$\pm1.2$    & 27.8 \scriptsize$\pm1.3$   & 10.0 \scriptsize$\pm1.5$   & 28.2 \scriptsize$\pm1.5$   & 28.2 \scriptsize$\pm1.5$   & 77.9 \scriptsize$\pm1.2$   & 53.5 \scriptsize$\pm2.5$   \\
CVaRDRO                    & 28.3 \scriptsize$\pm0.7$   & 8.3 \scriptsize$\pm0.7$    & 30.0 \scriptsize$\pm0.7$   & 7.9 \scriptsize$\pm0.2$    & 27.9 \scriptsize$\pm0.8$   & 8.0 \scriptsize$\pm0.2$    & 28.3 \scriptsize$\pm0.7$   & 28.3 \scriptsize$\pm0.7$   & 81.0 \scriptsize$\pm0.1$   & 33.2 \scriptsize$\pm4.1$   \\
\end{tabular}}
\end{center}

\subsubsection{Overall}
\label{appendix-subsubsec:overall-table-oracle-selection}

\begin{center}
\adjustbox{max width=\textwidth}{%
\begin{tabular}{lcccccccccccc|c}
\toprule[1.5pt]
\textbf{Algorithm}         & \waterbirds        & \celeba            & \civilcomments & \multinli          & \metashift         & \imagenetbg        & \nicopp            & \mimiccxr    & \mimicnotes        & \cxrmultisite      & \chexpert & \living          & \textbf{Avg}               \\
\midrule
ERM                        & 69.1 \scriptsize$\pm4.7$   & 62.6 \scriptsize$\pm1.5$   & 63.7 \scriptsize$\pm1.1$   & 66.8 \scriptsize$\pm0.5$   & 82.6 \scriptsize$\pm0.4$   & 81.0 \scriptsize$\pm0.9$   & 37.6 \scriptsize$\pm2.0$   & 68.9 \scriptsize$\pm0.3$   & 83.1 \scriptsize$\pm0.1$   & 50.1 \scriptsize$\pm0.9$   & 50.2 \scriptsize$\pm3.8$   & 28.2 \scriptsize$\pm1.5$   & 62.0                       \\
Mixup                      & 78.2 \scriptsize$\pm0.4$   & 57.8 \scriptsize$\pm0.8$   & 66.1 \scriptsize$\pm1.3$   & 68.5 \scriptsize$\pm0.6$   & 81.0 \scriptsize$\pm0.8$   & 82.4 \scriptsize$\pm0.3$   & 42.7 \scriptsize$\pm1.4$   & 68.1 \scriptsize$\pm0.9$   & 82.0 \scriptsize$\pm0.4$   & 50.1 \scriptsize$\pm0.9$   & 37.4 \scriptsize$\pm3.5$   & 29.8 \scriptsize$\pm1.8$   & 62.0                       \\
GroupDRO                   & 78.6 \scriptsize$\pm1.0$   & 89.0 \scriptsize$\pm0.7$   & 70.6 \scriptsize$\pm1.2$   & 76.0 \scriptsize$\pm0.7$   & 85.6 \scriptsize$\pm0.4$   & 78.2 \scriptsize$\pm0.3$   & 37.8 \scriptsize$\pm1.8$   & 74.4 \scriptsize$\pm0.2$   & 83.7 \scriptsize$\pm0.1$   & 59.6 \scriptsize$\pm1.0$   & 74.5 \scriptsize$\pm0.2$   & 27.2 \scriptsize$\pm1.5$   & 69.6                       \\
IRM                        & 74.5 \scriptsize$\pm1.5$   & 63.0 \scriptsize$\pm2.5$   & 63.2 \scriptsize$\pm0.8$   & 63.6 \scriptsize$\pm1.3$   & 83.0 \scriptsize$\pm0.1$   & 81.3 \scriptsize$\pm0.3$   & 40.0 \scriptsize$\pm0.0$   & 67.7 \scriptsize$\pm0.2$   & 83.2 \scriptsize$\pm0.2$   & 47.9 \scriptsize$\pm1.1$   & 34.4 \scriptsize$\pm1.7$   & 28.2 \scriptsize$\pm1.5$   & 60.8                       \\
\end{tabular}}
\end{center}

\vspace{0.1cm}
\subsection{Attributes Unknown in Training, but Known in Validation}
\label{appendix-subsec:complete-results-val-known-selection}

\subsubsection{Waterbirds}

\vspace{-10pt}
\begin{center}
\adjustbox{max width=\textwidth}{%
\begin{tabular}{lcccccccccc}
\toprule[1.5pt]
\textbf{Algorithm}         & \textbf{Avg Acc.}               & \textbf{Worst Acc.}             & \textbf{Avg Prec.}           & \textbf{Worst Prec.}         & \textbf{Avg F1}             & \textbf{Worst F1}           & \textbf{Adjusted Acc.}          & \textbf{Balanced Acc.}          & \textbf{AUROC}             & \textbf{ECE}               \\
\midrule
ERM                        & 84.1 \scriptsize$\pm1.7$   & 69.1 \scriptsize$\pm4.7$   & 77.4 \scriptsize$\pm2.0$   & 60.7 \scriptsize$\pm3.2$   & 79.4 \scriptsize$\pm2.1$   & 69.5 \scriptsize$\pm2.9$   & 83.1 \scriptsize$\pm2.0$   & 83.1 \scriptsize$\pm2.0$   & 91.0 \scriptsize$\pm1.4$   & 12.9 \scriptsize$\pm1.7$   \\
Mixup                      & 89.5 \scriptsize$\pm0.4$   & 78.2 \scriptsize$\pm0.4$   & 83.9 \scriptsize$\pm0.6$   & 71.6 \scriptsize$\pm1.3$   & 85.9 \scriptsize$\pm0.4$   & 78.8 \scriptsize$\pm0.6$   & 88.9 \scriptsize$\pm0.3$   & 88.9 \scriptsize$\pm0.3$   & 94.7 \scriptsize$\pm0.2$   & 7.0 \scriptsize$\pm0.6$    \\
GroupDRO                   & 86.9 \scriptsize$\pm0.9$   & 73.1 \scriptsize$\pm0.4$   & 80.7 \scriptsize$\pm1.1$   & 66.1 \scriptsize$\pm2.2$   & 82.8 \scriptsize$\pm0.9$   & 74.4 \scriptsize$\pm1.2$   & 86.3 \scriptsize$\pm0.5$   & 86.3 \scriptsize$\pm0.5$   & 94.0 \scriptsize$\pm0.3$   & 10.5 \scriptsize$\pm0.8$   \\
CVaRDRO                    & 89.9 \scriptsize$\pm0.4$   & 75.5 \scriptsize$\pm2.2$   & 84.5 \scriptsize$\pm0.7$   & 73.2 \scriptsize$\pm1.7$   & 86.2 \scriptsize$\pm0.3$   & 79.0 \scriptsize$\pm0.4$   & 88.5 \scriptsize$\pm0.3$   & 88.5 \scriptsize$\pm0.3$   & 95.4 \scriptsize$\pm0.2$   & 8.3 \scriptsize$\pm0.2$    \\
JTT                        & 88.9 \scriptsize$\pm0.6$   & 71.0 \scriptsize$\pm0.5$   & 83.2 \scriptsize$\pm0.8$   & 71.5 \scriptsize$\pm1.7$   & 84.7 \scriptsize$\pm0.6$   & 76.8 \scriptsize$\pm0.8$   & 86.8 \scriptsize$\pm0.2$   & 86.8 \scriptsize$\pm0.2$   & 94.2 \scriptsize$\pm0.1$   & 9.1 \scriptsize$\pm0.3$    \\
\end{tabular}}
\end{center}

\subsubsection{CelebA}

\vspace{-10pt}
\begin{center}
\adjustbox{max width=\textwidth}{%
\begin{tabular}{lcccccccccc}
\toprule[1.5pt]
\textbf{Algorithm}         & \textbf{Avg Acc.}               & \textbf{Worst Acc.}             & \textbf{Avg Prec.}           & \textbf{Worst Prec.}         & \textbf{Avg F1}             & \textbf{Worst F1}           & \textbf{Adjusted Acc.}          & \textbf{Balanced Acc.}          & \textbf{AUROC}             & \textbf{ECE}               \\
\midrule
ERM                        & 95.0 \scriptsize$\pm0.1$   & 57.6 \scriptsize$\pm0.8$   & 87.4 \scriptsize$\pm0.2$   & 76.3 \scriptsize$\pm0.3$   & 89.9 \scriptsize$\pm0.1$   & 82.7 \scriptsize$\pm0.2$   & 85.6 \scriptsize$\pm0.2$   & 93.0 \scriptsize$\pm0.0$   & 98.4 \scriptsize$\pm0.0$   & 2.9 \scriptsize$\pm0.1$    \\
Mixup                      & 95.4 \scriptsize$\pm0.1$   & 57.8 \scriptsize$\pm0.8$   & 88.4 \scriptsize$\pm0.3$   & 78.5 \scriptsize$\pm0.7$   & 90.6 \scriptsize$\pm0.2$   & 83.8 \scriptsize$\pm0.3$   & 85.8 \scriptsize$\pm0.2$   & 93.1 \scriptsize$\pm0.1$   & 98.4 \scriptsize$\pm0.1$   & 2.5 \scriptsize$\pm0.2$    \\
GroupDRO                   & 92.4 \scriptsize$\pm0.2$   & 78.5 \scriptsize$\pm1.1$   & 81.9 \scriptsize$\pm0.3$   & 64.5 \scriptsize$\pm0.5$   & 86.3 \scriptsize$\pm0.2$   & 77.1 \scriptsize$\pm0.4$   & 90.1 \scriptsize$\pm0.3$   & 93.9 \scriptsize$\pm0.1$   & 98.5 \scriptsize$\pm0.0$   & 7.4 \scriptsize$\pm0.5$    \\
CVaRDRO                    & 95.1 \scriptsize$\pm0.1$   & 62.2 \scriptsize$\pm3.1$   & 87.8 \scriptsize$\pm0.2$   & 77.1 \scriptsize$\pm0.5$   & 90.2 \scriptsize$\pm0.1$   & 83.2 \scriptsize$\pm0.1$   & 86.8 \scriptsize$\pm0.7$   & 93.2 \scriptsize$\pm0.2$   & 98.4 \scriptsize$\pm0.1$   & 3.0 \scriptsize$\pm0.2$    \\
JTT                        & 88.1 \scriptsize$\pm0.5$   & 66.0 \scriptsize$\pm11.9$  & 75.2 \scriptsize$\pm1.1$   & 54.1 \scriptsize$\pm1.0$   & 76.7 \scriptsize$\pm3.3$   & 60.5 \scriptsize$\pm6.6$   & 81.7 \scriptsize$\pm5.5$   & 82.7 \scriptsize$\pm6.7$   & 91.3 \scriptsize$\pm0.2$   & 5.0 \scriptsize$\pm1.0$    \\
\end{tabular}}
\end{center}

\subsubsection{CivilComments}

\vspace{-10pt}
\begin{center}
\adjustbox{max width=\textwidth}{%
\begin{tabular}{lcccccccccc}
\toprule[1.5pt]
\textbf{Algorithm}         & \textbf{Avg Acc.}               & \textbf{Worst Acc.}             & \textbf{Avg Prec.}           & \textbf{Worst Prec.}         & \textbf{Avg F1}             & \textbf{Worst F1}           & \textbf{Adjusted Acc.}          & \textbf{Balanced Acc.}          & \textbf{AUROC}             & \textbf{ECE}               \\
\midrule
ERM                        & 85.4 \scriptsize$\pm0.2$   & 63.2 \scriptsize$\pm1.2$   & 75.4 \scriptsize$\pm0.3$   & 57.6 \scriptsize$\pm0.7$   & 77.0 \scriptsize$\pm0.1$   & 63.2 \scriptsize$\pm0.1$   & 77.7 \scriptsize$\pm0.1$   & 79.4 \scriptsize$\pm0.2$   & 89.8 \scriptsize$\pm0.1$   & 7.8 \scriptsize$\pm0.4$    \\
Mixup                      & 84.9 \scriptsize$\pm0.3$   & 66.1 \scriptsize$\pm1.3$   & 74.8 \scriptsize$\pm0.3$   & 56.2 \scriptsize$\pm0.8$   & 76.7 \scriptsize$\pm0.2$   & 62.8 \scriptsize$\pm0.3$   & 78.0 \scriptsize$\pm0.1$   & 79.6 \scriptsize$\pm0.1$   & 89.8 \scriptsize$\pm0.1$   & 9.0 \scriptsize$\pm0.6$    \\
GroupDRO                   & 83.1 \scriptsize$\pm0.3$   & 69.5 \scriptsize$\pm0.7$   & 73.2 \scriptsize$\pm0.2$   & 52.0 \scriptsize$\pm0.6$   & 75.7 \scriptsize$\pm0.2$   & 62.2 \scriptsize$\pm0.2$   & 79.1 \scriptsize$\pm0.1$   & 80.9 \scriptsize$\pm0.1$   & 89.9 \scriptsize$\pm0.1$   & 12.3 \scriptsize$\pm0.5$   \\
CVaRDRO                    & 83.5 \scriptsize$\pm0.3$   & 68.7 \scriptsize$\pm1.3$   & 73.5 \scriptsize$\pm0.3$   & 52.8 \scriptsize$\pm0.6$   & 75.9 \scriptsize$\pm0.2$   & 62.4 \scriptsize$\pm0.2$   & 78.6 \scriptsize$\pm0.2$   & 80.7 \scriptsize$\pm0.1$   & 89.8 \scriptsize$\pm0.1$   & 32.9 \scriptsize$\pm0.4$   \\
JTT                        & 83.3 \scriptsize$\pm0.1$   & 64.3 \scriptsize$\pm1.5$   & 72.8 \scriptsize$\pm0.1$   & 52.4 \scriptsize$\pm0.3$   & 74.8 \scriptsize$\pm0.1$   & 60.3 \scriptsize$\pm0.2$   & 76.8 \scriptsize$\pm0.2$   & 78.4 \scriptsize$\pm0.2$   & 88.2 \scriptsize$\pm0.1$   & 10.2 \scriptsize$\pm0.3$   \\
\end{tabular}}
\end{center}

\subsubsection{MultiNLI}

\vspace{-10pt}
\begin{center}
\adjustbox{max width=\textwidth}{%
\begin{tabular}{lcccccccccc}
\toprule[1.5pt]
\textbf{Algorithm}         & \textbf{Avg Acc.}               & \textbf{Worst Acc.}             & \textbf{Avg Prec.}           & \textbf{Worst Prec.}         & \textbf{Avg F1}             & \textbf{Worst F1}           & \textbf{Adjusted Acc.}          & \textbf{Balanced Acc.}          & \textbf{AUROC}             & \textbf{ECE}               \\
\midrule
ERM                        & 80.9 \scriptsize$\pm0.3$   & 69.5 \scriptsize$\pm0.3$   & 81.3 \scriptsize$\pm0.3$   & 73.7 \scriptsize$\pm0.4$   & 81.0 \scriptsize$\pm0.3$   & 77.6 \scriptsize$\pm0.3$   & 79.7 \scriptsize$\pm0.3$   & 80.9 \scriptsize$\pm0.3$   & 93.6 \scriptsize$\pm0.1$   & 10.7 \scriptsize$\pm0.5$   \\
Mixup                      & 81.4 \scriptsize$\pm0.3$   & 68.5 \scriptsize$\pm0.6$   & 81.6 \scriptsize$\pm0.3$   & 76.0 \scriptsize$\pm0.5$   & 81.4 \scriptsize$\pm0.3$   & 78.0 \scriptsize$\pm0.2$   & 80.1 \scriptsize$\pm0.3$   & 81.4 \scriptsize$\pm0.3$   & 93.6 \scriptsize$\pm0.1$   & 9.4 \scriptsize$\pm0.9$    \\
GroupDRO                   & 81.0 \scriptsize$\pm0.0$   & 69.3 \scriptsize$\pm1.5$   & 81.3 \scriptsize$\pm0.1$   & 74.9 \scriptsize$\pm0.6$   & 81.1 \scriptsize$\pm0.1$   & 77.7 \scriptsize$\pm0.1$   & 79.5 \scriptsize$\pm0.3$   & 81.0 \scriptsize$\pm0.0$   & 93.8 \scriptsize$\pm0.0$   & 8.7 \scriptsize$\pm0.4$    \\
CVaRDRO                    & 75.1 \scriptsize$\pm0.1$   & 63.0 \scriptsize$\pm1.5$   & 76.2 \scriptsize$\pm0.2$   & 65.6 \scriptsize$\pm0.2$   & 75.2 \scriptsize$\pm0.1$   & 72.1 \scriptsize$\pm0.2$   & 74.2 \scriptsize$\pm0.4$   & 75.1 \scriptsize$\pm0.1$   & 86.3 \scriptsize$\pm0.2$   & 41.4 \scriptsize$\pm0.1$   \\
JTT                        & 81.4 \scriptsize$\pm0.1$   & 68.4 \scriptsize$\pm0.6$   & 81.6 \scriptsize$\pm0.1$   & 75.7 \scriptsize$\pm0.3$   & 81.5 \scriptsize$\pm0.1$   & 78.1 \scriptsize$\pm0.1$   & 80.2 \scriptsize$\pm0.2$   & 81.4 \scriptsize$\pm0.1$   & 93.9 \scriptsize$\pm0.0$   & 9.4 \scriptsize$\pm0.5$    \\
\end{tabular}}
\end{center}

\subsubsection{MetaShift}

\begin{center}
\adjustbox{max width=\textwidth}{%
\begin{tabular}{lcccccccccc}
\toprule[1.5pt]
\textbf{Algorithm}         & \textbf{Avg Acc.}               & \textbf{Worst Acc.}             & \textbf{Avg Prec.}           & \textbf{Worst Prec.}         & \textbf{Avg F1}             & \textbf{Worst F1}           & \textbf{Adjusted Acc.}          & \textbf{Balanced Acc.}          & \textbf{AUROC}             & \textbf{ECE}               \\
\midrule
ERM                        & 91.5 \scriptsize$\pm0.2$   & 82.1 \scriptsize$\pm0.8$   & 91.5 \scriptsize$\pm0.1$   & 90.7 \scriptsize$\pm0.3$   & 91.4 \scriptsize$\pm0.2$   & 90.9 \scriptsize$\pm0.2$   & 89.4 \scriptsize$\pm0.3$   & 91.4 \scriptsize$\pm0.2$   & 97.5 \scriptsize$\pm0.1$   & 5.7 \scriptsize$\pm0.4$    \\
Mixup                      & 91.4 \scriptsize$\pm0.2$   & 79.0 \scriptsize$\pm0.8$   & 91.4 \scriptsize$\pm0.2$   & 90.9 \scriptsize$\pm0.2$   & 91.3 \scriptsize$\pm0.2$   & 90.8 \scriptsize$\pm0.2$   & 88.7 \scriptsize$\pm0.1$   & 91.3 \scriptsize$\pm0.2$   & 97.2 \scriptsize$\pm0.0$   & 1.7 \scriptsize$\pm0.2$    \\
GroupDRO                   & 91.5 \scriptsize$\pm0.3$   & 82.6 \scriptsize$\pm1.1$   & 91.5 \scriptsize$\pm0.3$   & 90.8 \scriptsize$\pm0.3$   & 91.5 \scriptsize$\pm0.3$   & 91.0 \scriptsize$\pm0.3$   & 89.5 \scriptsize$\pm0.4$   & 91.5 \scriptsize$\pm0.3$   & 97.5 \scriptsize$\pm0.1$   & 5.9 \scriptsize$\pm0.5$    \\
CVaRDRO                    & 91.5 \scriptsize$\pm0.2$   & 82.6 \scriptsize$\pm1.1$   & 91.5 \scriptsize$\pm0.2$   & 90.7 \scriptsize$\pm0.5$   & 91.5 \scriptsize$\pm0.2$   & 91.0 \scriptsize$\pm0.2$   & 89.5 \scriptsize$\pm0.4$   & 91.5 \scriptsize$\pm0.2$   & 97.5 \scriptsize$\pm0.1$   & 7.8 \scriptsize$\pm2.1$    \\
JTT                        & 91.5 \scriptsize$\pm0.2$   & 82.6 \scriptsize$\pm0.4$   & 91.5 \scriptsize$\pm0.2$   & 91.0 \scriptsize$\pm0.1$   & 91.5 \scriptsize$\pm0.2$   & 90.9 \scriptsize$\pm0.3$   & 89.6 \scriptsize$\pm0.1$   & 91.4 \scriptsize$\pm0.3$   & 97.6 \scriptsize$\pm0.1$   & 6.5 \scriptsize$\pm0.1$    \\
\end{tabular}}
\end{center}

\subsubsection{ImageNetBG}

\vspace{-10pt}
\begin{center}
\adjustbox{max width=\textwidth}{%
\begin{tabular}{lcccccccccc}
\toprule[1.5pt]
\textbf{Algorithm}         & \textbf{Avg Acc.}               & \textbf{Worst Acc.}             & \textbf{Avg Prec.}           & \textbf{Worst Prec.}         & \textbf{Avg F1}             & \textbf{Worst F1}           & \textbf{Adjusted Acc.}          & \textbf{Balanced Acc.}          & \textbf{AUROC}             & \textbf{ECE}               \\
\midrule
ERM                        & 87.7 \scriptsize$\pm0.1$   & 76.8 \scriptsize$\pm0.9$   & 87.9 \scriptsize$\pm0.1$   & 78.3 \scriptsize$\pm1.3$   & 87.7 \scriptsize$\pm0.1$   & 80.7 \scriptsize$\pm0.4$   & 87.7 \scriptsize$\pm0.1$   & 87.7 \scriptsize$\pm0.1$   & 99.0 \scriptsize$\pm0.0$   & 5.1 \scriptsize$\pm0.1$    \\
Mixup                      & 87.9 \scriptsize$\pm0.2$   & 76.9 \scriptsize$\pm0.7$   & 88.4 \scriptsize$\pm0.1$   & 76.6 \scriptsize$\pm2.6$   & 88.0 \scriptsize$\pm0.2$   & 80.5 \scriptsize$\pm1.0$   & 87.9 \scriptsize$\pm0.2$   & 87.9 \scriptsize$\pm0.2$   & 98.7 \scriptsize$\pm0.0$   & 4.7 \scriptsize$\pm1.6$    \\
GroupDRO                   & 87.7 \scriptsize$\pm0.1$   & 76.4 \scriptsize$\pm0.2$   & 87.9 \scriptsize$\pm0.1$   & 76.2 \scriptsize$\pm0.5$   & 87.6 \scriptsize$\pm0.1$   & 81.1 \scriptsize$\pm0.3$   & 87.7 \scriptsize$\pm0.1$   & 87.7 \scriptsize$\pm0.1$   & 99.0 \scriptsize$\pm0.0$   & 5.1 \scriptsize$\pm0.1$    \\
CVaRDRO                    & 87.8 \scriptsize$\pm0.2$   & 74.8 \scriptsize$\pm0.8$   & 88.0 \scriptsize$\pm0.1$   & 79.9 \scriptsize$\pm0.9$   & 87.8 \scriptsize$\pm0.2$   & 79.5 \scriptsize$\pm0.3$   & 87.8 \scriptsize$\pm0.2$   & 87.8 \scriptsize$\pm0.2$   & 99.0 \scriptsize$\pm0.0$   & 5.6 \scriptsize$\pm0.2$    \\
JTT                        & 87.6 \scriptsize$\pm0.4$   & 77.0 \scriptsize$\pm0.4$   & 87.8 \scriptsize$\pm0.3$   & 78.3 \scriptsize$\pm3.0$   & 87.5 \scriptsize$\pm0.3$   & 80.4 \scriptsize$\pm0.6$   & 87.6 \scriptsize$\pm0.4$   & 87.6 \scriptsize$\pm0.4$   & 99.0 \scriptsize$\pm0.0$   & 3.7 \scriptsize$\pm0.2$    \\
\end{tabular}}
\end{center}

\subsubsection{NICO++}

\vspace{-10pt}
\begin{center}
\adjustbox{max width=\textwidth}{%
\begin{tabular}{lcccccccccc}
\toprule[1.5pt]
\textbf{Algorithm}         & \textbf{Avg Acc.}               & \textbf{Worst Acc.}             & \textbf{Avg Prec.}           & \textbf{Worst Prec.}         & \textbf{Avg F1}             & \textbf{Worst F1}           & \textbf{Adjusted Acc.}          & \textbf{Balanced Acc.}          & \textbf{AUROC}             & \textbf{ECE}               \\
\midrule
ERM                        & 85.3 \scriptsize$\pm0.3$   & 40.0 \scriptsize$\pm0.0$   & 86.2 \scriptsize$\pm0.2$   & 59.2 \scriptsize$\pm1.1$   & 85.4 \scriptsize$\pm0.3$   & 66.0 \scriptsize$\pm0.6$   & 84.8 \scriptsize$\pm0.3$   & 85.2 \scriptsize$\pm0.3$   & 99.4 \scriptsize$\pm0.0$   & 9.6 \scriptsize$\pm0.2$    \\
Mixup                      & 85.5 \scriptsize$\pm0.2$   & 30.0 \scriptsize$\pm4.1$   & 86.5 \scriptsize$\pm0.1$   & 55.7 \scriptsize$\pm2.5$   & 85.7 \scriptsize$\pm0.2$   & 66.5 \scriptsize$\pm1.6$   & 85.0 \scriptsize$\pm0.2$   & 85.4 \scriptsize$\pm0.2$   & 99.1 \scriptsize$\pm0.0$   & 2.0 \scriptsize$\pm0.3$    \\
GroupDRO                   & 83.0 \scriptsize$\pm0.3$   & 31.1 \scriptsize$\pm0.9$   & 84.5 \scriptsize$\pm0.2$   & 54.4 \scriptsize$\pm3.9$   & 83.2 \scriptsize$\pm0.3$   & 64.9 \scriptsize$\pm0.9$   & 82.6 \scriptsize$\pm0.3$   & 82.9 \scriptsize$\pm0.3$   & 99.3 \scriptsize$\pm0.0$   & 7.4 \scriptsize$\pm0.5$    \\
CVaRDRO                    & 84.4 \scriptsize$\pm0.6$   & 31.7 \scriptsize$\pm3.6$   & 85.5 \scriptsize$\pm0.5$   & 57.1 \scriptsize$\pm1.6$   & 84.6 \scriptsize$\pm0.6$   & 66.6 \scriptsize$\pm1.5$   & 84.0 \scriptsize$\pm0.6$   & 84.3 \scriptsize$\pm0.6$   & 99.4 \scriptsize$\pm0.0$   & 7.8 \scriptsize$\pm1.3$    \\
JTT                        & 85.4 \scriptsize$\pm0.2$   & 32.2 \scriptsize$\pm0.9$   & 86.2 \scriptsize$\pm0.2$   & 57.8 \scriptsize$\pm4.6$   & 85.5 \scriptsize$\pm0.2$   & 65.1 \scriptsize$\pm2.8$   & 84.9 \scriptsize$\pm0.2$   & 85.3 \scriptsize$\pm0.2$   & 99.4 \scriptsize$\pm0.0$   & 10.2 \scriptsize$\pm0.2$   \\
\end{tabular}}
\end{center}

\subsubsection{MIMIC-CXR}

\vspace{-10pt}
\begin{center}
\adjustbox{max width=\textwidth}{%
\begin{tabular}{lcccccccccc}
\toprule[1.5pt]
\textbf{Algorithm}         & \textbf{Avg Acc.}               & \textbf{Worst Acc.}             & \textbf{Avg Prec.}           & \textbf{Worst Prec.}         & \textbf{Avg F1}             & \textbf{Worst F1}           & \textbf{Adjusted Acc.}          & \textbf{Balanced Acc.}          & \textbf{AUROC}             & \textbf{ECE}               \\
\midrule
ERM                        & 78.6 \scriptsize$\pm0.2$   & 68.5 \scriptsize$\pm0.4$   & 77.7 \scriptsize$\pm0.2$   & 72.2 \scriptsize$\pm0.5$   & 77.8 \scriptsize$\pm0.2$   & 73.7 \scriptsize$\pm0.2$   & 77.5 \scriptsize$\pm0.2$   & 78.1 \scriptsize$\pm0.2$   & 85.5 \scriptsize$\pm0.2$   & 2.6 \scriptsize$\pm0.2$    \\
Mixup                      & 78.5 \scriptsize$\pm0.1$   & 67.2 \scriptsize$\pm1.1$   & 77.5 \scriptsize$\pm0.1$   & 72.3 \scriptsize$\pm0.5$   & 77.6 \scriptsize$\pm0.0$   & 73.4 \scriptsize$\pm0.1$   & 77.2 \scriptsize$\pm0.1$   & 77.8 \scriptsize$\pm0.1$   & 85.3 \scriptsize$\pm0.1$   & 2.4 \scriptsize$\pm0.7$    \\
GroupDRO                   & 78.3 \scriptsize$\pm0.1$   & 67.4 \scriptsize$\pm0.9$   & 77.4 \scriptsize$\pm0.1$   & 71.9 \scriptsize$\pm0.2$   & 77.6 \scriptsize$\pm0.1$   & 73.4 \scriptsize$\pm0.1$   & 77.2 \scriptsize$\pm0.1$   & 77.8 \scriptsize$\pm0.1$   & 85.1 \scriptsize$\pm0.0$   & 3.7 \scriptsize$\pm0.4$    \\
CVaRDRO                    & 78.4 \scriptsize$\pm0.3$   & 67.5 \scriptsize$\pm0.1$   & 77.4 \scriptsize$\pm0.3$   & 72.1 \scriptsize$\pm0.6$   & 77.6 \scriptsize$\pm0.3$   & 73.4 \scriptsize$\pm0.2$   & 77.2 \scriptsize$\pm0.2$   & 77.7 \scriptsize$\pm0.2$   & 84.9 \scriptsize$\pm0.2$   & 5.7 \scriptsize$\pm0.3$    \\
JTT                        & 78.2 \scriptsize$\pm0.1$   & 66.6 \scriptsize$\pm0.8$   & 77.3 \scriptsize$\pm0.1$   & 71.9 \scriptsize$\pm0.3$   & 77.4 \scriptsize$\pm0.1$   & 73.2 \scriptsize$\pm0.1$   & 77.0 \scriptsize$\pm0.1$   & 77.6 \scriptsize$\pm0.1$   & 85.0 \scriptsize$\pm0.1$   & 3.3 \scriptsize$\pm0.1$    \\
\end{tabular}}
\end{center}

\subsubsection{MIMICNotes}

\vspace{-10pt}
\begin{center}
\adjustbox{max width=\textwidth}{%
\begin{tabular}{lcccccccccc}
\toprule[1.5pt]
\textbf{Algorithm}         & \textbf{Avg Acc.}               & \textbf{Worst Acc.}             & \textbf{Avg Prec.}           & \textbf{Worst Prec.}         & \textbf{Avg F1}             & \textbf{Worst F1}           & \textbf{Adjusted Acc.}          & \textbf{Balanced Acc.}          & \textbf{AUROC}             & \textbf{ECE}               \\
\midrule
ERM                        & 91.1 \scriptsize$\pm0.1$   & 24.2 \scriptsize$\pm2.8$   & 76.4 \scriptsize$\pm0.9$   & 60.3 \scriptsize$\pm2.1$   & 65.8 \scriptsize$\pm1.2$   & 36.5 \scriptsize$\pm2.5$   & 62.2 \scriptsize$\pm1.3$   & 62.4 \scriptsize$\pm1.4$   & 85.2 \scriptsize$\pm0.1$   & 2.2 \scriptsize$\pm0.1$    \\
Mixup                      & 91.1 \scriptsize$\pm0.0$   & 22.7 \scriptsize$\pm3.2$   & 76.8 \scriptsize$\pm0.7$   & 61.2 \scriptsize$\pm1.6$   & 65.1 \scriptsize$\pm1.6$   & 35.0 \scriptsize$\pm3.2$   & 61.5 \scriptsize$\pm1.6$   & 61.7 \scriptsize$\pm1.7$   & 85.4 \scriptsize$\pm0.0$   & 2.0 \scriptsize$\pm0.8$    \\
GroupDRO                   & 83.2 \scriptsize$\pm2.4$   & 62.6 \scriptsize$\pm6.3$   & 64.7 \scriptsize$\pm1.4$   & 33.7 \scriptsize$\pm3.4$   & 66.4 \scriptsize$\pm1.3$   & 42.8 \scriptsize$\pm1.1$   & 74.3 \scriptsize$\pm1.5$   & 74.4 \scriptsize$\pm1.5$   & 85.1 \scriptsize$\pm0.1$   & 13.8 \scriptsize$\pm3.2$   \\
CVaRDRO                    & 90.2 \scriptsize$\pm0.0$   & 0.0 \scriptsize$\pm0.0$    & 45.1 \scriptsize$\pm0.0$   & 0.0 \scriptsize$\pm0.0$    & 47.4 \scriptsize$\pm0.0$   & 0.0 \scriptsize$\pm0.0$    & 50.0 \scriptsize$\pm0.0$   & 50.0 \scriptsize$\pm0.0$   & 71.6 \scriptsize$\pm0.4$   & 40.2 \scriptsize$\pm0.0$   \\
JTT                        & 71.3 \scriptsize$\pm3.7$   & 65.9 \scriptsize$\pm2.8$   & 60.3 \scriptsize$\pm0.7$   & 23.4 \scriptsize$\pm1.9$   & 58.6 \scriptsize$\pm2.3$   & 36.0 \scriptsize$\pm1.8$   & 75.5 \scriptsize$\pm0.4$   & 75.6 \scriptsize$\pm0.4$   & 84.9 \scriptsize$\pm0.1$   & 27.5 \scriptsize$\pm3.9$   \\
\end{tabular}}
\end{center}

\subsubsection{CXRMultisite}

\vspace{-10pt}
\begin{center}
\adjustbox{max width=\textwidth}{%
\begin{tabular}{lcccccccccc}
\toprule[1.5pt]
\textbf{Algorithm}         & \textbf{Avg Acc.}               & \textbf{Worst Acc.}             & \textbf{Avg Prec.}           & \textbf{Worst Prec.}         & \textbf{Avg F1}             & \textbf{Worst F1}           & \textbf{Adjusted Acc.}          & \textbf{Balanced Acc.}          & \textbf{AUROC}             & \textbf{ECE}               \\
\midrule
ERM                        & 98.3 \scriptsize$\pm0.0$   & 0.0 \scriptsize$\pm0.0$    & 49.2 \scriptsize$\pm0.0$   & 0.0 \scriptsize$\pm0.0$    & 49.6 \scriptsize$\pm0.0$   & 0.0 \scriptsize$\pm0.0$    & 50.0 \scriptsize$\pm0.0$   & 50.0 \scriptsize$\pm0.0$   & 93.1 \scriptsize$\pm0.1$   & 0.3 \scriptsize$\pm0.1$    \\
Mixup                      & 98.3 \scriptsize$\pm0.0$   & 0.0 \scriptsize$\pm0.0$    & 49.2 \scriptsize$\pm0.0$   & 0.0 \scriptsize$\pm0.0$    & 49.6 \scriptsize$\pm0.0$   & 0.0 \scriptsize$\pm0.0$    & 50.0 \scriptsize$\pm0.0$   & 50.0 \scriptsize$\pm0.0$   & 92.9 \scriptsize$\pm0.1$   & 0.3 \scriptsize$\pm0.0$    \\
GroupDRO                   & 90.2 \scriptsize$\pm0.1$   & 0.0 \scriptsize$\pm0.0$    & 56.7 \scriptsize$\pm0.1$   & 13.6 \scriptsize$\pm0.1$   & 59.2 \scriptsize$\pm0.1$   & 23.7 \scriptsize$\pm0.2$   & 50.2 \scriptsize$\pm0.1$   & 90.4 \scriptsize$\pm0.0$   & 92.8 \scriptsize$\pm0.2$   & 13.5 \scriptsize$\pm0.7$   \\
CVaRDRO                    & 98.3 \scriptsize$\pm0.0$   & 0.0 \scriptsize$\pm0.0$    & 61.2 \scriptsize$\pm4.9$   & 24.0 \scriptsize$\pm9.8$   & 50.7 \scriptsize$\pm0.7$   & 2.2 \scriptsize$\pm1.5$    & 50.2 \scriptsize$\pm0.2$   & 50.6 \scriptsize$\pm0.4$   & 93.0 \scriptsize$\pm0.0$   & 0.9 \scriptsize$\pm0.3$    \\
JTT                        & 94.1 \scriptsize$\pm0.9$   & 0.0 \scriptsize$\pm0.0$    & 59.0 \scriptsize$\pm0.7$   & 18.5 \scriptsize$\pm1.4$   & 62.9 \scriptsize$\pm0.8$   & 28.9 \scriptsize$\pm1.2$   & 55.2 \scriptsize$\pm0.9$   & 82.2 \scriptsize$\pm2.4$   & 93.2 \scriptsize$\pm0.1$   & 6.4 \scriptsize$\pm0.5$    \\
\end{tabular}}
\end{center}

\subsubsection{CheXpert}

\vspace{-10pt}
\begin{center}
\adjustbox{max width=\textwidth}{%
\begin{tabular}{lcccccccccc}
\toprule[1.5pt]
\textbf{Algorithm}         & \textbf{Avg Acc.}               & \textbf{Worst Acc.}             & \textbf{Avg Prec.}           & \textbf{Worst Prec.}         & \textbf{Avg F1}             & \textbf{Worst F1}           & \textbf{Adjusted Acc.}          & \textbf{Balanced Acc.}          & \textbf{AUROC}             & \textbf{ECE}               \\
\midrule
ERM                        & 88.6 \scriptsize$\pm0.7$   & 41.7 \scriptsize$\pm3.4$   & 68.3 \scriptsize$\pm1.1$   & 42.3 \scriptsize$\pm2.4$   & 68.7 \scriptsize$\pm0.2$   & 43.8 \scriptsize$\pm0.7$   & 69.5 \scriptsize$\pm1.3$   & 70.0 \scriptsize$\pm1.5$   & 85.4 \scriptsize$\pm0.4$   & 5.0 \scriptsize$\pm1.1$    \\
Mixup                      & 81.9 \scriptsize$\pm6.2$   & 37.4 \scriptsize$\pm3.5$   & 63.5 \scriptsize$\pm5.0$   & 33.9 \scriptsize$\pm9.3$   & 62.5 \scriptsize$\pm5.9$   & 35.7 \scriptsize$\pm7.6$   & 63.8 \scriptsize$\pm4.7$   & 64.1 \scriptsize$\pm4.6$   & 76.1 \scriptsize$\pm8.5$   & 16.1 \scriptsize$\pm9.0$   \\
GroupDRO                   & 79.0 \scriptsize$\pm0.3$   & 75.1 \scriptsize$\pm0.6$   & 62.9 \scriptsize$\pm0.1$   & 28.5 \scriptsize$\pm0.3$   & 64.6 \scriptsize$\pm0.2$   & 42.0 \scriptsize$\pm0.3$   & 78.6 \scriptsize$\pm0.0$   & 79.3 \scriptsize$\pm0.1$   & 86.3 \scriptsize$\pm0.1$   & 22.0 \scriptsize$\pm0.4$   \\
CVaRDRO                    & 73.7 \scriptsize$\pm1.0$   & 50.2 \scriptsize$\pm1.8$   & 57.3 \scriptsize$\pm0.1$   & 20.1 \scriptsize$\pm0.3$   & 56.8 \scriptsize$\pm0.4$   & 29.9 \scriptsize$\pm0.2$   & 65.7 \scriptsize$\pm0.6$   & 67.0 \scriptsize$\pm0.4$   & 72.9 \scriptsize$\pm0.4$   & 40.4 \scriptsize$\pm0.0$   \\
JTT                        & 75.2 \scriptsize$\pm0.8$   & 60.4 \scriptsize$\pm4.8$   & 59.4 \scriptsize$\pm1.1$   & 23.0 \scriptsize$\pm1.5$   & 59.6 \scriptsize$\pm1.4$   & 34.4 \scriptsize$\pm2.3$   & 70.7 \scriptsize$\pm2.6$   & 72.0 \scriptsize$\pm2.5$   & 79.0 \scriptsize$\pm2.5$   & 24.4 \scriptsize$\pm0.5$   \\
\end{tabular}}
\end{center}

\subsubsection{Living17}

\vspace{-10pt}
\begin{center}
\adjustbox{max width=\textwidth}{%
%
}
\end{center}

\subsubsection{Overall}
\label{appendix-subsubsec:overall-table-val-known-selection}

\begin{center}
\adjustbox{max width=\textwidth}{%
%
}
\end{center}

\vspace{0.1cm}
\subsection{Attributes Unknown in Both Training \& Validation}
\label{appendix-subsec:complete-results-val-unknown-selection}

\subsubsection{Waterbirds}

\vspace{-10pt}
\begin{center}
\adjustbox{max width=\textwidth}{%
%
}
\end{center}

\subsubsection{CelebA}

\vspace{-10pt}
\begin{center}
\adjustbox{max width=\textwidth}{%
%
}
\end{center}

\subsubsection{CivilComments}

\vspace{-10pt}
\begin{center}
\adjustbox{max width=\textwidth}{%
%
}
\end{center}

\subsubsection{MultiNLI}

\vspace{-10pt}
\begin{center}
\adjustbox{max width=\textwidth}{%
%
}
\end{center}

\subsubsection{MetaShift}

\vspace{-10pt}
\begin{center}
\adjustbox{max width=\textwidth}{%
%
}
\end{center}

\subsubsection{ImageNetBG}

\vspace{-10pt}
\begin{center}
\adjustbox{max width=\textwidth}{%
%
}
\end{center}

\subsubsection{NICO++}

\vspace{-10pt}
\begin{center}
\adjustbox{max width=\textwidth}{%
%
}
\end{center}

\subsubsection{MIMIC-CXR}

\vspace{-10pt}
\begin{center}
\adjustbox{max width=\textwidth}{%
%
}
\end{center}

\subsubsection{MIMICNotes}

\vspace{-10pt}
\begin{center}
\adjustbox{max width=\textwidth}{%
%
}
\end{center}

\subsubsection{CXRMultisite}

\vspace{-10pt}
\begin{center}
\adjustbox{max width=\textwidth}{%
%
}
\end{center}

\subsubsection{CheXpert}

\vspace{-10pt}
\begin{center}
\adjustbox{max width=\textwidth}{%
%
}
\end{center}

\subsubsection{Living17}

\vspace{-10pt}
\begin{center}
\adjustbox{max width=\textwidth}{%
%
}
\end{center}

\subsubsection{Overall}
\label{appendix-subsubsec:overall-table-val-unknown-selection}

\begin{center}
\adjustbox{max width=\textwidth}{%
%
}
\end{center}

\end{document}